\documentclass[sigconf]{acmart}
\copyrightyear{2026}
\acmYear{2026}
\setcopyright{cc}
\setcctype{by-nc-nd}
\acmConference[KDD '26] {Proceedings of the 32nd ACM SIGKDD Conference on Knowledge Discovery and Data Mining V.2}{August 09--13, 2026}{Jeju Island, Republic of Korea}
\acmBooktitle{Proceedings of the 32nd ACM SIGKDD Conference on Knowledge Discovery and Data Mining V.2 (KDD '26), August 09--13, 2026, Jeju Island, Republic of Korea}
\acmISBN{979-8-4007-2259-2/2026/08}
\acmDOI{10.1145/3770855.3817449}
\usepackage{fontawesome}
\usepackage{listings}
\usepackage{booktabs}
\usepackage{multirow}
\usepackage{graphicx}
\usepackage{array}
\usepackage{makecell} 
\usepackage{xcolor}
\usepackage{threeparttable}
\usepackage{tabularx}
\usepackage{natbib}
\usepackage{pifont}
\lstdefinestyle{datasetstyle}{
    basicstyle=\ttfamily\small,   
    frame=single,                 
    backgroundcolor=\color{gray!10}, 
    breaklines=true,              
    postbreak=\mbox{\textcolor{red}{$\hookrightarrow$}\space}, 
    showstringspaces=false,       
}
\AtBeginDocument{%
  }

\begin{document}
\settopmatter{printfolios=true} 

\title[PipeMFL-240K: A Dataset and Benchmark for Object Detection in Pipeline Magnetic Flux Leakage Imaging]{PipeMFL-240K: A Large-scale Dataset and Benchmark for Object Detection in Pipeline Magnetic Flux Leakage Imaging}


\author{Tianyi Qu}
\authornote{Tianyi Qu, Songxiao Yang and Haolin Wang contributed equally to this research.}
\email{qutianyi@hb-sais.com}
\orcid{0009-0005-1064-7320}
\affiliation{%
  \institution{SINOMACH Sensing Technology \\Co., Ltd}
  \city{Shenyang}
  \state{Liaoning}
  \country{China}
}

\author{Songxiao Yang}
\authornotemark[1]
\email{syang@ok.sc.e.titech.ac.jp}
\orcid{0000-0001-9036-4817}
\affiliation{%
  \institution{Institute of Science Tokyo}
  \city{Tokyo}
  \country{Japan}
}

\author{Haolin Wang}
\authornotemark[1]
\email{haolin.wang.k3@elms.hokudai.ac.jp}
\orcid{0000-0002-5767-0019}
\affiliation{%
  \institution{Hokkaido University}
  \city{Sapporo}
  \state{Hokkaido}
  \country{Japan}
}

\author{Huadong Song}
\email{songhuadong@hb-sais.com}
\orcid{0009-0002-0376-6055}
\affiliation{%
  \institution{SINOMACH Sensing Technology \\Co., Ltd}
  \city{Shenyang}
  \state{Liaoning}
  \country{China}
}

\author{Xiaoting Guo}
\email{guoxiaoting@hb-sais.com}
\orcid{0000-0001-9198-4065}
\affiliation{%
  \institution{SINOMACH Sensing Technology \\Co., Ltd}
  \city{Shenyang}
  \state{Liaoning}
  \country{China}
}

\author{Wenguang Hu}
\email{huwenguang@hb-sais.com}
\orcid{0000-0002-4306-048X}
\affiliation{%
  \institution{SINOMACH Sensing Technology \\Co., Ltd}
  \city{Shenyang}
  \state{Liaoning}
  \country{China}
}

\author{Guanlin Liu}
\email{liuguanlin@hb-sais.com}
\orcid{0009-0001-7788-3142}
\affiliation{%
  \institution{SINOMACH Sensing Technology \\Co., Ltd}
  \city{Shenyang}
  \state{Liaoning}
  \country{China}
}

\author{Honghe Chen}
\email{chenhonghe@hb-sais.com}
\orcid{0009-0001-4527-1985}
\affiliation{%
  \institution{SINOMACH Sensing Technology \\Co., Ltd}
  \city{Shenyang}
  \state{Liaoning}
  \country{China}
}

\author{Yafei Ou}\authornote{Corresponding author: Yafei Ou (yafei.ou@riken.jp).}
\email{yafei.ou@riken.jp}
\orcid{0000-0001-7510-0813}
\affiliation{%
  \institution{RIKEN}
  \city{Tokyo}
  \country{Japan}
}
\renewcommand{\shortauthors}{Tianyi Qu et al.}

\begin{abstract}
Pipeline integrity is critical to industrial safety and environmental protection, with Magnetic Flux Leakage (MFL) detection being a primary non-destructive testing technology.
Despite the promise of deep learning for automating MFL interpretation, progress toward reliable models has been constrained by the absence of a large-scale public dataset and benchmark, making fair comparison and reproducible evaluation difficult. We introduce \textbf{PipeMFL-240K}, a large-scale, meticulously annotated dataset and benchmark for complex object detection in pipeline MFL pseudo-color images.
PipeMFL-240K reflects real-world inspection complexity and poses several unique challenges: (i) an extremely long-tailed distribution over \textbf{12} categories, (ii) a high prevalence of tiny objects that often comprise only a handful of pixels and (iii) substantial intra-class variability. The dataset contains \textbf{249,320} images and \textbf{200,020} high-quality bounding-box annotations, collected from 12 pipelines spanning approximately \textbf{1,530} km.
Extensive experiments are conducted with state-of-the-art object detectors to establish baselines. Results show that modern detectors still struggle with the intrinsic properties of MFL data, highlighting considerable headroom for improvement, while PipeMFL-240K provides a reliable and challenging testbed to drive future research. As the first public dataset and the first benchmark of this scale and scope for pipeline MFL inspection, it provides a critical foundation for efficient pipeline diagnostics as well as maintenance planning and is expected to accelerate algorithmic innovation and reproducible research in MFL-based pipeline integrity assessment.
  \\
  \small \textbf{\mbox{\faGithub\hspace{.25em} Benchmark \& Code:}} \href{https://github.com/TQSAIS/PipeMFL-240K}{github.com/TQSAIS/PipeMFL-240K}\\
  \raisebox{-0.3\height}{\includegraphics[width=0.35cm]{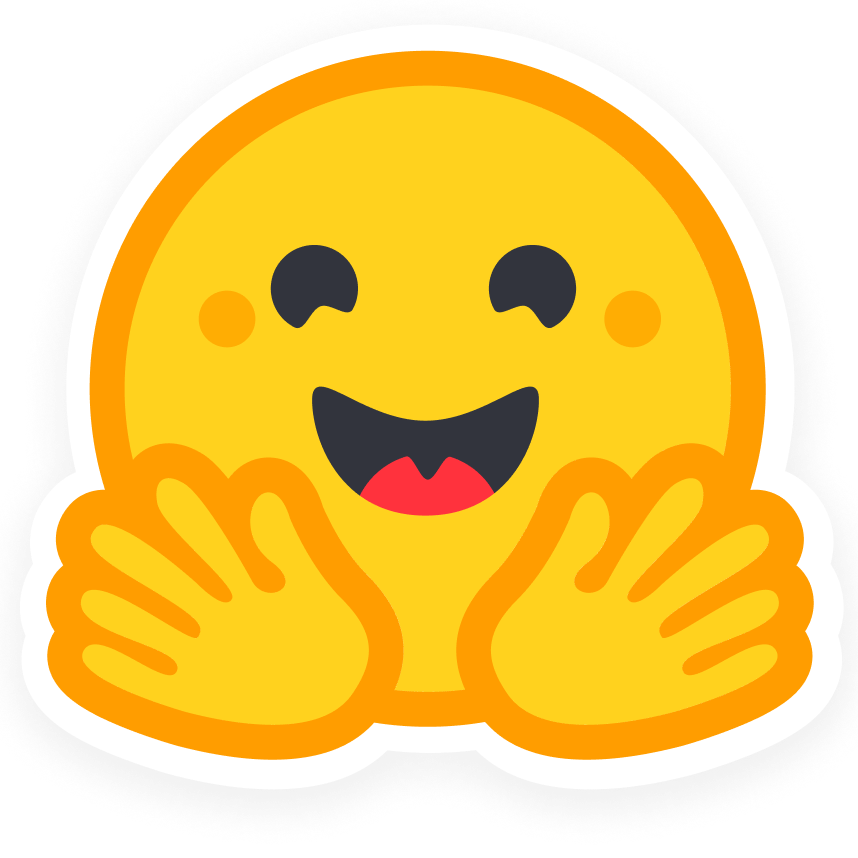}} \small \textbf{\mbox{Data \& Dataset Card:}} \href{https://huggingface.co/datasets/PipeMFL/PipeMFL-240K}{huggingface.co/datasets/PipeMFL/PipeMFL-240K}
\end{abstract}


\begin{teaserfigure}
  \includegraphics[width=\textwidth]{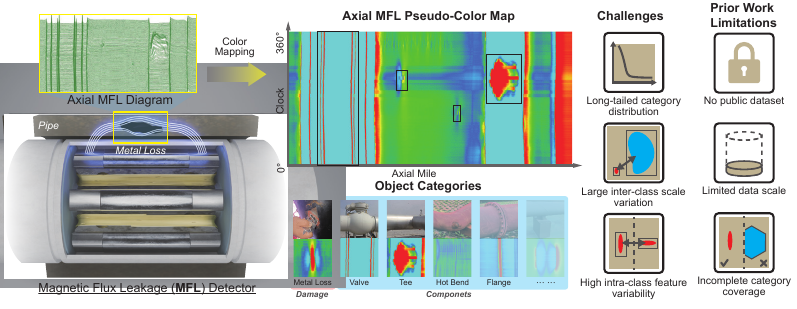}
  \caption{An overview of MFL detection, the key challenges it poses and the limitations of prior work.}
  \label{fig:teaser}
\end{teaserfigure}


\maketitle

\begin{table*}[!t]
\centering
\caption{Summary of related works on pipeline MFL image detection.}
\label{tab:related_works}
\begin{threeparttable}
\setlength{\tabcolsep}{1mm}{
\begin{tabular}{p{2.5cm}p{0.8cm}<{\centering}p{1cm}<{\centering}p{1.6cm}<{\centering}p{2cm}<{\centering}p{1cm}<{\centering}p{1.5cm}<{\centering}p{1.8cm}<{\centering}p{1.5cm}<{\centering}p{2cm}<{\centering}}
\toprule
\multirow{2.5}{*}{\textbf{Works}} & \multirow{2.5}{*}{\textbf{Year}} & \multirow{2.5}{*}{\textbf{Images}} & \multirow{2.5}{*}{\textbf{\makecell[c]{Image size\\{[}px$\times$px{]}}}} & \multirow{2.5}{*}{\textbf{\makecell[c]{Resolution\\{[}mm/px · °/px{]}}}} & \multirow{2.5}{*}{\textbf{Objects}} & \multicolumn{2}{c}{\textbf{Number of Categories}} & \multirow{2.5}{*}{\textbf{Pipe Type}} & \multirow{2.5}{*}{\textbf{\makecell[c]{Dataset\\Accessibility}}} \\ \cmidrule(lr){7-8}
 &  &  &  &  &  & \textbf{Damage} & \textbf{Component} &  &  \\
\midrule
Yang et al.~\cite{Yang2018SSD} & 2018 & 20,000 & 560$\times$420 & [- · 0.86]  & 3,100 & 1 & 2 & In-service & \ding{55}\\
Yuksel et al.~\cite{Yuksel2023single} & 2023 & 540 & - & - & 100 & 1 & 0 & In-service & \ding{55}\\
Zhao et al.~\cite{zhao2023single} & 2023 & - & - & - & 2,000 & 1 & 0 & Experimental & \ding{55}\\
Han et al.~\cite{Han2023Single} & 2023 & 24,633 & 640$\times$640 &  [- · 0.56]  & - & 1 & 0 & Experimental& \ding{55}\\
Xu et al.~\cite{Xu2023Image} & 2023 & 1,000 & -$\times$288 & [- · 1.25] & 2,483 & 2 & 3 & In-service & \ding{55}\\
Han et al.~\cite{Han2023anchor} & 2023 & 23,739 & 512$\times$512 & [- · 0.7]  & - & 1 & 2 & In-service & \ding{55}\\
Shen et al.~\cite{Shen2023multi} & 2024 & 841 & 432$\times$288 & [2 · 1.25] & - & 1 & 2  & Experimental& \ding{55}\\
Manzoor et al.~\cite{manzoor2024multimodal}& 2024 & 33,000 & 128$\times$128 & [- · 2.8125] & 33,000 & 1 & 0 & In-service& \ding{55}\\
Lin et al.~\cite{Lin2025single} & 2025 & 200 & - & - & - & 1 & 0 & Experimental& \ding{55}\\
Wang et al.~\cite{Wang2025single} & 2025 & 2,550 & 200$\times$200 & [- · 1.8]  & - & 1 & 0 &  Both & \ding{55}\\
Li et al.~\cite{li2025yolov8} & 2025 & 36,098 & 960$\times$640  & [4.16 · 0.56] & 8,194 & 1 & 0 & In-service & \ding{55}\\
Liu et al.~\cite{Liu2025Online} & 2025 & 841 & 7,640$\times$640  & [0.27 · 0.56] & - & 3 & 2 & Both & \ding{55} \\
Zhao et al.~\cite{Zhao2025masked} & 2025 & - & 1,333$\times$888 & [- · 0.41] & 7,898 & 1 & 6 & Experimental& \ding{55} \\
\textbf{This work} & \textbf{2026} & \textbf{249,320} & \textbf{5,000$\times$2,400} & \textbf{[1 · 0.15]} & \textbf{200,020} & \textbf{4} & \textbf{8} & \textbf{In-service} & \ding{51} \\
\bottomrule
\end{tabular}
}
\end{threeparttable}
\end{table*}

\section{Introduction}

Long-distance metal pipelines serve as critical infrastructure for the transportation of energy resources such as oil and natural gas, owing to their high efficiency, large capacity and ability to operate continuously across vast and remote terrains~\cite{peng2020intro, song2023intro}. With the majority of pipelines now entering a stable operational phase, ensuring their structural integrity and safety through regular inspection and maintenance has become increasingly imperative. 
As shown in Fig.~\ref{fig:teaser}, Magnetic Flux Leakage (MFL) detection operates by magnetizing the pipeline wall and detecting variations in magnetic flux caused by metal loss, weld anomalies, or other structural/component features ~\cite{tang17}. 
Due to its high inspection speed, strong penetration capability and robustness in various operating conditions, MFL stands as one of the most widely adopted non-destructive testing (NDT) methods for in-line pipeline inspection~\cite{Feng2022NDT}, particularly for ferromagnetic pipelines and corrosion-related anomalies~\cite{shi2015theory, Ma2025NDT}.

The original pipeline MFL signal comprises three-axis data, among which the axial component exhibits the most significant features~\cite{HWANG2000review}. When transformed into pseudo-color maps, the resulting images reveal several domain-specific characteristics: (i) The upper and lower boundaries of the image are physically connected, forming a circular data structure.
(ii) Certain categories frequently co-occur (e.g., valves and tees often appear within the same image, typically within an actual spatial distance of less than 5 meters). (iii) Target categories are highly correlated with their positional distributions in the image (e.g., corrosion is predominantly located near the pipeline bottom, while branches are mainly concentrated at the top). These domain-specific attributes remain largely underexplored in existing research. A systematic investigation and effective integration of such prior knowledge can establish a more robust and explainable foundation for the detection and diagnosis of  MFL-based pipeline features.

\subsection{MFL Data Acquisition and Data Analysis Procedure}

Pipeline Magnetic Flux Leakage (MFL) inspection represents the dominant non-destructive testing technology for long-distance transmission pipelines. The inspection process begins with an in-line inspection robot autonomously traversing the pipeline interior. This device carries high-sensitivity magnetic sensor arrays, acquiring real-time three-axis leakage field signals (axial, radial, circumferential) while the pipe wall is magnetized near saturation. Raw data are stored in frame format, comprising frame headers, MFL/Eddy Current/inductance sensor signals, and motion measurements including odometer readings and accelerometer data.

The typical analysis workflow follows a rigorous engineering chain: after signal acquisition, data quality assessment is performed to eliminate noise segments caused by equipment vibration or abnormal operating speed; subsequent pseudo-color map generation converts signal intensities into intuitive 2D images. The algorithm deployed on server-side software subsequently performs object detection analysis to identify all MFL features and provide their categories and locations; for detected loss-type features, defect quantification further estimates severity (currently relying primarily on traditional feature extraction methods such as signal amplitude and morphological analysis); finally, expert review validates the annotations, and field excavation testing confirms critical defects. This workflow aligns with standard industrial practices and constitutes the authentic scenario underlying our dataset construction.

The AI-related tasks in Pipeline MFL detection can be summarized at two levels: (i) Object Detection: detecting all MFL features, outputting feature categories and precise location information; (ii) Defect Quantification: based on detection results, further estimating the 3D size of loss-type features to provide a quantitative basis for maintenance decisions.

\subsection{Related Works}
Early MFL defect detection~\cite{Ruiz2015PCA, Liu2018PR, Lang2021SVM} relied on traditional pattern recognition approaches and sliding-window matching, which suffered from poor generalization. Currently, deep learning has become the predominant choice for MFL detection owing to its ability to automatically learn discriminative features from raw MFL images~\cite{Huang2023overview}. Table~\ref{tab:related_works} summarizes related MFL object detection research with deep learning. Yang et al.~\cite{Yang2018SSD} first introduced a CNN-based~\cite{CNN} approach using an SSD~\cite{SSD} network, demonstrating the feasibility of casting defect localization as an object-detection task. Subsequent studies~\cite{Xu2023Image, Han2023Single, li2025yolov8} increasingly prioritized real-time performance to keep pace with high-speed in-line inspection. YOLO~\cite{YOLO} family, such as YOLOv5~\cite{yolov5} and YOLOv8~\cite{yolov8_ultralytics}, soon emerged as the mainstream solution, trading two-stage refinement for a lightweight, rapid-forward architecture that can be deployed on edge devices with minimal memory and latency. The most recent innovations aim to recover the precision traditionally sacrificed in single-stage architectures. This has led to the integration of attention mechanisms~\cite{Han2023anchor}, anchor-free prediction heads and lightweight convolutional blocks into YOLO-based frameworks, representing a concerted trend toward models that are both highly accurate and computationally efficient. Beyond architectural improvements, there is also a growing emphasis on advanced learning paradigms. Researchers are increasingly exploring techniques such as semi-supervised learning~\cite{Zhao2025masked} and self-supervised learning~\cite{zhao2023single} to reduce dependency on large-scale, fully annotated datasets. These methods leverage unlabeled or weakly labeled MFL data to improve model generalization, offering promising pathways to address annotation bottlenecks inherent in specialized industrial inspection tasks.

In parallel, datasets for MFL defect detection research have undergone significant quantitative and qualitative expansion. Early collections were typically limited to single-class metal loss localization. In contrast, contemporary datasets have broadened their scope to include auxiliary pipeline components such as welds and flanges, with label vocabularies now often encompassing more distinct categories~\cite{Shen2023multi, Xu2023Image, Han2023anchor, Liu2025Online, Zhao2025masked}. This evolution is accompanied by marked improvements in data quality: spatial resolution has increased, and dataset volumes have grown from a few thousand to several tens of thousands of annotated images.

\subsection{Scenario Challenges}
Despite these significant advances in model architectures and learning paradigms, the transition of deep learning-based MFL analysis from research to robust field application is hindered by several persistent data-centric challenges, as illustrated in Fig.~\ref{fig:teaser}. (i) The field lacks a standardized, public benchmark dataset. Research continues to rely on privately collected data, which hinders reproducible comparisons and slows collective progress. (ii) The high cost of expert annotation severely limits dataset scale; most studies are validated on only thousands of images. This lack of large-scale validation undermines confidence in model robustness and generalization to diverse, real-world pipeline conditions. (iii) The annotation focus remains narrow. Although recent datasets include more categories, critical pipeline structural components (e.g., tees, valves and bends) are often omitted. Accurate detection of these components is essential for spatially mapping detected defects to precise physical locations along the pipeline~\cite{wang2010local}. Their absence can lead to erroneous excavation decisions, resulting in substantial economic losses~\cite{Zhang2024NDT}.

Besides, the MFL detection exhibits several inherent challenges, as shown in Fig.~\ref{fig:teaser}: (i) A severely long-tailed category distribution, where damage and defect objects are often hundreds or even thousands of times more frequent than most pipeline components, which can drive detectors to overfit to these dominant categories and lead to systematically degraded performance on rare yet safety-critical components. (ii) Large inter-class scale variation, as components are often tens of times larger than damage or defect regions, resulting in substantial size disparities that make it difficult for conventional detectors to simultaneously capture small defects and large structural components within a unified representation. (iii) High intra-class feature variability, particularly for the metal loss/corrosion cluster and valves. As corrosion depth progresses, the MFL color response of metal loss gradually evolves from light yellow to dark red, leading to substantial appearance differences between mild and severe corrosion. Furthermore, valves include unidirectional and bidirectional types, whose MFL signatures differ markedly (with unique elliptic or symmetric line). Such intra-class variability can significantly complicate robust feature learning.

\begin{figure*}[!t]
    \centering
    \includegraphics[width=\linewidth]{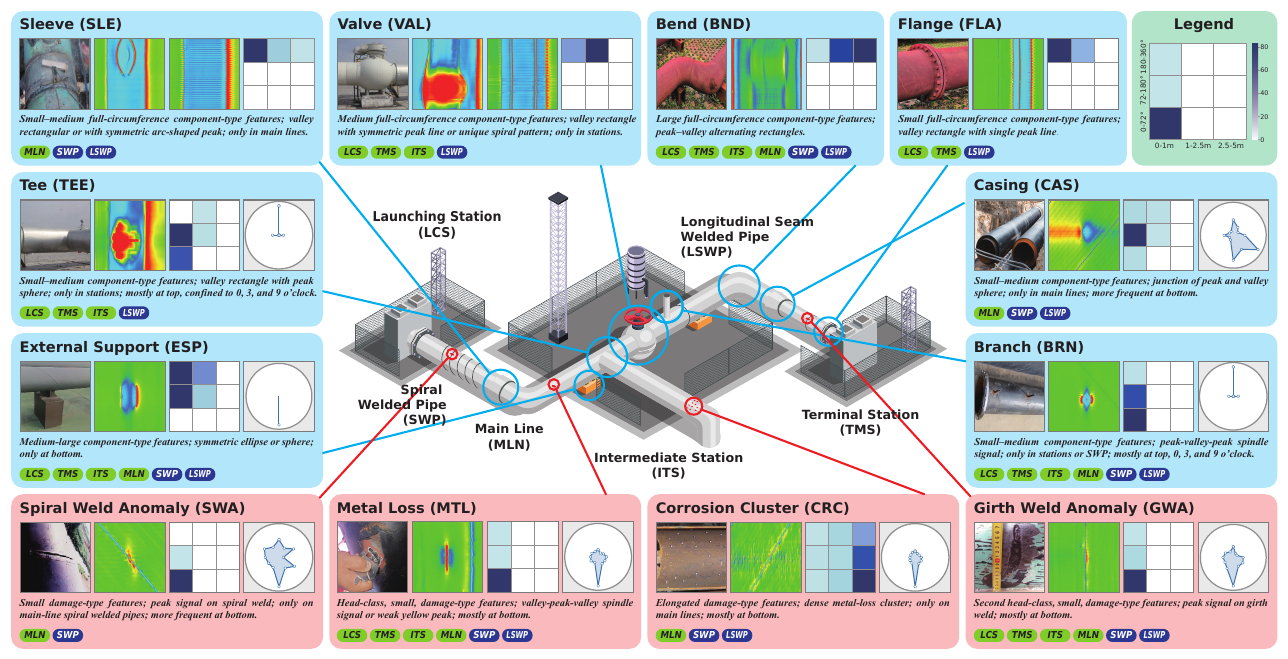}
    \caption{
    Feature taxonomy and annotation characteristics of the PipeMFL-240K dataset. The figure illustrates the pipeline topology and the complete label space used in the dataset. The central schematic provides the spatial context of in-line inspection, including launching, intermediate and terminal stations (LCS, ITS and TMS) and main line (MLN), as well as different pipe types such as spiral welded pipe (SWP) and longitudinal seam welded pipe (LSWP). Surrounding panels present the annotated feature taxonomy.
    For each category, representative optical images and MFL signal maps are shown together with schematic circumferential signal patterns and occurrence distributions. The colored tags indicate the inspection contexts in which each category appears, highlighting the multi-context nature of the labels.
    }
\label{fig:overview}
\end{figure*}

\subsection{Our Contributions}
This work introduces PipeMFL-240K, a large-scale and multi-class object detection dataset of pipeline MFL images. It comprises 249,320 pseudo-color MFL images collected from 12 pipelines spanning approximately 1,530 km in total, with 200,020 labeled objects across 12 critical categories, including defects and key structural components. PipeMFL-240K establishes a much-needed benchmark for the MFL inspection community, providing a common ground for evaluating and comparing state-of-the-art object detection models. We expect this dataset to become a foundational resource for the advancement of industrial AI and domain-specific object detection research. Our key contributions are threefold:
\begin{itemize}
\item \textbf{The first large-scale and multi-class object detection dataset of pipeline MFL imaging.} PipeMFL-240K is the largest publicly available dataset for pipeline MFL-based object detection, comprising 249,320 MFL images and 200,020 labeled objects. Its scale and real-world origin ensure that model training is conducted with sufficient variety and practical relevance, which can advance the feasibility of data-driven inspection systems.
\item \textbf{Fine-grained, multi-class annotations with detailed statistics.}
PipeMFL-240K provides the most comprehensive set of object categories for MFL-based detection to date, with 12 categories covering damage types and structural components. Accurate detection of these components offers crucial spatial references, thereby facilitating the precise localization of damage or defects in the physical space along the pipeline.
\item \textbf{Comprehensive Benchmarking and Reproducible Experimental Framework.} 
We present a benchmark for multi-class object detection in pipeline MFL images, enabling standardized evaluation and comparison of algorithms for pipeline MFL detection.
\end{itemize}

\begin{figure}[!t]
  \includegraphics[width=\linewidth]{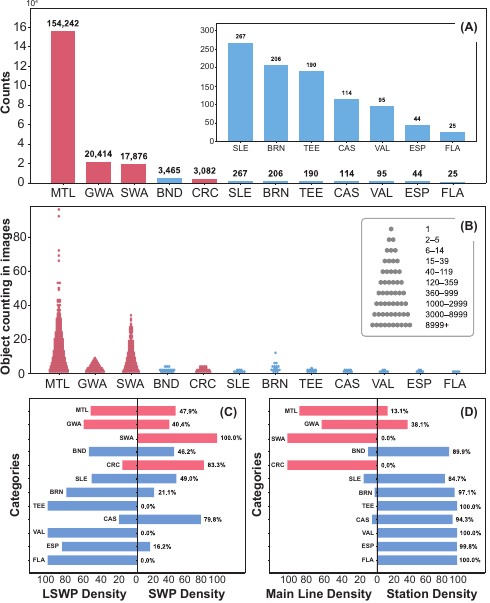}
  \caption{(A) Overall object counts for each annotated category, showing a highly long-tailed distribution across damage-type features (MTL, GWA, SWA, CRC) and component-type features (BND, SLE, BRN, TEE, CAS, VAL, ESP, FLA). (B) Distribution of object counting in images for each category, illustrating strong density imbalance, where a small subset of images contains a disproportionately large number of objects, particularly for metal loss and weld-related anomalies. (C) Relative density ratios of objects observed on LSWP versus SWP, revealing category-specific structural bias in pipe type. (D) Relative density ratios of objects appearing on main lines versus stations, highlighting pronounced contextual imbalance across inspection locations. Together, these statistics characterize the dataset as highly imbalanced in terms of category frequency, object density and spatial context, posing significant challenges for learning robust and generalizable MFL inspection models.}
  \label{fig:statistics}
\end{figure}

\section{Overview of Dataset}
\label{sec:dataset}


      
      

\subsection{Image and Annotation}
The dataset consists of 249,320 pseudo-color images from 12 different pipelines. 
The dataset is organized using the Portable Network Graphics (PNG) format. The detailed data pre-processing procedure is summarized in Section~\ref{sec:prepro}.

Annotation is performed by a dedicated team consisting of six data analysis experts (4 junior, 2 senior). Each pipeline is divided into 1 km segments with rotating junior annotators to minimize bias. 6 experts establish unified standards via joint discussion after a warm-up annotation phase. When junior annotators finish, senior experts review the annotations together to ensure the annotations are correct. This expert team has annotation experience for more than ten years. According to over 5,000 reports of real excavation testing statistics, the precision of the annotation exceeds 98\%, which ensures that the annotations are both accurate and relevant for real-world engineering applications. The precision is not an intra-annotator k-value but the post-excavation confirmation rate obtained from the field reports of the pipeline operator. This engineering-based validation ensures that the bounding boxes are aligned with real-world defect confirmations rather than with purely academic consensus metrics.

The annotations consist of precise bounding-boxes for 4 types of damage and 8 types of components, including Metal Loss (MTL), Corrosion Cluster (CRC), Girth Weld Anomaly (GWA), Spiral Weld Anomaly (SWA), Bend (BND), Sleeve (SLE), Branch (BRN), Tee (TEE), Casing (CAS), Valve (VAL), External Support (ESP) and Flange (FLA). A more detailed problem formulation, along with category-wise example detections, is provided in Section~\ref{sec:problem}.

Fig.~\ref{fig:overview} presents an overview of the PipeMFL-240K dataset, highlighting the spatial distribution of various component types and damage categories along the pipeline. This global layout provides intuitive insight into where different structural elements and defect patterns are most likely to occur, thereby revealing the strong coupling between inspection context and object appearance.

PipeMFL-240K exhibits highly heterogeneous signal characteristics across categories. Component-related categories typically generate stable and repetitive MFL responses that are closely aligned with specific structural configurations. In contrast, damage-related anomalies tend to manifest as localized, low-contrast perturbations embedded within dominant background signals. Moreover, several damage categories share visually similar MFL signatures, which substantially increases inter-class ambiguity and renders fine-grained discrimination particularly challenging.

In addition, PipeMFL-240K demonstrates pronounced multi-context variability. While some categories consistently appear across diverse inspection positions and pipe segments, others are strongly constrained by local structural conditions. This interaction between category semantics and spatial context induces a significant distributional imbalance and contributes to the long-tailed nature of the dataset. Consequently, effective learning on PipeMFL-240K necessitates models capable of jointly capturing signal appearance, structural priors, and spatial dependencies, rather than relying solely on isolated local features.

\begin{table*}[!t]
\centering
\caption{Object distribution across training, validation and testing sets of PipeMFL-240K and Zero-shot Set.}
\label{tab:data_set}
\begin{threeparttable}
\setlength{\tabcolsep}{1mm}{
\begin{tabular}{p{2cm}p{1cm}<{\centering}p{1cm}<{\centering}p{1cm}<{\centering}p{1cm}<{\centering}p{1cm}<{\centering}p{1cm}<{\centering}p{1cm}<{\centering}p{1cm}<{\centering}p{1cm}<{\centering}p{1cm}<{\centering}p{1cm}<{\centering}p{1cm}<{\centering}p{1cm}<{\centering}}
\toprule
& \multicolumn{4}{c}{\textbf{Damage}} & \multicolumn{8}{c}{\textbf{Component}} & \\
\cmidrule(lr){2-5} \cmidrule(lr){6-13}
\textbf{Dataset} & \textbf{MTL} & \textbf{CRC} & \textbf{GWA} & \textbf{SWA} & \textbf{BND} & \textbf{SLE} & \textbf{BRN} & \textbf{TEE} & \textbf{CAS} & \textbf{VAL} & \textbf{ESP} & \textbf{FLA} & \textbf{Total}\\
\midrule
Training Set & 87,617 & 1,630 & 12,191 & 10,244 & 2,049 & 154 & 99 & 113 & 60 & 48 & 25 & 13 & \textbf{114,243} \\
Validation Set & 29,575 & 566 & 4,032 & 3,482 & 675 & 54 & 50 & 31 & 28 & 20 & 9 & 4 & \textbf{38,526} \\
Testing Set & 29,720 & 573 & 4,135 & 3,423 & 708 & 58 & 45 & 38 & 25 & 23 & 9 & 4 & \textbf{38,761} \\
Zero-shot Set & 7,330 & 313 & 56 & 727 & 33 & 1 & 12 & 8 & 1 & 4 & 1 & 4 & \textbf{8,490}\\
\midrule
\textbf{Total} & \textbf{154,242} & \textbf{3,082} & \textbf{20,414} & \textbf{17,876} & \textbf{3,465} & \textbf{267} & \textbf{206} & \textbf{190} & \textbf{114} & \textbf{95} & \textbf{44} & \textbf{25} & \textbf{200,020} \\
\bottomrule
\end{tabular}}
\end{threeparttable}
\end{table*}

\subsection{Statistics of PipeMFL-240K}
We present a statistical analysis of the PipeMFL-240K dataset to characterize its category distribution, object density and spatial contextual bias. Specifically, we analyze circumferential signal patterns and size distributions (Fig.~\ref{fig:overview}) and examine category-level object counts, per-image object density, pipe-type distribution and category-scenario distribution (Fig.~\ref{fig:statistics}). Additionally, we examine the relationship between pipeline attributes and category objects, as detailed in Section~\ref{sec:pipe}.

Circumferential signal patterns (Fig.~\ref{fig:overview}) indicate pronounced spatial contextual bias. Full-circumference component-type categories (SLE, VAL, BND, FLA) exhibit complete circumferential coverage, whereas several other component categories show clear azimuthal preferences (TEE and BRN are biased toward the top and concentrated near 0, 3 and 9 o'clock; CAS occurs only at 6 o'clock). Damage-type categories (SWA, CRC, GWA, MTL) are predominantly observed at 6 o'clock, corresponding to the bottom of the pipeline. In horizontal/stratified flow, free water preferentially wets and accumulates at the pipe invert, providing a persistent electrolyte and promoting the accumulation of solids and deposits~\cite{paolinelli2018study}; under-deposit corrosion in such low areas is highly localized and accelerates metal loss~\cite{brown2017under}. Weld-related anomalies can be exacerbated by preferential weld corrosion mechanisms, particularly where welds intersect the bottom wetting zone~\cite{lee2005preferential}.

Fig.~\ref{fig:overview} also provides category-wise size distributions, summarized on a 2D grid of axial length and circumferential coverage. Component-type categories are generally larger, whereas damage-type categories (MTL, GWA, SWA) are typically small and localized. CRC is an exception: it can be viewed as a spatially clustered MTL, representing regions where corrosion is dense and recurrent, and thus tends to manifest itself as elongated areas of comparatively large extent. This extreme inter-class scale disparity substantially increases the difficulty of detection, as models must simultaneously handle very large structural components and tiny damage signals within the same inspection imagery.

As shown in Fig.~\ref{fig:statistics}~(A), the dataset exhibits a highly long-tailed category distribution. Among damage-type anomalies, MTL dominates the dataset by a large margin, followed by GWA and SWA, while CRC is comparatively rare. Component-type features show substantially lower object counts overall, with BND, SLE and BRN being the most frequent and FLA and ESP appearing only sparsely. This pronounced imbalance reflects real-world inspection conditions, where certain types of defects occur much more frequently than others.

Fig.~\ref{fig:statistics}~(B) further analyzes the number of objects per image for each category. The results reveal a severe object-density imbalance, particularly for MTL and weld-related anomalies, where a small subset of images contains a very large number of objects. In contrast, most component-type features appear only a few times per image. This heavy-tailed density pattern indicates that learning algorithms must cope not only with category imbalance but also with large variations in object concentration across samples.

Spatial and structural biases are analyzed in Fig.~\ref{fig:statistics}~(C) and Fig.~\ref{fig:statistics}~(D). Fig.~\ref{fig:statistics}~(C) compares the relative density of objects on longitudinal seam welded pipes (LSWP) versus spiral welded pipes (SWP). Several categories exhibit strong pipe-type preference, indicating that their occurrence is tightly coupled with specific manufacturing structures. Fig.~\ref{fig:statistics}~(D) shows the distribution of objects between main lines and inspection stations. Component-type features are predominantly observed at stations, whereas damage-type anomalies are concentrated on main lines. This clear contextual separation highlights the multi-context nature of the dataset and the strong correlation between object categories and inspection locations.

\section{Experiments and Benchmarks}

\begin{table*}[!t]
\centering
\caption{Model comparison on PipeMFL-240K Testing Set (Confidence score=0.25, IoU=0.5). The best results in each column are highlighted in bold and the second-best values are underlined.}
\label{tab:result_full_overall}
\begin{threeparttable}
\setlength{\tabcolsep}{0.7mm}{  
\begin{tabular}{p{3.2cm}@{\centering} 
               p{1.2cm}<{\centering}
               p{1.5cm}<{\centering}
               p{1.5cm}<{\centering}
               p{2.4cm}<{\centering}
               p{2.4cm}<{\centering}
               p{2.4cm}<{\centering}
               p{1cm}<{\centering}
               p{1cm}<{\centering}
}
\toprule
\textbf{Model} & \textbf{Year} & \textbf{mAP50} & \textbf{mAP50:95} & \textbf{Precision} & \textbf{Recall} & \textbf{F1-score} & \textbf{Param} & \textbf{FLOPs}  \\
 &  &  &  & \textbf{(macro, micro)} & \textbf{(macro, micro)} & \textbf{(macro, micro)} & \textbf{[M]} & \textbf{[G]}  \\
\midrule 
Faster R-CNN~\cite{2015FASTRCNN} & 2017 & 0.106 & 0.060 & 0.065, 0.070 &0.189, 0.065  & 0.091, 0.068 & 41.4 & 535.9   \\
RetinaNet~\cite{retina} & 2017 & 0.173 & 0.086 & 0.111, 0.055 & 0.356, \underline{0.467} & 0.157, 0.098 & 32.4 &  519.8  \\
YOLOv5-s~\cite{yolov5}     & 2020 &0.426  &0.265  &0.453, 0.377  &0.564, 0.333  &0.480, 0.354  & 9.1  &23.9    \\
YOLOv5-m~\cite{yolov5}   & 2020 & 0.452 & 0.277 & 0.446, 0.388 & 0.577, 0.345 & 0.477, 0.365 & 25.1 & 64.0   \\
YOLOv5-l~\cite{yolov5}   & 2020 & \textbf{0.498} & 0.285 & 0.493,  0.377 & \underline{0.622}, 0.410 & 0.538, \underline{0.393} & 53.1 & 134.7  \\
YOLOv8-s~\cite{yolov8_ultralytics}   & 2023 & 0.402 & 0.248 &0.413, 0.361 & 0.544, 0.369& 0.449, 0.365 & 11.1&28.5    \\
YOLOv8-m~\cite{yolov8_ultralytics}   & 2023 &0.450  &0.288  &0.448, 0.375  &0.596, 0.356  &0.490, 0.365  &25.9  & 78.7    \\
YOLOv8-l~\cite{yolov8_ultralytics}   & 2023 & 0.475&0.298 &0.529, 0.410 & 0.587, 0.368 &\underline{0.541}, 0.388 & 43.6& 164.9    \\
YOLO11-s~\cite{yolo11}   & 2024 &0.353 &0.202 & 0.453, 0.376 & 0.485, 0.351 & 0.437, 0.363 & 9.4 & 21.3  \\
YOLO11-m~\cite{yolo11}   & 2024 & 0.406 & 0.251 & 0.511, 0.467 & 0.515, 0.297 & 0.488, 0.363 & 20.0  &  67.7  \\
YOLO11-l~\cite{yolo11}   & 2024 & 0.449 & 0.270& 0.583, \underline{0.469} & 0.544, 0.273 & 0.529, 0.345 & 25.3  &  86.6 \\
YOLOv8-s-world-v2~\cite{2024yoloworld}   & 2024 & 0.467 & 0.297 & 0.444, 0.365&0.616, 0.375 & 0.493, 0.370 & 12.7 & 34.7   \\
YOLOv8-m-world-v2~\cite{2024yoloworld}   & 2024 & 0.494  &\textbf{0.327}  &0.533, 0.462  &0.604, 0.298  &0.535, 0.362  & 28.4 & 88.8   \\
YOLOv8-l-world-v2~\cite{2024yoloworld}   & 2024 & 0.469 & \underline{0.302} & 0.507, 0.416 & 0.585, 0.386& 0.530, \textbf{0.401} & 46.8 & 179.0  \\
YOLOv8-x-world-v2~\cite{2024yoloworld}   & 2024 & \textbf{0.498} & 0.300 & \underline{0.613}, \textbf{0.504} & 0.585, 0.282& \textbf{0.571}, 0.362 & 72.9 &  277.4 \\
RT-DETR~\cite{2024RTDETR}  & 2024 & 0.174 & 0.074 & 0.253, 0.151 & 0.268, 0.400 & 0.238, 0.219 & 32.0 & 103.5   \\
RF-DETR-Base~\cite{rfdetr} & 2025 & 0.472 & 0.273 & 0.378, 0.151 & \textbf{0.650}, \textbf{0.514} & 0.460, 0.233 & 28.6 & 89.3 \\
YOLO26-n~\cite{yolo26}   & 2026 & 0.317 & 0.223 & 0.522, 0.394 & 0.397, 0.301 & 0.388, 0.342 & 2.4& 9.5  \\
YOLO26-s~\cite{yolo26}   & 2026 & 0.432 & 0.251 & 0.474, 0.389 & 0.549, 0.312 & 0.472, 0.346 & 9.5 & 22.5  \\
YOLO26-m~\cite{yolo26}   & 2026 & 0.442 & 0.263 & 0.585, 0.436 & 0.550, 0.291 & 0.536, 0.349 & 20.4 & 67.9  \\
YOLO26-l~\cite{yolo26}   & 2026 & 0.363 & 0.210 & \textbf{0.616}, 0.442 & 0.449, 0.279 & 0.475, 0.342 & 24.8 & 86.1 \\
YOLO26-x~\cite{yolo26}   & 2026 & 0.453 & 0.290 & 0.487, 0.398& 0.575, 0.364 & 0.504, 0.380 & 55.6 & 193.4  \\
\bottomrule
\end{tabular}}
\end{threeparttable}
\end{table*}

\begin{figure*}[!t]
    \centering
    \includegraphics[width=\linewidth]{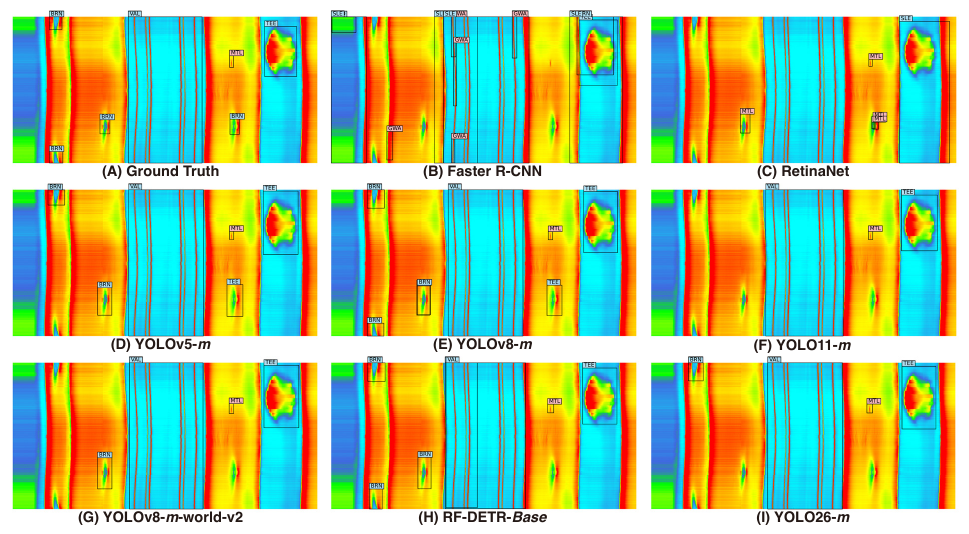}
    \caption{Qualitative benchmark results on representative MFL samples. Predicted bounding boxes from different detectors are compared under the same evaluation setting.}
    \label{fig:visual_results}
\end{figure*}

\subsection{Experimental Setup}

Pseudo-color MFL images in PipeMFL-240K represent a full circumferential unwrap in the vertical direction and an axial scan in the horizontal direction, meaning the inspection data are inherently continuous along the pipe axis, and the axial extent is effectively unbounded compared with the fixed circumference. To reflect this deployment reality, a unified patch-based pipeline is adopted in which both training and inference operate on full-circumference square slices: each slice always covers the entire full circumferential context while moving along the axial direction. This design serves two goals: (i) it preserves geometry-dependent patterns of structural components that require complete circumferential context and (ii) it enables an evaluation that explicitly tests a detector’s robustness and localization stability on continuous axial imagery, where defects can be sparse, tiny, and sensitive to boundary or context changes. The following paragraphs provide the exact training-time cropping, sampling strategy, and the sliding-window inference, merging procedure.


During training, two 2400 $\times$ 2400 windows with horizontal coordinates corresponding to 100-2500 and 2500-4900, are cropped from each 5000 $\times$ 2400 images and each window is then resized to 640 $\times$ 640 to match the detector input. To ensure stable training, we maintain a foreground-to-background sampling ratio of $9{:}1$~\cite{yolo11}, while the validation and testing datasets remain unchanged. A slice is defined as a foreground sample if the patch center lies inside the bounding box of any annotated object; otherwise, it is treated as a background sample. In the training set, 53,375 patch slices contain objects and all of them are included in the training process. To make full use of the entire dataset, background patches are randomly sampled with sampling ratio $9{:}1$, resulting in 5,930 background samples. 

During inference, Slicing Aided Hyper Inference~\cite{akyon2022sahi} is applied with a sliding window of 2400 $\times$ 2400 and a stride of 1300 px (45.8\% overlap). Each slice is resized to 640 $\times$ 640 before being fed to the network; the resulting detections are mapped back to the original slice coordinates and merged across all windows via standard NMS~\cite{NMS} (IoU~\cite{IoU} $\geq$ 0.5). Only boxes whose confidence exceeds 0.25~\cite{yolo11} are retained.



\subsection{Implementation Setup}
All experiments are performed on a server with 4 $\times$ RTX 4090 (24 GB), CPU Xeon Platinum 8470Q and 360 GB RAM. 
The dataset contains 240,320 pseudo-color MFL images (excluding Zero-shot Set), which is split into training/validation/testing sets with a ratio of 6:2:2. The dataset is split on a per-pipeline basis, instead of globally shuffling. Within each pipeline, samples are allocated by category frequency (from tail to head) following a 6:2:2 ratio. After positive sample allocation, pure background images are supplemented to achieve 144,192/48,064/48,064 splits with approximately 3:7 positive-negative ratio, reflecting realistic MFL characteristics. The object distribution is kept as balanced as possible in these splits, as summarized in Table~\ref{tab:data_set}. 
Each model is trained for 300 epochs with early-stopping of 20 epochs and the training process takes approximately one day per model. To guarantee fair comparison, the original hyperparameters released by each baseline repository are adopted without manual re-tuning; the only changes are dataset-specific input dimensions and batch size adjusted to fit GPU memory.

\subsection{Benchmark Results}
To evaluate MFL detection performance, comprehensive experiments are conducted on the PipeMFL-240K dataset, benchmarking a series of widely-used supervised detection architectures. The supervised models encompass: (i) Two-stage detectors: Faster R-CNN~\cite{2015FASTRCNN}; (ii) One-stage detectors: RetinaNet ~\cite{retina} and the YOLO series (YOLOV5~\cite{yolov5}, YOLOv8~\cite{yolov8_ultralytics}, YOLO11~\cite{yolo11}, YOLO-World~\cite{2024yoloworld} and YOLO26~\cite{yolo26}); (iii) End-to-end transformers~\cite{attention}: DETR-based\cite{detr} models (RT-DETR~\cite{2024RTDETR} and RF-DETR~\cite{rfdetr}). 

Table~\ref{tab:result_full_overall} presents the benchmark comparison of all methods under the same evaluation setting. Detection accuracy is quantified following standard evaluation protocols using the COCO-style metrics~\cite{coco}: mean Average Precision at IoU=0.5 (mAP50), mean Average Precision at IoU=0.50:0.05:0.95 (mAP50:95), Precision (P), Recall (R) and F1-score with macro and micro settings. Furthermore, model efficiency is assessed in terms of the number of parameters count (Params) and computational complexity (FLOPs). The macro F1-score is the average of each category's F1-score, not calculated by macro P and R. Fig.~\ref{fig:visual_results} visualizes representative detection outputs for qualitative assessment. More detailed quantitative results (including per-category metrics) and additional qualitative visualizations are provided in Section~\ref{sec:detail_exp}.

Among all tested benchmark models, the early one-stage model RetinaNet and the two-stage model Faster R-CNN have much more parameters than other models, but these models perform poorly. YOLO variants achieve superior accuracy-efficiency trade-offs, maintaining stable recall on dominant damage types and structural parts. Within the YOLO family, the YOLO-World series achieves the strongest results. Meanwhile, we notice that: (i) the latest release YOLO26 lags behind YOLOv8 and even YOLOv5, possibly because its optimization revisions, loss redesign and end-to-end NMS-free inference do not align with MFL characteristics, while YOLOv8 and YOLOv5 have been upgraded and deployed in the field for a considerable period. (ii) As model size grows, performance typically improves from small to medium backbones but then saturates or fluctuates: further scaling yields diminishing returns and can even degrade certain metrics. Larger models also tend to adopt a noticeably more conservative decision boundary, consequently suppressing false positives more effectively and thus increasing precision at the expense of recall under a fixed threshold, while still achieving a higher overall F1-score in most cases.

DETR-based models are less competitive. Although RF-DETR approximates YOLO-level accuracy, RT-DETR fails to detect entire cohorts of categories, especially continuous structures such as VAL, SLE and FLA (Table~\ref{tab:result_full_ap50}-\ref{tab:result_full_f1} in Section~\ref{sec:detail_exp}). The RT-DETR decoder lacks explicit global-scale reasoning, crippling the localization of elongated vertical components and dense micro-defects. Moreover, transformer detectors rely on pre-training over semantically balanced natural images; this prior conflicts with the heavily long-tailed MFL distribution, thereby limiting applicability to this task even further.
Grounding-DINO~\cite{dino} is evaluated under the same protocol; however, its results are omitted from Table~\ref{tab:result_full_overall} due to consistently poor performance. Zero-shot foundation models show limited transferability, as pre-training on natural imagery yields representations whose texture and semantic statistics diverge markedly from pseudo-color MFL maps; discriminative ability fails to transfer without domain-specific adaptation.


\subsection{Data Scaling Study}
To investigate the impact of data scale on model performance, we extract several subsets from the training set of PipeMFL-240K at different scales: 1/2, 1/4, 1/8 and 1/16. These subsets preserve approximately the same category distribution as the full training set, while ensuring that each category contains at least one object. Three models with moderate size are trained with these subsets: YOLOv8-m, YOLO26-m and RF-DETR-Base. When the scale is less than 1/16, all models perform poorly (mAP50 < 0.05). 

\begin{figure*}[!t]
  \includegraphics[width=\linewidth]{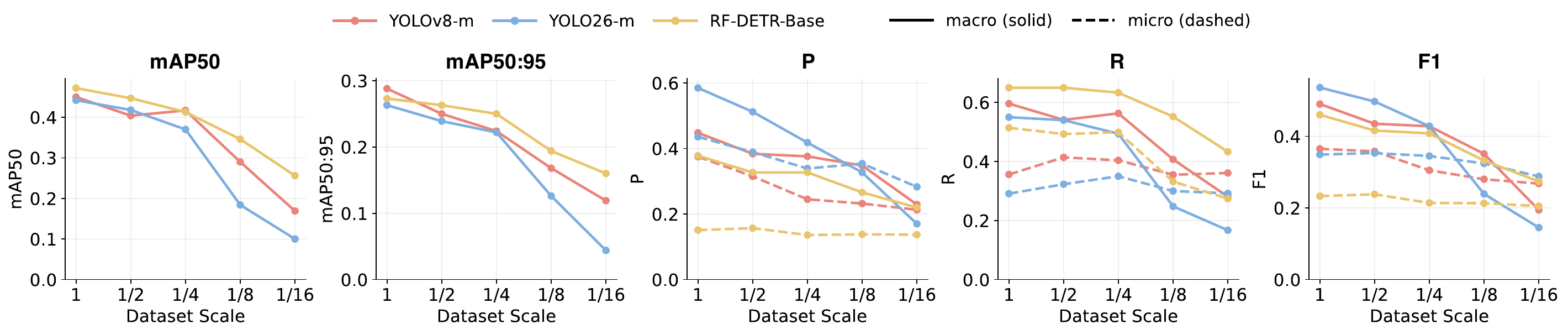}
  \caption{Dataset scale study results on YOLOv8-m, YOLO26-m and RF-DETR-Base, illustrating performance variations in mAP50, mAP50:95, precision, recall and F1-score as the training data decreases.}
  \label{fig:scale plot}
\end{figure*}

Fig.~\ref{fig:scale plot} demonstrates the performance of these models under different dataset scales: (i) mAP50, mAP50:95, Precision, Recall and F1-score decrease when training with less data, indicating that a large-scale dataset is essential to achieve better detection performance. (ii) Among all models, RF-DETR-Base has the best robustness; when trained with only 1/16 of the data, the performance does not drop significantly, showing that RF-DETR-Base can be trained with insufficient samples. The per-category metrics of these models are summarized in Section~\ref{sec:detail_small_data}.

\subsection{Zero-shot Domain Generalization Testing}

To investigate model generalization on completely unseen pipelines, we conduct Zero-shot domain generalization experiments. We select a pipeline (Zero-shot Set, 240K-Z) with 406 mm diameter (not present in PipeMFL-240K), featuring steel material and purity significantly different from the 240K dataset. Additionally, this pipeline is inspected using a newly developed in-line inspection robot with distinct sensor configurations. We maximize the disparities between test and training environments to rigorously evaluate cross-domain robustness.
This pipeline spans approximately 50 km, containing 9,000 images with 8,490 annotated objects across all 12 categories. Table \ref{tab:data_set} compares key statistical characteristics between the Zero-shot Set and the main dataset. The Zero-shot Set exhibits significantly higher object density (0.943 vs. 0.131 obj/m) and greater corrosion defect proportion (MTL: 86.3\% vs. 76.7\%), reflecting more severe degradation in this pipeline. This "same task, different distribution" setting provides a strict test for assessing domain robustness.
We evaluate 22 models trained on PipeMFL-240K on this external pipeline. Table \ref{tab:extra_ap_zero} presents the performance of 22 trained models on Zero-shot Set. Results demonstrate that most models, particularly YOLO variants, maintain a stable generalization capability on the new pipeline, with mAP50 and mAP50:95 comparable to those on PipeMFL-240K Testing Set. Consistent with findings on PipeMFL-240K, YOLO-World series achieves superior performance (mAP50=\textbf{0.516}, mAP50:95=\textbf{0.300}), indicating this family maintains strong performance in pipeline MFL detection under both fully-supervised training and zero-shot transfer scenarios.
Under zero-shot scenarios, most models exhibit a "high recall, low precision" tendency. For instance, RF-DETR-Base achieves macro Recall of 0.482 but Precision of merely 0.198. This characteristic is also observed on the main test set but less severely. This likely stems from lowered discrimination thresholds for background noise under domain shift, causing increased false positives. For tail categories, metric instability becomes particularly pronounced due to extreme scarcity: with merely 1 to 4 instances for FLA, ESP and similar categories, their AP50, Precision and Recall demonstrate binary "all-or-nothing" behavior, authentically reflecting the evaluation dilemma of extremely rare categories in industrial inspection. More detailed quantitative results are provided in~\ref{sec:detail_zero_test}.

\subsection{Sampling Ratio Selection}
To investigate the performance of models trained with different sampling ratios, we conduct experiments with the YOLOv8 series on both the standard testing set and the zero-shot testing set. Tables \ref{tab:sample_240}--\ref{tab:sample_zero_f1} present the quantitative metrics of models trained with various sampling ratios. The results demonstrate that the 9:1 sampling ratio consistently outperforms the alternatives across model scales and testing environments. Notably, the 3:7 "real-world distribution" (closer to natural occurrence frequencies) leads to substantial performance degradation: for YOLOv8-l, mAP50 drops from 0.475 (9:1) to 0.248 (3:7) on PipeMFL-240K, and from 0.411 (9:1) to 0.227 (3:7) on the zero-shot testing set. All 3:7 models converge within 60 epochs due to background gradient dominance, whereas most models trained with the 9:1 ratio continue training stably for 200+ epochs stably. This suggests that, in extreme imbalance scenarios, such as MFL detection, foreground feature learning should be prioritized over background distribution fidelity. The 9:1 ratio ensures sufficient learning signals for sparse features, which proves critical for generalization, as validated by consistent trends across both standard and zero-shot evaluations. More detailed analysis is discussed in Section~\ref{sec:detail_sampling}.

\section{Discussion}
We present a systematic investigation of complex-object detection in pipeline MFL imagery. The benchmark results demonstrate the performance boundary of contemporary detectors on this task and reflect the structural challenges that MFL imaging possesses compared with natural scene images.

\paragraph{\textbf{Challenge Analysis}}
Pipeline MFL detection exhibits clear, task-specific data and signal characteristics. (i) An extremely long-tailed category distribution induces severe imbalance, biasing training toward high-frequency categories and eroding sensitivity to infrequent yet safety-critical categories. (ii) Inter-class scale variance is drastic, head-type anomalies are spatially tiny and have a weak signal response, whereas tail-type features are large but rare, degrading precision. These properties require a unified architecture with exceptional scale generalization. (iii) Inter-class appearance variability is high, complicating stable feature learning. The categories are tightly coupled to pipe geometry and axial position in the vertical direction, suggesting that incorporating structural priors could be beneficial.

\paragraph{\textbf{Baseline Comparison}}
Even with large-scale training data, mainstream detectors struggle to deliver consistent quality across all categories. The dominant performance bottleneck appears to stem from a structural mismatch between prevailing detection paradigms and the characteristics of MFL data. Standard architectures implicitly assume relatively balanced categories and largely independent objects, whereas MFL imagery exhibits dense, position-dependent and severely imbalanced object distributions. Simultaneously localizing massive structural components and tiny defects further stresses the scale generalization of existing backbones. Moreover, purely data-driven feature learning neglects domain knowledge such as circumferential continuity, axial position priors and defect-scene associations, limiting both discrimination and stability.

\section{Conclusion and Future Work}

We present PipeMFL-240K, the first large-scale public dataset dedicated to pipeline MFL inspection, targeting the detection of both pipeline defect and component patterns in real inspection scenarios.
The dataset provides a substantial collection of high-quality annotations, complemented by explicit pipeline type and scene metadata, enabling systematic analysis across diverse inspection conditions. In addition to dataset construction, we conduct a comprehensive benchmark study and statistical analysis, revealing pronounced long-tailed category distributions, extreme scale variation, and high intra-class diversity, which jointly characterize the intrinsic difficulty of MFL-based pipeline inspection. Benchmark results show that even state-of-the-art object detectors, including single-stage, two-stage, and Transformer-based architectures, achieve only limited performance on PipeMFL-240K. Performance degradation is particularly evident for rare categories and small-scale defects, highlighting unresolved challenges in current detection paradigms. These findings indicate that PipeMFL-240K exposes fundamental limitations of existing object detection methods under realistic industrial inspection conditions and provides a demanding evaluation platform for future research on robust detection under long-tailed distributions and severe multi-scale variations.




Future work should focus on improving robustness to category imbalance and scale variation in pipeline MFL inspection, with particular emphasis on rare and small-scale defects. In addition, exploring more efficient detection frameworks is important to support practical deployment, while more flexible labeling and detection paradigms may be considered to accommodate diverse and evolving inspection requirements.

\begin{acks}
This research was supported by National Natural Science Foundation of China (62573206, General Program), Science \& Technology Research Special Fund of China National Machinery Industry Corporation (SINOMACH) and the Open Fund of National Engineering Research Center of Transducer (2025 No.5), China.
\end{acks}

\bibliographystyle{ACM-Reference-Format}
\bibliography{reference}

\clearpage

\appendix


\section{PipeMFL-240K Data Access and Format}

The data can be accessed on Hugging Face at \url{https://huggingface.co/datasets/PipeMFL/PipeMFL-240K}. The dataset has a permanent DOI: \url{https://doi.org/10.57967/hf/7651}. The benchmark and code can be accessed on GitHub at \url{https://github.com/TQSAIS/PipeMFL-240K}. The Zero-shot set 240K-Z is provided in a separate folder, while the folder arrangement remains the same.

The dataset PipeMFL-240K is organized in one main folder. The dataset structure is shown as follows:

\begin{lstlisting}[style=datasetstyle]
PipeMFL-240K/
|-- train/
|   |-- images/              # Contains all train images in PNG format
|       |-- Train_A_0000011.png
|       |-- Train_A_0000047.png
|       |-- ...
|   |-- labels/              # Contains all train labels in TXT format (YOLO)
|       |-- Train_A_0000011.txt
|       |-- Train_A_0000047.txt
|       |-- ...
|   |-- train.coco.json     # Contains all train labels in JSON format (COCO)
|-- val/
|   |-- images/              # Contains all val images in PNG format
|       |-- Val_A_0000051.png
|       |-- Val_A_0000117.png
|       |-- ...
|   |-- labels/              # Contains all val labels in TXT format (YOLO)
|       |-- Val_A_0000051.txt
|       |-- Val_A_0000117.txt
|       |-- ...
|   |-- val.coco.json     # Contains all val labels in JSON format (COCO)
|-- test/
|   |-- images/              # Contains all test images in PNG format
|       |-- Train_A_0000086.png
|       |-- Train_A_0000096.png
|       |-- ...
|   |-- labels/              # Contains all test labels in TXT format (YOLO)
|       |-- Train_A_0000086.txt
|       |-- Train_A_0000096.txt
|       |-- ...
|   |-- test.coco.json     # Contains all train labels in JSON format (COCO)
|-- metadata_image.xlsx       # Image metadata of PipeMFL-240K
|-- metadata_pipe.xlsx       # Pipeline metadata of PipeMFL-240K
|-- data.yaml         # Pre-defined classes for PipeMFL-240K (YOLO)
|-- classes.txt       # Pre-defined classes for PipeMFL-240K (YOLO)

            
\end{lstlisting}

\begin{itemize}
    \item \texttt{PipeMFL-240K/\{split\}/images/}: Contains all train/val/test images in PNG format. Each file is named as: \\ \texttt{[Split]\_[PipelineID]\_[ImageID].png}
    \item \texttt{PipeMFL-240K/\{split\}/labels/}: Contains all the corresponding detection labels in train/val/test set stored in YOLO TXT format (.txt), with filenames matching the corresponding images.
    \item \texttt{PipeMFL-240K/\{split\}/xxx.coco.json}: Contains all the corresponding detection labels in train/val/test set stored in COCO JSON format (.json). The format of entries in JSON file is shown as follows:
    \begin{samepage}
        \begin{lstlisting}[style=datasetstyle]
{
  "categories": [
    {
      "id": 1,
      "name": "MTL",
    },
    {
      "id": 2,
      "name": "TEE",
    },
    {
      "id": 3,
      "name": "BND",
    },
    {
      "id": 4,
      "name": "CRC",
    },
    {
      "id": 5,
      "name": "BRN",
    },
    {
      "id": 6,
      "name": "GWA",
    },
    {
      "id": 7,
      "name": "SWA",
    },
    {
      "id": 8,
      "name": "ESP",
    },
    {
      "id": 9,
      "name": "VAL",
    },
    {
      "id": 10,
      "name": "FLA",
    },
    {
      "id": 11,
      "name": "CAS",
    },
    {
      "id": 12,
      "name": "SLE",
    }
  ],
  "images": [
    {
      "id": 0,
      "file_name": "Train_A_0000011.png",
      "width": 5000,
      "height": 2400
    },
    {
      "id": 1,
      "file_name": "Train_A_0000047.png",
      "width": 5000,
      "height": 2400
    },
    {
      "id": 2,
      "file_name": "Train_A_0000053.png",
      "width": 5000,
      "height": 2400
    },
    // ... other images (continued)
  ],
    "annotations": [
        {
            "id": 1,
            "image_id": 1,
            "category_id": 1,
            "bbox": [
                2911.0,
                330.0012,
                35.0,
                88.99920000000003
            ],
            "area": 3114.972000000001,
            "iscrowd": 0
        },
        {
            "id": 2,
            "image_id": 1,
            "category_id": 1,
            "bbox": [
                1347.0,
                1956.0012,
                31.0,
                70.99919999999997
            ],
            "area": 2200.975199999999,
            "iscrowd": 0
        },
        {
            "id": 3,
            "image_id": 2,
            "category_id": 1,
            "bbox": [
                1716.0,
                1703.0004000000001,
                49.0,
                124.00080000000003
            ],
            "area": 6076.039200000001,
            "iscrowd": 0
        },
    // ... other annotations (continued)
  ],
}

        \end{lstlisting}
    \end{samepage}
    \item \texttt{data.yaml}: Contains pre-defined categories and data path for YOLO training.
    \item \texttt{classes.txt}: Contains pre-defined categories for result visualization with LabelImg for YOLO.
    \item \texttt{metadata\_image.xlsx}: An Excel file containing mapping information (Pipe NO., Scene Type and Pipe Type) for each image.
    \begin{itemize}
        \item \texttt{Image Name}: A string identifier following the pattern \texttt{Split\_PipeNo\_Index.png}, where
            \texttt{Split} denotes the dataset partition (\texttt{Train}, \texttt{Val}, or \texttt{Test}),
            \texttt{PipeNo} indicates the pipeline identifier (\texttt{A}--\texttt{K}),
            and \texttt{Index} is a 7-digit numeric identifier unique within each split and pipeline.
        \item \texttt{Pipe No.}: A categorical identifier indicating the pipeline from which the image is acquired. Valid values are single uppercase letters (e.g., \texttt{240K-A}, \texttt{240K-B}, \texttt{240K-C}, \ldots, \texttt{240K-K}), each corresponding to a predefined pipeline.
        \item \texttt{Scene Type}: A categorical variable specifying the scene category of the MFL image. Valid values include \texttt{ITS}, \texttt{LCS}, \texttt{MLN} and \texttt{TMS}.
        \item \texttt{Pipe Type}: A categorical attribute indicating the pipeline type. Valid values include \texttt{SWP} and \texttt{LSWP}.
    \end{itemize}
    \item \texttt{metadata\_pipe.xlsx}: An Excel file containing information (Length, Diameter, Age, Probe Layout Type) for each pipeline.

    \begin{itemize}
        \item \texttt{Pipe No.}: A unique identifier assigned to each pipeline segment in the dataset. The identifier consists of a dataset prefix \texttt{240K} followed by an alphabetical suffix (e.g., \texttt{240K-A}, \texttt{240K-B}, \texttt{240K-C}, \ldots, \texttt{240K-K}) to distinguish individual pipelines. 
        \item \texttt{Length (km)}: The physical length of each pipeline, measured in kilometers (km). 
        \item \texttt{Diameter (mm)}: The nominal outer diameter of the pipeline, measured in millimeters (mm). 
        \item \texttt{Age (Year)}: The service age of the pipeline, expressed in years since commissioning.
        \item \texttt{Probe Layout Type}: The sensor arrangement used in the MFL inspection detector. This field is a categorical variable indicating the probe layout configuration, with possible values including \texttt{Single Row} and \texttt{Double Row}.
    \end{itemize}
\end{itemize}

\begin{figure}[!t]
  \includegraphics[width=\linewidth]{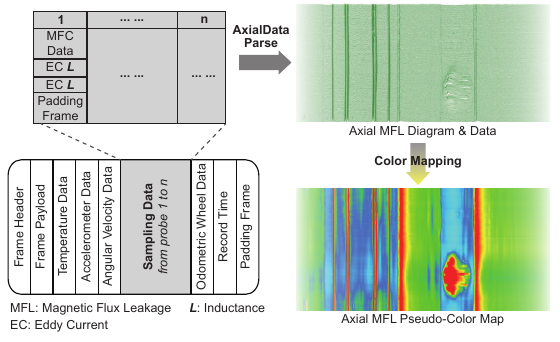}
  \caption{Overview of data collection and acquisition}
  \label{fig:datacollection}
\end{figure}

\section{Detailed Information of PipeMFL-240K}

\subsection{License and Attribution}
The pipeline MFL pseudo-color images and associated annotations (bounding-box annotations of features) in the dataset are licensed under the Creative Commons Attribution Non Commercial 4.0 International License (CC BY-NC 4.0).
For proper attribution when using this dataset in any publications or research outputs, please cite with
the DOI: \url{https://doi.org/10.57967/hf/7651}. The benchmark and code can be accessed on GitHub at \url{https://github.com/TQSAIS/PipeMFL-240K}.

\textbf{\textit{Suggested Citation:}} Qu, T., Yang, S., Wang, H., Song, H., Guo, X., Hu, W., Liu, G., Chen, H. \& Ou, Y. (2026). PipeMFL-240K: A Large-scale Dataset and Benchmark for Object Detection in Pipeline Magnetic Flux Leakage Imaging. \url{https://doi.org/10.57967/hf/7651}

\subsection{Data Pre-Processing}
\label{sec:prepro}
\paragraph{\textbf{Collection and Acquisition}}
As shown in Fig.~\ref{fig:datacollection}, during pipeline MFL inspection, the inspection detector travels along the axial direction of the pipeline while synchronously acquiring multi-channel sensor data. The raw data are organized in a frame-based format, with each frame comprising a frame header, padding information and detection signals from MFL, eddy current (EC) and inductance (L) sensors, together with motion-related measurements such as acceleration, angular velocity and odometer readings.

After acquisition, the raw frame data undergo axial data parsing and are reconstructed into a continuous MFL signal distribution referenced to the axial distance along the pipeline. By integrating odometer information for spatial mapping, an axial MFL data curve is generated. Subsequently, pseudo-color mapping is applied to convert the signal intensity into an intuitive two-dimensional pseudo-color representation, providing a unified data format for subsequent defect identification and analysis. 

\begin{figure}[!t]
  \includegraphics[width=\linewidth]{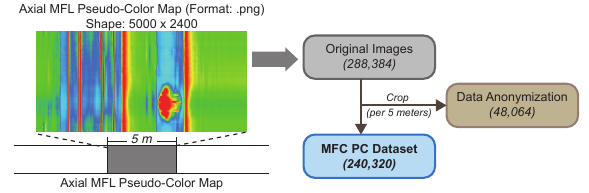}
  \caption{Overview of data selection and filtering.}
  \label{fig:datafilter}
\end{figure}

\paragraph{\textbf{Selection and Filtering}} As illustrated in Fig.~\ref{fig:datafilter}, the complete pipeline pseudo-color images are cropped into fixed-size patches at intervals of 5 m. Each patch has a resolution of 5000 $\times$ 2400 pixels, resulting in a total of 288,384 images (without Zero-shot Set). Subsequently, 20\% of the dataset is randomly removed to achieve data anonymization and conceal location-related information, corresponding to the exclusion of 48,064 images. After data filtering, the remaining 240,320 images, together with the Zero-shot Set, constitute the PipeMFL-240K dataset.

\begin{figure*}[!t]
    \centering
    \includegraphics[width=\linewidth]{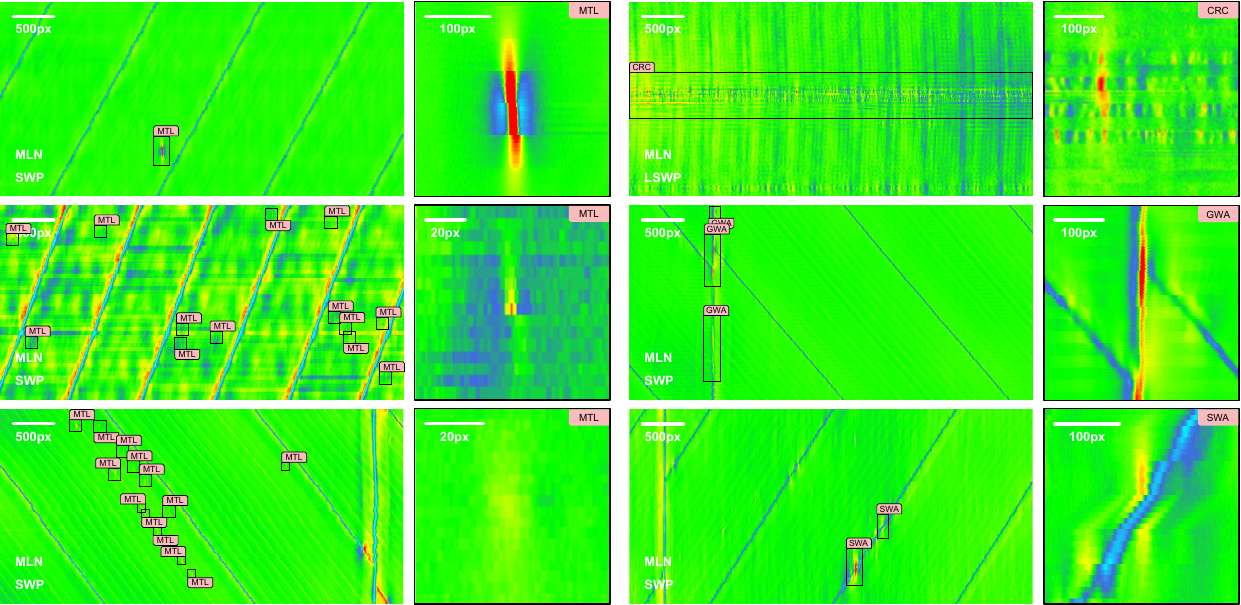}
    \caption{Pattern visualization of damage-type categories in MFL imaging cases: MTL, CRC, GWA and SWA with MLN scene and 2 pipe types (LSWP and SWP).}
\label{fig:damageCase}
\end{figure*}

\begin{figure*}[!t]
    \centering
    \includegraphics[width=\linewidth]{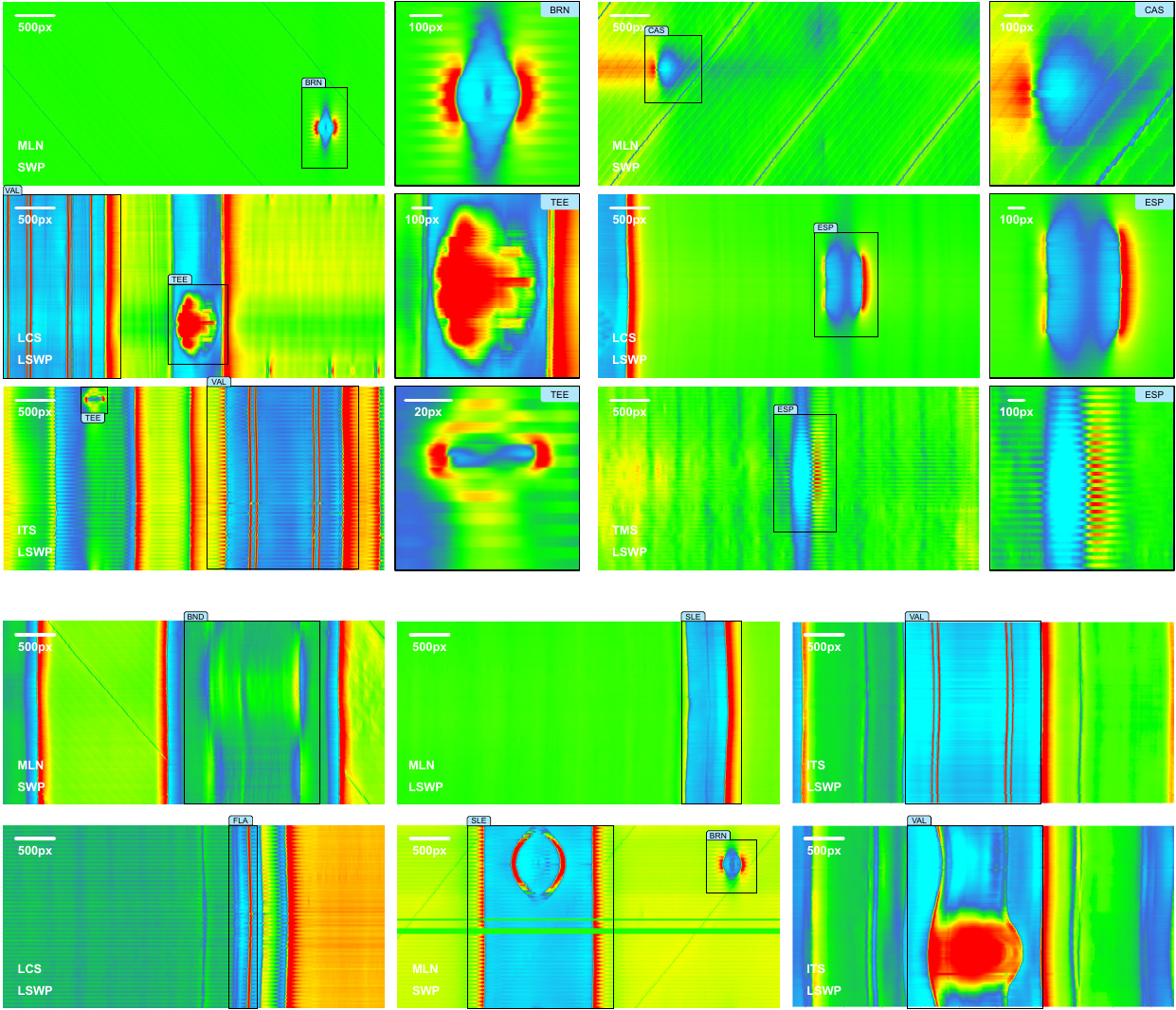}
    \caption{Pattern visualization of component-type categories in MFL imaging cases: BRN, CAS, TEE, ESP, BND, SLE, VAL and FLA  with 4 scenes (LCS, ITS, TMS and MLN) and 2 pipe types (LSWP and SWP).}
\label{fig:componentCase}
\end{figure*}

\subsection{Problem Formulation} 
The PipeMFL-240K consists of 249,320 pipeline magnetic flux leakage pseudo-color images with 200,020 objects of 12 categories. To facilitate further research on the correlation between features and scenes, MFL images are categorized into 4 types of scenes and 2 types of pipelines. 
The definitions of each category, scene and pipe type are as follows:

\paragraph{\textbf{Damage}}
The core mission of this dataset is to detect these damage objects, including 4 types of features: metal loss (MTL), girth weld anomaly (GWA), spiral weld anomaly (SWA) and corrosion cluster (CRC).
Fig.~\ref{fig:damageCase} illustrates the pseudo-color features of 4 damage-type categories. These categories have different distributions in sizes, shapes, colors and patterns.

\textbf{MTL:} the most common type of damage in pipelines, characterized by local wall thinning or pitting. In MFL imaging, it typically appears as a blue–red–blue spindle-shaped structure or a light yellow spot or stripe. The affected regions are relatively small compared to the overall size of the image. MTL is commonly observed around the 6 o’clock direction (near the image center), corresponding to the bottom of the pipeline where corrosion tends to occur. It can appear in all scene types (LCS, TMS, ITS and MLN) and in both pipe types (LSWP and SWP).

\textbf{CRC:} the aggregation of a large number of closely distributed metal loss objects, forming an extensive corrosion region. CRC can only be found in the MLN scene and occurs in both LSWP and SWP pipe types. It is most commonly observed around the 6 o’clock direction (near the image center), corresponding to the bottom of the pipeline.

\textbf{GWA:} the anomaly occurring exclusively at the girth weld, which appears as a straight line spanning the vertical direction in an MFL image and results in local wall thinning at the girth weld. In MFL imaging, GWA is typically indicated by a light red or yellow response along the girth weld. It is most commonly observed around the 6 o’clock direction (near the image center), corresponding to the bottom of the pipeline. GWA can occur in any scene (LCS, TMS, ITS and MLN) and in both types of pipe (LSWP and SWP).

\textbf{SWA:} the anomaly occurring exclusively at the spiral weld, which appears as an inclined straight line spanning the vertical direction in an MFL image and results in local wall thinning at the spiral weld. In MFL imaging, SWA is typically indicated by a light red or yellow response along the spiral weld. It is most commonly observed around the 6 o’clock direction (near the image center), corresponding to the bottom of the pipeline. SWA only exists in the MLN scene with SWP pipe type.


\paragraph{\textbf{Component}}
The core mission of this dataset is to detect these component objects, including 8 types of features: bend (BND), sleeve (SLE), branch (BRN), tee (TEE), casing (CAS), valve (VAL), external support (ESP) and flange (FLA).
Fig.~\ref{fig:componentCase} presents the pseudo-color features of 8 component type categories. These categories also have distinct distributions in sizes, shapes, colors and patterns. In addition, most components exist in the station scene, with pipe type of LSWP, except BND, SLE and BRN.

\textbf{BND:} the pipeline component used to change the flow direction. In MFL imaging, BND typically appears as a very large valley-colored (blue or green) rectangular region spanning the vertical direction of the image, often accompanied by a symmetric peak–valley alternating pattern. BND can be found in any scene (LCS, TMS, ITS and MLN) and in both pipe types (LSWP and SWP).

\textbf{SLE:} the component installed to repair, reinforce, isolate or adapt pipeline sections without replacing the entire line. In MFL imaging, SLE typically appears as a blue rectangular region spanning the vertical direction, with or without a symmetric elliptical pattern inside. SLE can only be found in the MLN scene and occurs in both LSWP and SWP pipe types.

\textbf{BRN:} the pipe component used to divert part of the fluid flow to other pipelines. In MFL imaging, BRN typically appears as a red–blue–red spindle-shaped structure. BRN can be found in any scenes, but only at positions 0, 3 and 9 o’clock.

\textbf{TEE:} the pipeline component used to control or split the flow direction. In MFL imaging, TEE typically appears as a light spherical region embedded within a darker rectangular region spanning the vertical direction of the image. TEE can be found in the LCS, ITS and TMS scenes with the LSWP pipe type and only at positions 0, 3 and 9 o’clock.

\textbf{CAS:} the component installed in a well to seal formation fluids, stabilize the well and maintain well integrity. In MFL imaging, CAS typically appears as a junction of adjacent red and blue spherical patterns. CAS can only be found in the MLN scene and occurs in both LSWP and SWP pipe types.

\textbf{VAL:} the component used to regulate fluid flow by opening, closing, or partially obstructing the pipeline. In MFL imaging, VAL typically appears as either a blue rectangular region with symmetric red lines spanning the vertical direction or as a distinctive spiral elliptical pattern. Both patterns extend across the vertical direction. VAL can only be found in station scenes (LCS, ITS and TMS).

\textbf{ESP:} the component used to prevent excessive settlement or sagging of the pipeline. In MFL imaging, ESP typically appears as a very large, vertically symmetric dark-colored pattern. ESP is only observed at the 6 o’clock direction (near the image center), corresponding to the bottom of the pipeline and can occur in any scene, although it is most commonly found in station scenes.

\textbf{FLA:} the component that enables detachable and sealed connections between pipes, valves and equipment. In MFL imaging, FLA typically appears as a blue rectangular region spanning the vertical direction with a single red line along the vertical axis. FLA can only be found in the ITS and TMS scenes.

\paragraph{\textbf{Scenes}}
To facilitate further research on the correlation between features and scenes, MFL images are categorized into 4 types of scenes: launching station (LCS), intermediate station (ITS), terminal station (TMS) and main line (MLN).

\textbf{LCS:} the launching station, representing the starting point of a pipeline and typically containing various pipeline components, with an approximate length of less than 100 meters. The pipe type in LCS is LSWP.

\textbf{ITS:} the intermediate station located along a long-distance transmission line, typically housing components such as valves and tees, with an approximate length of less than 50 meters. The pipe type in ITS is LSWP.

\textbf{TMS:} the terminal station, representing the end point of a pipeline and serving functions similar to a launching station for receiving operations. The pipe type in TMS is LSWP.

\textbf{MLN:} the main line scene, representing the most frequent operating condition in long-distance pipeline transmission, where damage types are frequently observed. The pipe type in MLN can be SWP or LSWP.


\paragraph{\textbf{Pipeline types}}
To facilitate further research on the correlation between features and scenes, MFL images are categorized into 2 types of pipelines: longitudinal seam welded pipe (LSWP) and spiral welded pipe (SWP). 

\textbf{SWP:} the pipe type containing spiral welds, which appear as inclined straight lines spanning the vertical direction in MFL images. Station-type scenes (LCS, ITS and TMS) do not exist in SWP.

\textbf{LSWP:} the pipe type that contains longitudinal seam welds and no spiral welds. Station-type scenes (LCS, ITS and TMS) exist only in LSWP. In addition, TEE, VAL and FLA are only observed in LSWP.

\label{sec:problem}

\begin{table*}[!t]
\centering
\caption{Object distribution per pipeline in PipeMFL-240K.}
\label{tab:pipeline_distribution}
\begin{threeparttable}
\setlength{\tabcolsep}{0.9mm}{
\begin{tabular}{p{2cm}p{1cm}<{\centering}p{1cm}<{\centering}p{1cm}<{\centering}p{1cm}<{\centering}p{1cm}<{\centering}p{1cm}<{\centering}p{1cm}<{\centering}p{1cm}<{\centering}p{1cm}<{\centering}p{1cm}<{\centering}p{1cm}<{\centering}p{1cm}<{\centering}p{1cm}<{\centering}}
\toprule
& \multicolumn{4}{c}{\textbf{Damage}} & \multicolumn{8}{c}{\textbf{Component}} & \\
\cmidrule(lr){2-5} \cmidrule(lr){6-13}
\textbf{Pipeline ID} & \textbf{MTL} & \textbf{CRC} & \textbf{GWA} & \textbf{SWA} & \textbf{BND} & \textbf{SLE} & \textbf{BRN} & \textbf{TEE} & \textbf{CAS} & \textbf{VAL} & \textbf{ESP} & \textbf{FLA}  & \textbf{Total} \\
\midrule
240K-A & 13,221 & 1,241 & 41 & 563 & 90 & 13 & 45 & 2 & 0 & 4 & 4 & 3 & 15,227 \\
240K-B & 12,452 & 136 & 4,150 & 8,204 & 64 & 60 & 25 & 10 & 13 & 5 & 2 & 2 & 25,123\\
240K-C & 6,300 & 259 & 1,440 & 300 & 82 & 24 & 20 & 10 & 15 & 3 & 2 & 3 & 8,458\\
240K-D & 6,838 & 8 & 22 & 597 & 38 & 0 & 4 & 6 & 2 & 3 & 0 & 2 & 7,520\\
240K-E & 11,257 & 806 & 536 & 1,511 & 312 & 9 & 6 & 14 & 0 & 3 & 1 & 2 & 14,457 \\
240K-F & 6,467 & 0 & 121 & 609 & 252 & 8 & 34 & 5 & 1 & 3 & 5 & 2 & 7,507 \\
240K-G & 6,245 & 49 & 3,170 & 736 & 1,440 & 15 & 12 & 35 & 12 & 8 & 15 & 2 & 11,739\\
240K-H & 21,836 & 136 & 641 & 762 & 332 & 6 & 8 & 22 & 63 & 13 & 2 & 1 & 23,822 \\
240K-I & 21,650 & 87 & 4,888 & 332 & 282 & 1 & 4 & 23 & 3 & 10 & 4 & 1 & 27,285\\
240K-J & 28,625 & 9 & 3,892 & 2,790 & 149 & 123 & 20 & 17 & 2 & 16 & 5 & 2 & 35,650 \\
240K-K & 12,021 & 38 & 1,457 & 745 & 391 & 7 & 16 & 38 & 2 & 23 & 3 & 1 & 14,742\\
240K-Z & 7,330 & 313 & 56 & 727 & 33 & 1 & 12 & 8 & 1 & 4 & 1 & 4 & 8,490\\
\midrule
\textbf{Total} & \textbf{154,242} & \textbf{3,082} & \textbf{20,414} & \textbf{17,876} & \textbf{3,465} & \textbf{267} & \textbf{206} & \textbf{190} & \textbf{114} & \textbf{95} & \textbf{44} & \textbf{25} & \textbf{200,020} \\
\bottomrule
\end{tabular}}
\end{threeparttable}
\end{table*}

\begin{table}[!t]
\centering
\caption{Pipeline characteristics in PipeMFL-240K.}
\label{tab:pipeline_characteristics}
\begin{threeparttable}
\setlength{\tabcolsep}{1.5mm}{
\begin{tabular}{p{1.6cm}p{1.0cm}<{\centering}p{1.3cm}<{\centering}p{1.2cm}<{\centering}p{1.35cm}<{\centering}p{0.85cm}<{\centering}}
\toprule
\textbf{ID} & \textbf{Image} & \textbf{\makecell[c]{Density\\(Obj./m)}} & \textbf{\makecell[c]{Length\\(km)$^*$}} & \textbf{\makecell[c]{Diameter\\(mm)}} & \textbf{\makecell[c]{Age\\(year)}}\\
\midrule
240K-A & 12,304 & 0.218 & 70 & 457 & 20\\
240K-B & 7,910 & 0.502 & 50 & 457 & 17\\
240K-C & 6,871 & 0.211 & 40 & 457 & 17\\
240K-D & 15,831 & 0.084 & 90 & 508 & 1\\
240K-E & 12,640 & 0.180 & 80 & 610 & 16\\
240K-F & 7,900 & 0.150 & 50 & 711 & 12\\
240K-G & 38,690 & 0.049 & 240 & 711 & 8\\
240K-H & 40,370 & 0.088 & 270 & 813 & 9\\
240K-I & 26,614 & 0.182 & 150 & 813 & 9\\
240K-J & 30,800 & 0.210 & 170 & 813 & 9\\
240K-K & 40,390 & 0.055 & 270 & 813 & 9\\
240K-Z& 9,000 & 0.943 & 50 & 406 & -\\
\midrule
\textbf{Total} & \textbf{249,320} & \textbf{0.131} & \textbf{1,530} & \textbf{--} & \textbf{--}\\
\bottomrule
\end{tabular}}
\begin{tablenotes}[flushleft]  
\small  
\item \textbf{*} To ensure data security and confidentiality, the dataset is de-identified: the approximate length of each pipeline is reported and approximately 80\% of the images for each pipeline are randomly retained.
\end{tablenotes}
\end{threeparttable}
\end{table}

\begin{figure}[!t]
  \includegraphics[width=\linewidth]{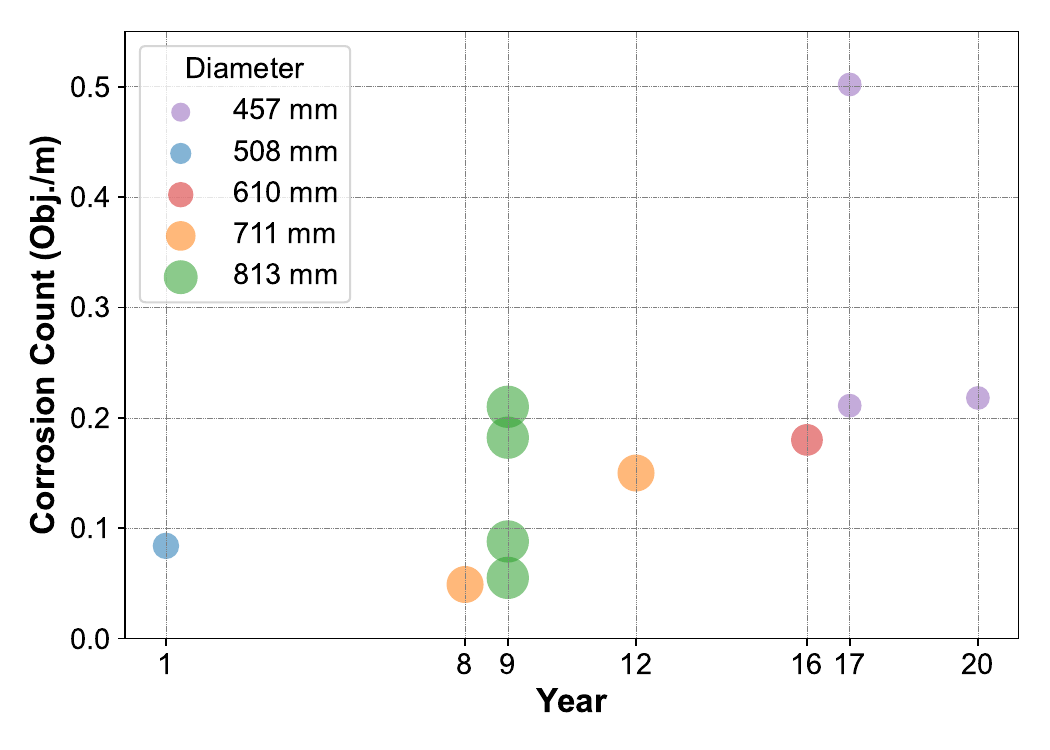}
  \caption{Corrosion density as a function of service age for pipelines with different diameters. Bubble sizes are scaled by diameter, illustrating the variation in corrosion occurrence across pipeline sizes over time.}
  \label{fig:yeardiameter}
\end{figure}

\begin{table*}[!t]
\centering
\caption{Comparison of AP50 for each category in different models on PipeMFL-240K Testing Set. The best results in each column are highlighted in bold and the second-best values are underlined. "-" indicates not detected.}
\label{tab:result_full_ap50}
\begin{threeparttable}
\resizebox{\textwidth}{!}{
\setlength{\tabcolsep}{0.65mm}{
\begin{tabular}{p{3.3cm}p{1cm}<{\centering}p{1cm}<{\centering}p{1cm}<{\centering}p{1cm}<{\centering}p{1cm}<{\centering}p{1cm}<{\centering}p{1cm}<{\centering}p{1cm}<{\centering}p{1cm}<{\centering}p{1cm}<{\centering}p{1cm}<{\centering}p{1cm}<{\centering}p{1cm}<{\centering}p{1cm}<{\centering}}
\toprule
& \multicolumn{4}{c}{\textbf{Damage}} & \multicolumn{8}{c}{\textbf{Component}} \\
\cmidrule(lr){2-5} \cmidrule(lr){6-13}
\textbf{Model} & \textbf{MTL} & \textbf{CRC} & \textbf{GWA} & \textbf{SWA} & \textbf{BND} & \textbf{SLE} & \textbf{BRN} & \textbf{TEE} & \textbf{CAS} & \textbf{VAL} & \textbf{ESP} & \textbf{FLA} & \textbf{Overall}  \\
\midrule 
Faster R-CNN~\cite{2015FASTRCNN} & - & 0.023 & 0.028 & 0.104 & 0.296 & 0.747 & - & - & - & - & - & - & 0.106 \\
RetinaNet~\cite{retina} & 0.168 & 0.015 & 0.008 & 0.037 & 0.329 & 0.754 & -  & 0.075 & 0.052 & 0.425 & 0.206 & - & 0.173\\
YOLOv5-s~\cite{yolov5}& 0.221 &0.131 &0.041 &0.081 &0.541 &0.831 &0.351 &0.701 &0.411 &0.851 &0.621 &0.333 &0.426 \\
YOLOv5-m~\cite{yolov5}  & 0.243 & 0.116 & 0.045 & 0.082 & 0.555 & 0.851 & 0.324 & 0.695 & 0.453 & 0.880 & 0.684 & 0.500 & 0.452 \\
YOLOv5-l~\cite{yolov5} & \textbf{0.286} & 0.117 & \underline{0.072} & \underline{0.118} & 0.548 & 0.838 & 0.401 & 0.747 & 0.524 & 0.835 & \underline{0.737} & \textbf{0.750} & \textbf{0.498}\\
YOLOv8-s~\cite{yolov8_ultralytics} &0.252 &0.091 &0.043 &0.085 &0.514 &0.863 &0.352 &0.485 &0.420 &0.816 &0.569 &0.333 &0.402\\
YOLOv8-m~\cite{yolov8_ultralytics} &0.243 &0.113 &0.051 &0.090 &0.571 &0.871 &0.362 &0.771 &0.531 &\underline{0.891} &0.651 &0.250 &0.450\\
YOLOv8-l~\cite{yolov8_ultralytics} & 0.260&0.105 &0.065 &0.108 &\textbf{0.587} & 0.854&0.279 &0.828 & 0.460&0.875 &0.630 & 0.650&0.475\\
YOLO11-s~\cite{yolo11} &0.239 & 0.079& 0.034& 0.084& 0.451& 0.824& 0.165&0.624 &0.392 &0.812 &0.400 & 0.125&0.353\\
YOLO11-m~\cite{yolo11} & 0.227 & 0.082 & 0.036 & 0.073 & 0.451 & 0.832 & 0.290 & 0.676 & 0.381 & 0.829 & 0.578 & 0.417 & 0.406\\
YOLO11-l~\cite{yolo11} & 0.207&0.107 &0.026 &0.064 & 0.450& 0.861& 0.344&0.685 & 0.325& 0.863& 0.704& \textbf{0.750}& 0.449\\
YOLOv8-s-world-v2~\cite{2024yoloworld} & 0.256 & 0.102 & 0.054 & 0.099 & 0.568 & 0.838 & 0.408 & 0.830 &  0.547 & 0.857 & 0.547 & 0.510 & 0.467 \\
YOLOv8-m-world-v2~\cite{2024yoloworld} & 0.222 & 0.114 & 0.044 & 0.083 & 0.560 & 0.868 & 0.361 & 0.732 &  \textbf{0.627} & 0.881 & 0.690 & \textbf{0.750} & 0.494 \\
YOLOv8-l-world-v2~\cite{2024yoloworld} & \underline{0.273} & 0.132 & \textbf{0.076} & \textbf{0.131} & 0.559 & 0.848 & 0.394 & 0.825 &  \underline{0.595} & \textbf{0.919} & \textbf{0.750} & 0.125 & 0.469 \\
YOLOv8-x-world-v2~\cite{2024yoloworld} & 0.216 & 0.103 & 0.048 & 0.073 & \underline{0.577} & \underline{0.878} & \underline{0.435} & 0.773 &  0.534 & 0.878 & 0.711 & \textbf{0.750} & \textbf{0.498} \\
RT-DETR~\cite{2024RTDETR} & 0.202 & 0.020 & 0.035 & 0.075 & 0.361 & - & 0.374 & \textbf{0.894} & 0.122 & - & - & - & 0.174\\
RF-DETR-Base~\cite{rfdetr} & 0.246 & 0.128 & 0.070 & 0.108 & 0.501 & 0.810 & \textbf{0.479} & \underline{0.878} & 0.365 &0.849  & 0.477 &  \textbf{0.750}& 0.472\\ 
YOLO26-n~\cite{yolo26}  & 0.202 & 0.093 & 0.019 & 0.064 & 0.489 & 0.840 & 0.308 & 0.208 & 0.240 & 0.844 & - & 0.500 & 0.317\\ 
YOLO26-s~\cite{yolo26}   & 0.212 &0.136  & 0.027 & 0.079 & 0.494 & 0.861 & 0.405 & 0.684 & 0.460 & 0.840 & 0.333 & 0.650 & 0.432\\ 
YOLO26-m~\cite{yolo26}   & 0.206 & \textbf{0.149} & 0.032 & 0.088 & 0.430 & 0.829 & 0.346 &0.775  & 0.452 & 0.864 & 0.386 & \textbf{0.750} & 0.442\\ 
YOLO26-l~\cite{yolo26} & 0.200 & 0.120 & 0.024 & 0.065 & 0.445 & 0.836 & 0.300 & 0.519 & 0.360 & 0.759 & 0.222 & 0.500 & 0.363\\ 
YOLO26-x~\cite{yolo26}   & 0.256 &  \underline{0.139}&  0.044& 0.108 & 0.567 & \textbf{0.880} & 0.399 & 0.599 & 0.444 & 0.833 & 0.477 & 0.688 & 0.453\\  
\bottomrule
\end{tabular}}}
\end{threeparttable}
\end{table*}

\begin{table*}[!t]
\centering
\caption{Comparison of Precision for each category in different models on PipeMFL-240K Testing Set. The best results in each column are highlighted in bold and the second-best values are underlined. "-" indicates not detected.}
\label{tab:result_full_p}
\begin{threeparttable}
\resizebox{\textwidth}{!}{
\setlength{\tabcolsep}{0.65mm}{
\begin{tabular}{p{3.3cm}p{1cm}<{\centering}p{1cm}<{\centering}p{1cm}<{\centering}p{1cm}<{\centering}p{1cm}<{\centering}p{1cm}<{\centering}p{1cm}<{\centering}p{1cm}<{\centering}p{1cm}<{\centering}p{1cm}<{\centering}p{1cm}<{\centering}p{1cm}<{\centering}p{1cm}<{\centering}p{1cm}<{\centering}}
\toprule
& \multicolumn{4}{c}{\textbf{Damage}} & \multicolumn{8}{c}{\textbf{Component}} \\
\cmidrule(lr){2-5} \cmidrule(lr){6-13}
\textbf{Model} & \textbf{MTL} & \textbf{CRC} & \textbf{GWA} & \textbf{SWA} & \textbf{BND} & \textbf{SLE} & \textbf{BRN} & \textbf{TEE} & \textbf{CAS} & \textbf{VAL} & \textbf{ESP} & \textbf{FLA} & \textbf{Overall}  \\
\midrule
Faster R-CNN~\cite{2015FASTRCNN} & - & 0.015 & 0.076 & 0.101 & 0.152 &  0.436 & - & - & - & - & - & - & 0.065\\
RetinaNet~\cite{retina} & 0.068 &0.012 & 0.020 & 0.023 & 0.281 & 0.130 & - & 0.189 & 0.103 & 0.320 & 0.188 & - & 0.111   \\
YOLOv5-s~\cite{yolov5} &0.402 &0.112 &0.312 &0.292 &0.441 &0.632 &0.638 &0.709 &0.409 &0.790 &0.500 &0.200 &0.453\\
YOLOv5-m~\cite{yolov5}  & 0.415 & 0.116 & 0.348 & 0.312 & 0.449 & 0.598 & 0.605 & 0.614 & 0.357 & 0.767 & 0.583 & 0.200 & 0.446\\
YOLOv5-l~\cite{yolov5} & 0.391 & 0.147 & 0.296 & 0.313 &  0.479& 0.634 & 0.650 & 0.714 & 0.484 & \textbf{0.793} & 0.636 &0.375 & 0.493\\
YOLOv8-s~\cite{yolov8_ultralytics} & 0.375& 0.122& 0.321&0.295 &0.396 &0.461 & 0.618&0.500 &0.400 &0.719 &0.462 & 0.286&0.413\\
YOLOv8-m~\cite{yolov8_ultralytics} &0.401 &0.111 &0.339 &0.269 &0.451 &0.639 &0.609 &0.689 &0.329 &0.789 &0.500 &0.250 &0.448\\
YOLOv8-l~\cite{yolov8_ultralytics} & 0.423&0.175 &0.332 &0.338 &0.491 &0.642 & 0.613&0.846 &0.546 &\textbf{0.793} &0.546 & 0.600&0.529\\
YOLO11-s~\cite{yolo11} &0.392 &0.093 &0.318 &0.344 &0.425 & 0.445& 0.382& 0.833& 0.469& 0.657& 0.571& 0.500& 0.453\\
YOLO11-m~\cite{yolo11} & 0.484 & 0.170 & 0.402 & 0.421 &0.447  & 0.725 & 0.529 & 0.700 & 0.458 & \textbf{0.793} & 0.600 & 0.400 & 0.511\\
YOLO11-l~\cite{yolo11} & 0.489 & 0.142 & \textbf{0.447} & \textbf{0.471}& 0.468& 0.765& 0.700& 0.778& 0.500& \textbf{0.793}& \underline{0.700}& 0.750& 0.583\\
YOLOv8-s-world-v2~\cite{2024yoloworld} & 0.384 & 0.105 & 0.301 & 0.321 & 0.443 & 0.604 & 0.641 & 0.800 & 0.419 & 0.677 & 0.294 & 0.333 & 0.444\\
YOLOv8-m-world-v2~\cite{2024yoloworld} & \underline{0.493} & 0.136 & 0.401 & 0.377 & 0.475 & 0.639 & 0.667 & 0.769 & 0.529 & 0.719 & 0.438 & 0.750 & 0.533\\
YOLOv8-l-world-v2~\cite{2024yoloworld} & 0.428 & 0.231 & 0.310 & 0.382 & 0.464 & 0.708 & 0.641 & 0.821 & 0.500 & \textbf{0.793} & 0.382 & 0.333 & 0.507\\
YOLOv8-x-world-v2~\cite{2024yoloworld} & \textbf{0.517} & \textbf{0.281} & 0.400 & \underline{0.455} & \underline{0.498} & \textbf{0.800} & \textbf{0.758} & 0.811 & 0.484 & 0.767 & 0.583 & \textbf{1.000} & \underline{0.613}\\
RT-DETR~\cite{2024RTDETR} & 0.174 & \underline{0.256} & 0.108 & 0.057 & \textbf{0.508} & - & 0.667 & 0.875 & 0.389 & - & - & - & 0.253 \\
RF-DETR-Base~\cite{rfdetr} & 0.167 & 0.108 & 0.173 & 0.064 & 0.326 & 0.398 & 0.518 & 0.694 & 0.333 & 0.697 & 0.313 & 0.750 & 0.378\\ 
YOLO26-n~\cite{yolo26}  & 0.412 & 0.100 & 0.387 & 0.353 & 0.435 & 0.584 & \underline{0.739} & \underline{0.889} & \textbf{1.000} & 0.697 & - & 0.667 & 0.522\\ 
YOLO26-s~\cite{yolo26}   & 0.418 & 0.086 & 0.349 & 0.366 & 0.440 & 0.663 & 0.710 & 0.875 & 0.520 & 0.639 & 0.191 & 0.429 & 0.474\\ 
YOLO26-m~\cite{yolo26}   & 0.454 & 0.156 & 0.411 & 0.396 & 0.457 & \underline{0.781} & 0.677 & 0.861 & \underline{0.750} & 0.697 & 0.385 & 0.750 & 0.585\\ 
YOLO26-l~\cite{yolo26}    & 0.460 & 0.143 & \underline{0.427} & 0.360 & 0.479 & 0.641 & 0.586 & \textbf{0.913} & 0.632 & 0.750 & \textbf{1.000} & \textbf{1.000} & \textbf{0.616}\\ 
YOLO26-x~\cite{yolo26}   & 0.427 & 0.101 &  0.356& 0.314 & 0.448 & 0.680 & 0.657 & 0.703 & 0.520 & 0.639 & 0.500 &0.500  & 0.487\\  

\bottomrule
\end{tabular}}}
\end{threeparttable}
\end{table*}

\begin{table*}[!t]
\centering
\caption{Comparison of Recall for each category in different models on PipeMFL-240K Testing Set. The best results in each column are highlighted in bold and the second-best values are underlined. "-" indicates not detected.}
\label{tab:result_full_recall}
\begin{threeparttable}
\resizebox{\textwidth}{!}{
\setlength{\tabcolsep}{0.65mm}{
\begin{tabular}{p{3.3cm}p{1cm}<{\centering}p{1cm}<{\centering}p{1cm}<{\centering}p{1cm}<{\centering}p{1cm}<{\centering}p{1cm}<{\centering}p{1cm}<{\centering}p{1cm}<{\centering}p{1cm}<{\centering}p{1cm}<{\centering}p{1cm}<{\centering}p{1cm}<{\centering}p{1cm}<{\centering}p{1cm}<{\centering}}
\toprule
& \multicolumn{4}{c}{\textbf{Damage}} & \multicolumn{8}{c}{\textbf{Component}} \\
\cmidrule(lr){2-5} \cmidrule(lr){6-13}
\textbf{Model} & \textbf{MTL} & \textbf{CRC} & \textbf{GWA} & \textbf{SWA} & \textbf{BND} & \textbf{SLE} & \textbf{BRN} & \textbf{TEE} & \textbf{CAS} & \textbf{VAL} & \textbf{ESP} & \textbf{FLA} & \textbf{Overall}  \\
\midrule
Faster R-CNN~\cite{2015FASTRCNN} & - & 0.338 & 0.122 & \underline{0.403} & 0.576 & 0.828 & - & - &-  & - & - & - &0.189 \\
RetinaNet~\cite{retina} & \underline{0.524} & 0.157 & 0.156 & 0.400 & 0.490 & \underline{0.914} & - & 0.368 & 0.240 & 0.696 & 0.333 & -  &0.356 \\
YOLOv5-s~\cite{yolov5} &0.368  &0.369  &0.078  &0.191  &0.823  &0.900  &0.511  &0.760  &0.600  &\textbf{1.000}  &0.667  &0.500  &0.564 \\
YOLOv5-m~\cite{yolov5} &0.382  &0.397  &0.092 &0.188 &0.846  &0.897  &0.549  &0.750  &0.600  &\textbf{1.000}  &\textbf{0.778}  &0.500  &0.577 \\
YOLOv5-l~\cite{yolov5} &0.451  &0.342  &\underline{0.168} &0.250 &0.858 &0.897 &\underline{0.578}  &0.790  &0.600  &\textbf{1.000}  &\textbf{0.778}  &\textbf{0.750}  & \underline{0.622} \\
YOLOv8-s~\cite{yolov8_ultralytics} & 0.413&0.305 &0.097 &0.208 & 0.816& \underline{0.914}&0.467& 0.579& 0.560&\textbf{1.000} &0.667 &0.500 &0.544\\
YOLOv8-m~\cite{yolov8_ultralytics} &0.390  &0.401  &0.112  &0.221  &0.880  &0.900  &0.508  &0.820  &0.640  &\textbf{1.000}  &\textbf{0.778}  & 0.500 & 0.596\\
YOLOv8-l~\cite{yolov8_ultralytics} & 0.403& 0.310& 0.136&0.218 &\textbf{0.889} &0.897 & 0.422& 0.868& 0.480&\textbf{1.000} &0.667 &\textbf{0.750} &0.587\\
YOLO11-s~\cite{yolo11} & 0.400 &0.263  &0.074  & 0.175 & 0.754& \underline{0.914} & 0.289 & 0.658& 0.600 & \textbf{1.000} & 0.444 &0.250 & 0.485\\
YOLO11-m~\cite{yolo11} & 0.335 & 0.270 & 0.069 & 0.131 &  0.769 & 0.862 & 0.400 & 0.737 & 0.440 & \textbf{1.000} & 0.667 & 0.500 & 0.515\\
YOLO11-l~\cite{yolo11} & 0.308 &0.317  &0.046  & 0.108 & 0.761 & 0.897 & 0.467 & 0.737 & 0.360 & \textbf{1.000} & \textbf{0.778} & \textbf{0.750} & 0.544 \\
YOLOv8-s-world-v2~\cite{2024yoloworld} & 0.415  & \textbf{0.415} & 0.125 & 0.205 & 0.859 & \textbf{0.948} & 0.556  & 0.842 & \textbf{0.720} & \textbf{1.000} & 0.556 & \textbf{0.750} & 0.616\\
YOLOv8-m-world-v2~\cite{2024yoloworld} & 0.326  & 0.362 & 0.085 & 0.153 & 0.880 & \underline{0.914} & 0.489  & 0.790 & \textbf{0.720} & \textbf{1.000} & \textbf{0.778} & \textbf{0.750} & 0.604\\
YOLOv8-l-world-v2~\cite{2024yoloworld} & 0.419  & 0.324 & \underline{0.168} & 0.245 & \underline{0.883} & 0.879 & 0.556  & 0.842 & 0.680 & \textbf{1.000} & \textbf{0.778} & 0.250 & 0.585\\
YOLOv8-x-world-v2~\cite{2024yoloworld} & 0.309  & 0.247 & 0.096 & 0.126 & 0.872 & 0.897 & 0.556  & 0.790 & 0.600 & \textbf{1.000} & \textbf{0.778} &\textbf{0.750} & 0.585\\
RT-DETR~\cite{2024RTDETR} & 0.446 & 0.035 & 0.160 & 0.325 & 0.561 & - & 0.489 & \textbf{0.921} & 0.280 & - & - & - & 0.268\\
RF-DETR-Base~\cite{rfdetr} & \textbf{0.553} & 0.390 & \textbf{0.236} & \textbf{0.446} & 0.852 & 0.879 & \textbf{0.644} & \underline{0.895} & 0.600  &\textbf{1.000}  & 0.556 &\textbf{0.750}  & \textbf{0.650}\\ 
YOLO26-n~\cite{yolo26}  & 0.346 & 0.284 & 0.033 & 0.138 & 0.738 & 0.897 & 0.378 & 0.211 & 0.240 & \textbf{1.000} & - & 0.500 & 0.397\\ 
YOLO26-s~\cite{yolo26}   & 0.350 & 0.385 & 0.054 & 0.159 & 0.783  & \underline{0.914} & 0.489 & 0.737 & 0.520 & \textbf{1.000} & 0.444 & \textbf{0.750} & 0.549\\ 
YOLO26-m~\cite{yolo26}   & 0.322 & 0.368 & 0.059 & 0.169 & 0.748 & 0.862 & 0.467 & 0.816 & 0.480 & \textbf{1.000} & 0.556 & \textbf{0.750} & 0.550\\ 
YOLO26-l~\cite{yolo26}    & 0.315 & 0.279 & 0.047 & 0.137 & 0.706 &0.862  & 0.378 & 0.553 & 0.480 & 0.913 & 0.222 & 0.500  &  0.449\\ 
YOLO26-x~\cite{yolo26}   &0.402  & \underline{0.406} & 0.094 & 0.231 & 0.829 & \underline{0.914} & 0.511 & 0.684 & 0.520 & \textbf{1.000} & 0.556 & \textbf{0.750} & 0.575\\  
\bottomrule
\end{tabular}}}
\end{threeparttable}
\end{table*}

\begin{table*}[!t]
\centering
\caption{Comparison of F1-score for each category in different models on PipeMFL-240K Testing Set. The best results in each column are highlighted in bold and the second-best values are underlined. "-" indicates not detected.}
\label{tab:result_full_f1}
\begin{threeparttable}
\resizebox{\textwidth}{!}{
\setlength{\tabcolsep}{0.65mm}{
\begin{tabular}{p{3.3cm}p{1cm}<{\centering}p{1cm}<{\centering}p{1cm}<{\centering}p{1cm}<{\centering}p{1cm}<{\centering}p{1cm}<{\centering}p{1cm}<{\centering}p{1cm}<{\centering}p{1cm}<{\centering}p{1cm}<{\centering}p{1cm}<{\centering}p{1cm}<{\centering}p{1cm}<{\centering}p{1cm}<{\centering}}
\toprule
& \multicolumn{4}{c}{\textbf{Damage}} & \multicolumn{8}{c}{\textbf{Component}} \\
\cmidrule(lr){2-5} \cmidrule(lr){6-13}
\textbf{Model} & \textbf{MTL} & \textbf{CRC} & \textbf{GWA} & \textbf{SWA} & \textbf{BND} & \textbf{SLE} & \textbf{BRN} & \textbf{TEE} & \textbf{CAS} & \textbf{VAL} & \textbf{ESP} & \textbf{FLA} & \textbf{Overall}  \\
\midrule
Faster R-CNN~\cite{2015FASTRCNN} &-  & 0.029 & 0.094 & 0.161 & 0.240 & 0.571 & - & - & - &-  &-  & - & 0.068\\
RetinaNet~\cite{retina} & 0.121 & 0.023 & 0.035 & 0.043 & 0.356 & 0.227 & - & 0.250 & 0.145 & 0.438 & 0.240 & - & 0.157\\
YOLOv5-s~\cite{yolov5} &0.384 &0.172 &0.125 &0.231 &0.574 &0.742 &0.567 &0.734 &0.487 &0.882 &0.572 &0.286 &0.480\\
YOLOv5-m~\cite{yolov5} &0.398 &0.157 &0.146 &0.235 &0.586 &0.716 &0.542 &0.675 &0.447 &0.868 &0.667 &0.286 &0.477\\
YOLOv5-l~\cite{yolov5} &\underline{0.419} &0.205 & \underline{0.214}&\underline{0.278} &0.615 &0.743 &\underline{0.612} &0.750 &0.536 &\textbf{0.885} &\underline{0.700} &0.500 &0.538\\
YOLOv8-s~\cite{yolov8_ultralytics} & 0.393 &0.175  &0.148  & 0.244 & 0.533& 0.613 & 0.532 & 0.537& 0.467 & 0.836 & 0.545 &0.364 & 0.449\\
YOLOv8-m~\cite{yolov8_ultralytics}& 0.395 &0.174 &0.168 &0.243 &0.596 &0.747 &0.554 &0.749 &0.434 &0.882 &0.609 &0.333 &0.490\\
YOLOv8-l~\cite{yolov8_ultralytics} & 0.413  & 0.224 & 0.193 & 0.265 &\underline{0.633} & 0.748 &0.500  &\underline{0.857} & 0.511 & \textbf{0.885}&0.600  &0.667 &\underline{0.541}\\
YOLO11-s~\cite{yolo11} & 0.396 & 0.137 & 0.120 & 0.232 &0.543 &0.599 & 0.329 & 0.735 & 0.526& 0.793 & 0.500 & 0.333 & 0.437 \\
YOLO11-m~\cite{yolo11} & 0.396 & 0.209 & 0.117 & 0.199 & 0.563 & 0.787 & 0.456 & 0.718 & 0.449 & \textbf{0.885} & 0.632 & 0.444 & 0.488 \\
YOLO11-l~\cite{yolo11} & 0.378 & 0.196 & 0.084 & 0.175 &  0.579& \underline{0.825} & 0.560 & 0.757 & 0.419 & \textbf{0.885} & \textbf{0.737} &  0.750& 0.529\\
YOLOv8-s-world-v2~\cite{2024yoloworld} & 0.399 & 0.168 & 0.177 & 0.250 & 0.585 & 0.738 & 0.595 & 0.820 & 0.529 & 0.807 & 0.385 & 0.462 & 0.493\\
YOLOv8-m-world-v2~\cite{2024yoloworld} & 0.392 & 0.197 & 0.140 & 0.218 & 0.617 & 0.752 & 0.564 & 0.779 & \textbf{0.610} & 0.836 & 0.560 & 0.750 & 0.535\\
YOLOv8-l-world-v2~\cite{2024yoloworld} & \textbf{0.424} & \textbf{0.270} & \textbf{0.218} & \textbf{0.298} & 0.608 & 0.785 & 0.595 & 0.831 & 0.576 & \textbf{0.885} & 0.583 & 0.286 & 0.530\\
YOLOv8-x-world-v2~\cite{2024yoloworld} & 0.387 & \underline{0.263} & 0.155 & 0.198 & \textbf{0.634} & \textbf{0.846} & \textbf{0.641} & 0.800 & 0.536 & 0.868 & 0.667 & \textbf{0.857} & \textbf{0.571}\\
RT-DETR~\cite{2024RTDETR} & 0.251 & 0.061 & 0.129 &0.097  & 0.533 & - & 0.564 & \textbf{0.898} & 0.326 & - & - & - & 0.238\\
RF-DETR-Base~\cite{rfdetr} &0.256  & 0.169 & 0.200 & 0.112 & 0.472 & 0.548 & 0.574 & 0.782 & 0.429 & 0.821 & 0.400 & 0.750 & 0.460\\ 
YOLO26-n~\cite{yolo26}  & 0.376 & 0.148 & 0.061 & 0.198 & 0.547& 0.708 & 0.500 & 0.340 & 0.387 & 0.822 & - & 0.571 & 0.388\\ 
YOLO26-s~\cite{yolo26}   & 0.381 & 0.140 & 0.093 & 0.221 & 0.564 & 0.768 & 0.579 & 0.800 & 0.520 & 0.780 & 0.267 & 0.546 & 0.472\\ 
YOLO26-m~\cite{yolo26}   & 0.377 & 0.219 & 0.103 & 0.237 & 0.567 & 0.820 & 0.553 & 0.838  & \underline{0.585} & 0.821 & 0.455 & \textbf{0.857} & 0.536 \\ 
YOLO26-l~\cite{yolo26}    & 0.374 & 0.189 & 0.085 & 0.198  & 0.571 & 0.735 & 0.459 & 0.688 & 0.545 & 0.824 & 0.364 & 0.667 & 0.475\\ 
YOLO26-x~\cite{yolo26}   & 0.414 & 0.161 & 0.149 & 0.266 & 0.581 & 0.779 & 0.575 &0.693  &  0.520& 0.780 & 0.526 &0.600  & 0.504\\  
\bottomrule
\end{tabular}}}
\end{threeparttable}
\end{table*}

\begin{figure*}[!t]
    \centering
    \vspace{0.7cm}
    \includegraphics[width=\linewidth]{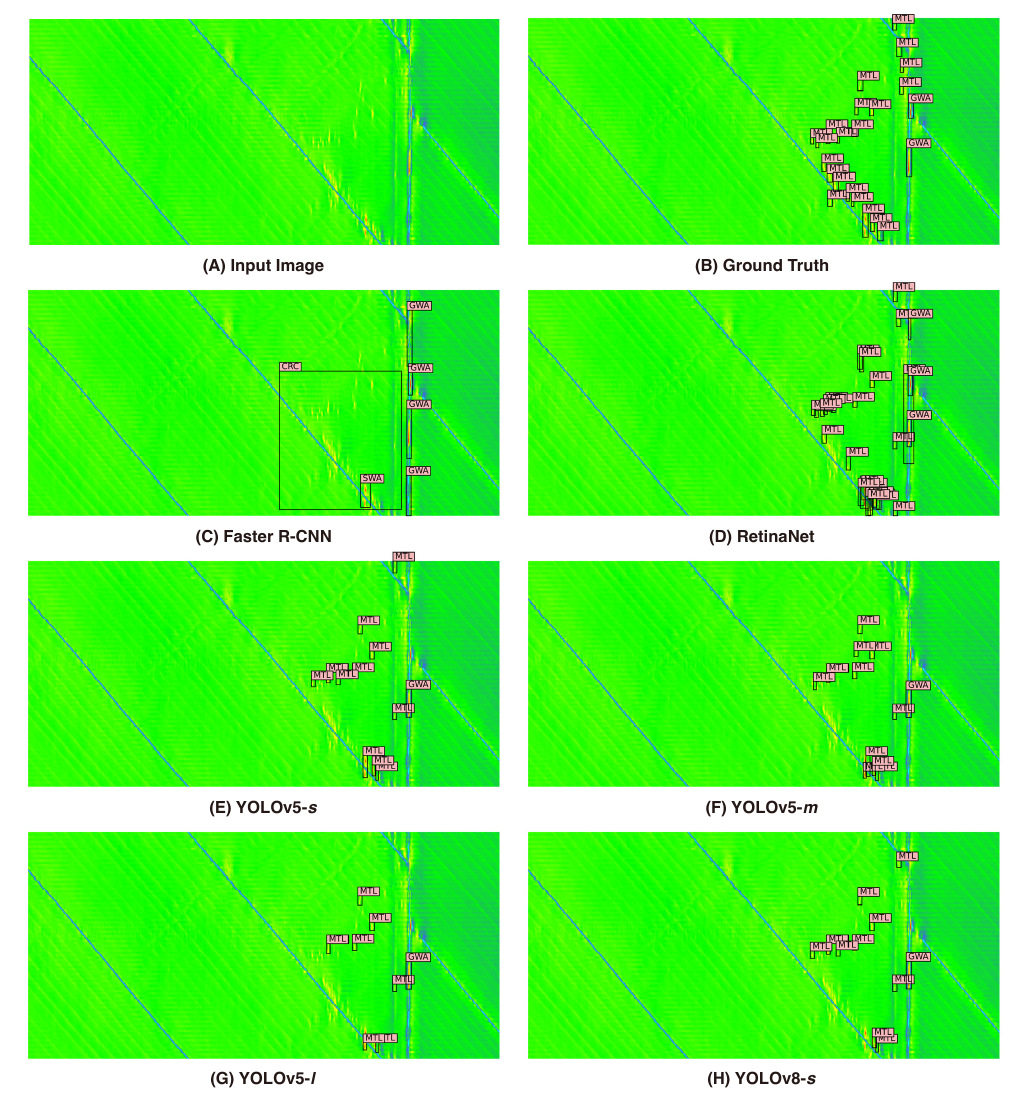}
    \caption{Qualitative benchmark results on representative damage samples (Part A).}
    \vspace{0.7cm}
    \label{fig:visual_results_1_a}
\end{figure*}
\begin{figure*}[!t]
    \centering
    \vspace{0.7cm}
    \includegraphics[width=\linewidth]{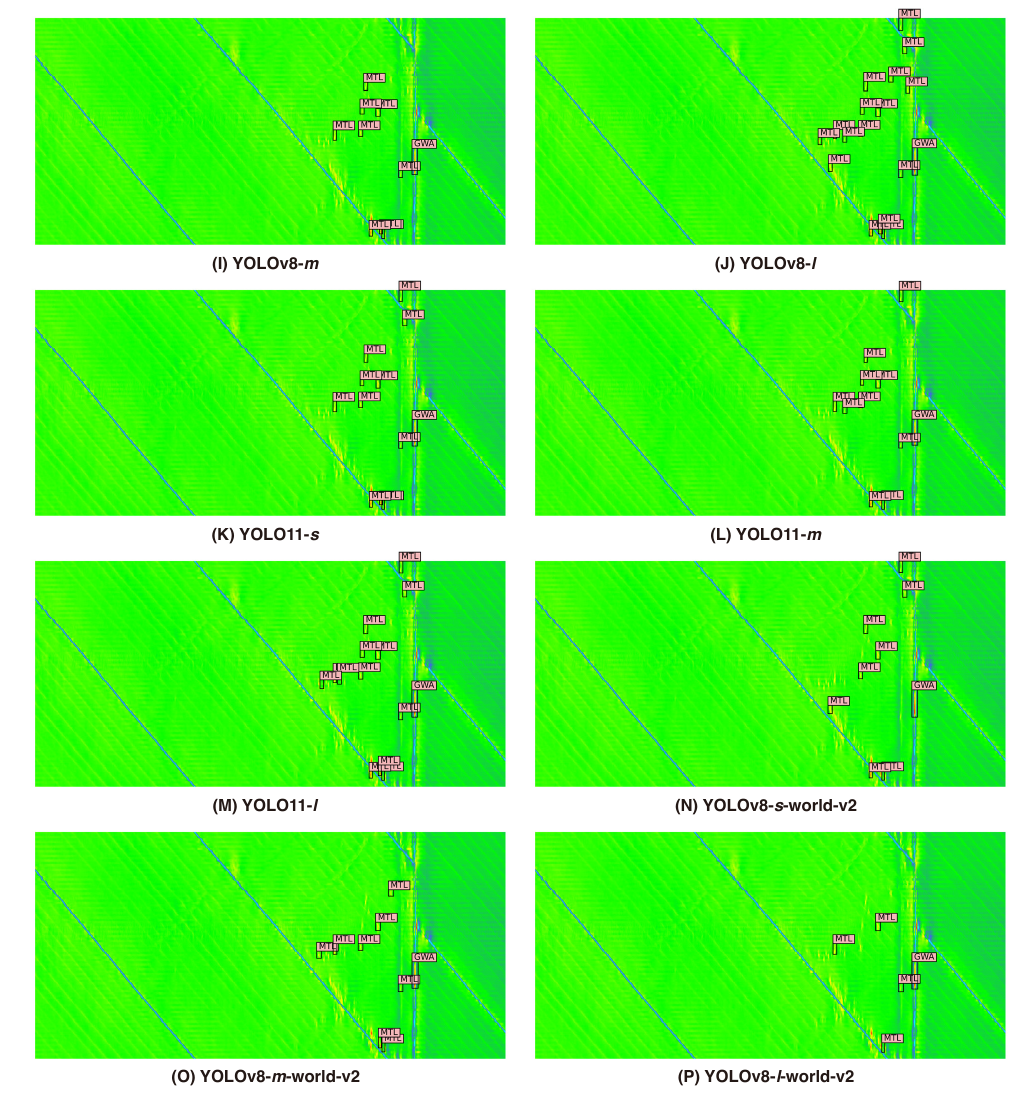}
    \caption{Qualitative benchmark results on representative damage samples (Part B).}
    \vspace{0.7cm}
    \label{fig:visual_results_1_b}
\end{figure*}
\begin{figure*}[!t]
    \centering
    \vspace{0.7cm}
    \includegraphics[width=\linewidth]{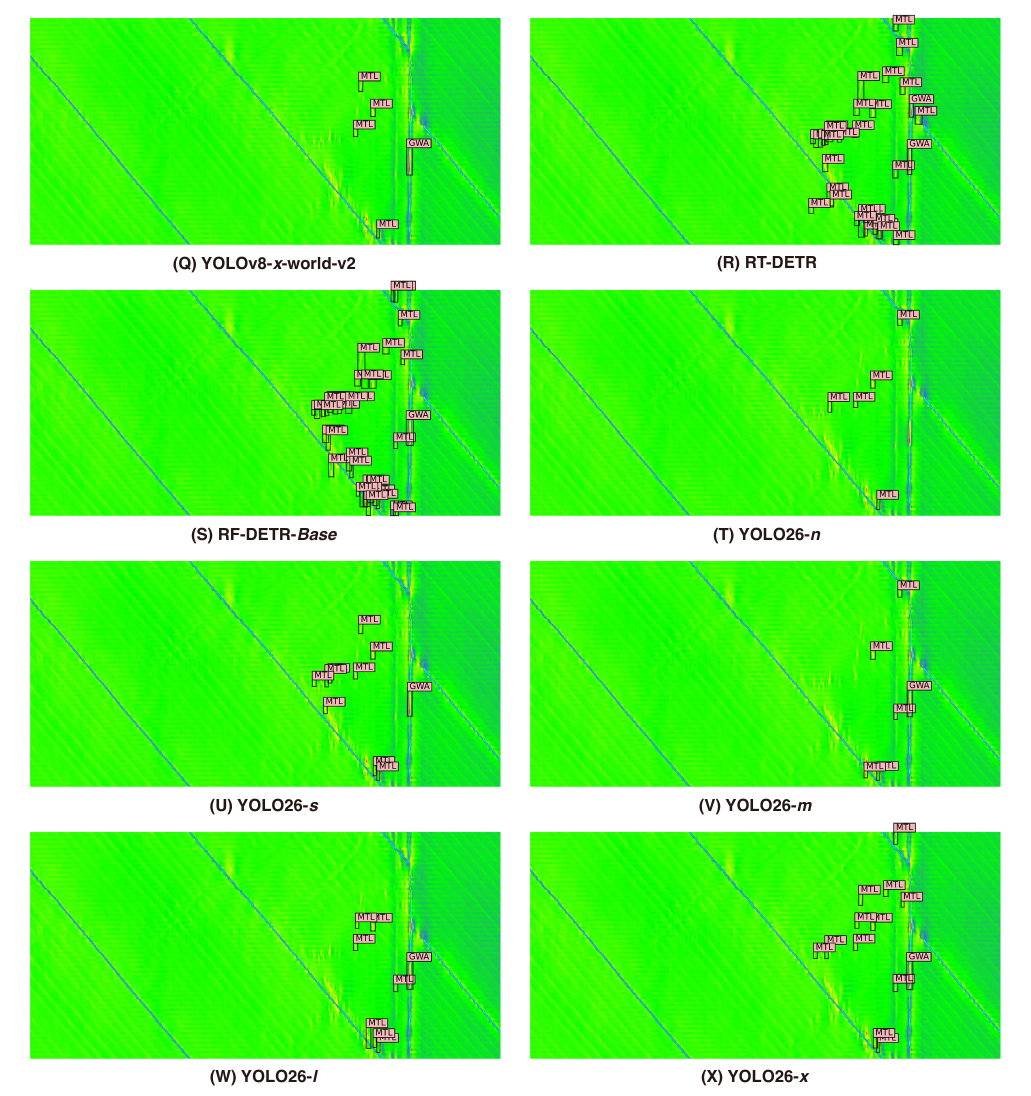}
    \caption{Qualitative benchmark results on representative damage samples (Part C).}
    \vspace{0.7cm}
    \label{fig:visual_results_1_c}
\end{figure*}

\begin{figure*}[!t]
    \centering
    \vspace{0.7cm}
    \includegraphics[width=\linewidth]{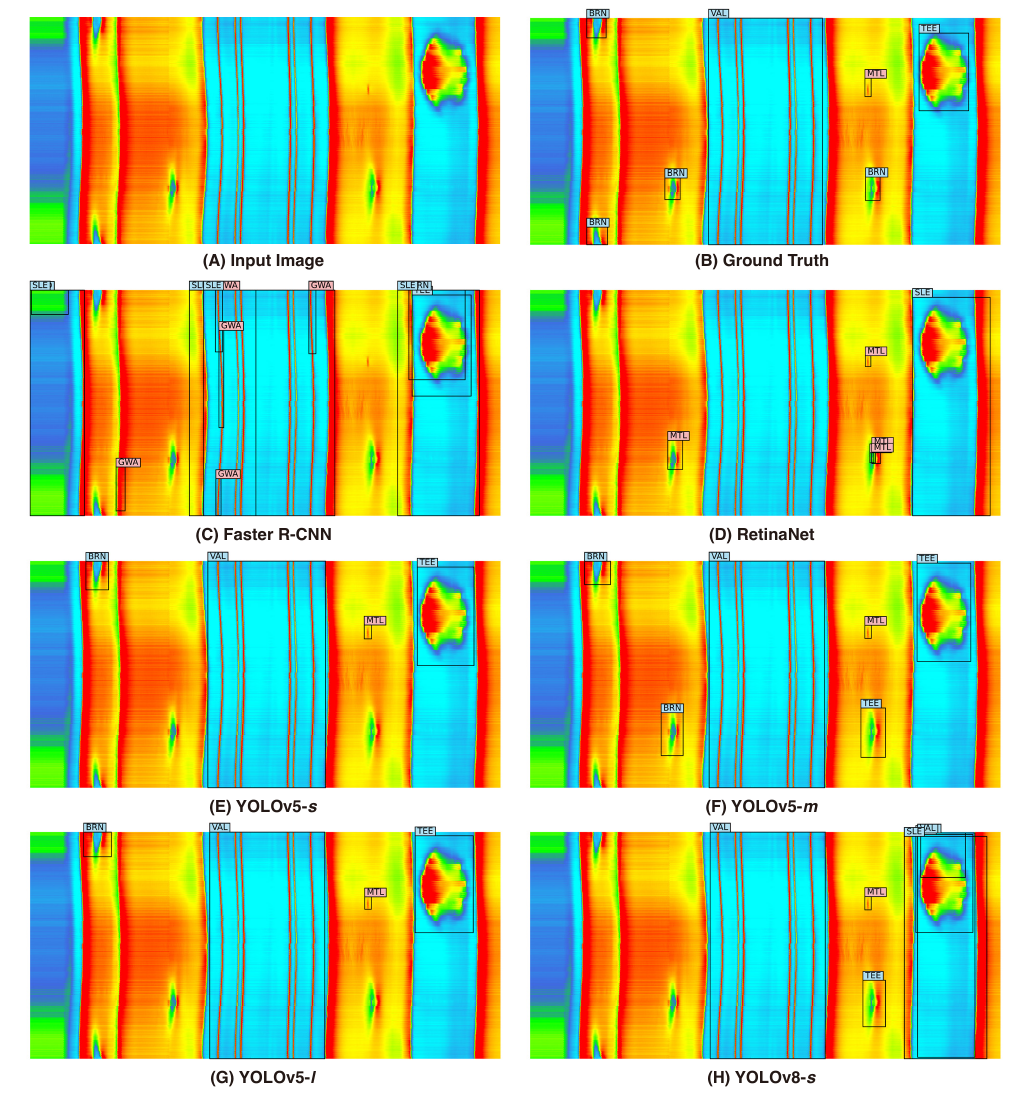}
    \caption{Qualitative benchmark results on representative component samples (Part A).}
    \vspace{0.7cm}
    \label{fig:visual_results_2_a}
\end{figure*}
\begin{figure*}[!t]
    \centering
    \vspace{0.7cm}
    \includegraphics[width=\linewidth]{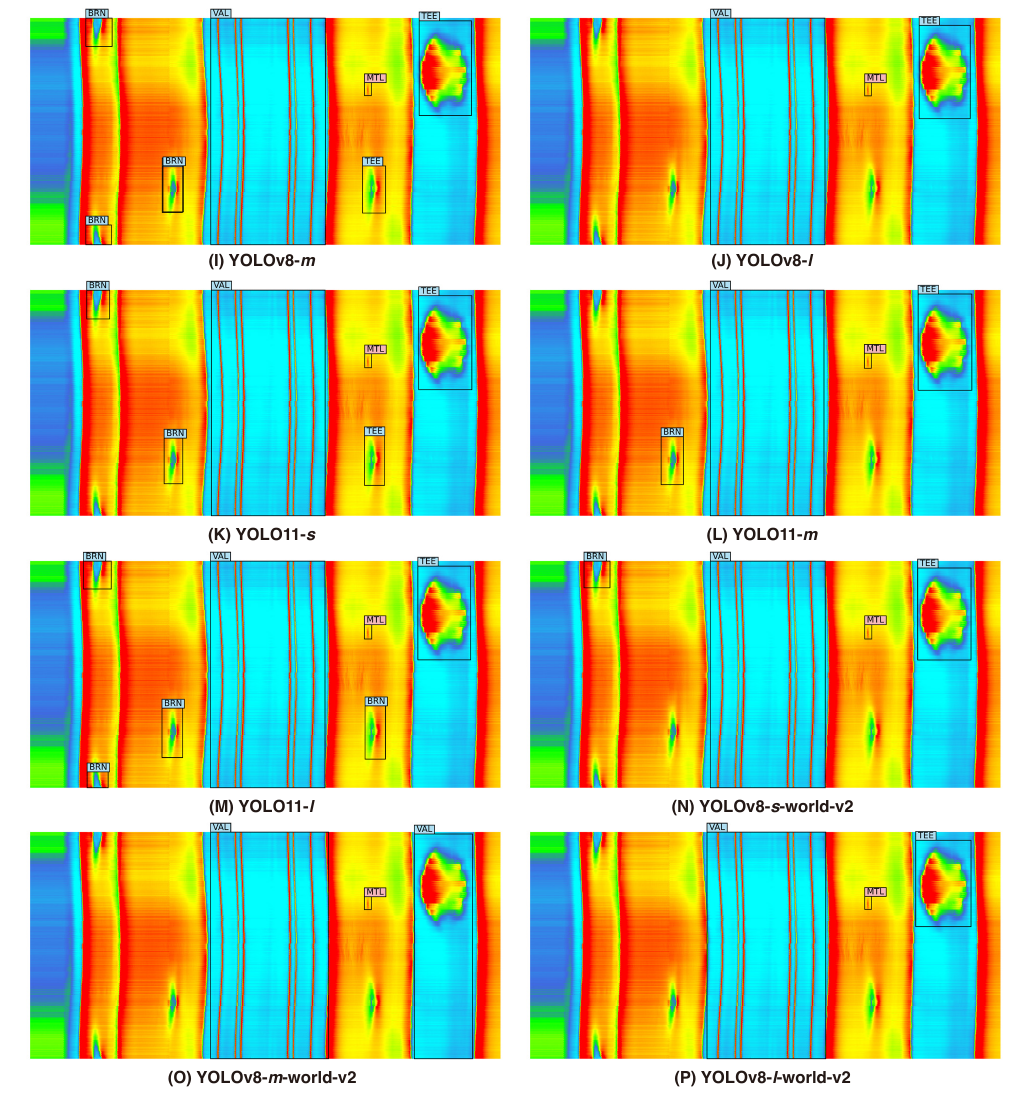}
    \caption{Qualitative benchmark results on representative component samples (Part B).}
    \vspace{0.7cm}
    \label{fig:visual_results_2_b}
\end{figure*}
\begin{figure*}[!t]
    \centering
    \vspace{0.7cm}
    \includegraphics[width=\linewidth]{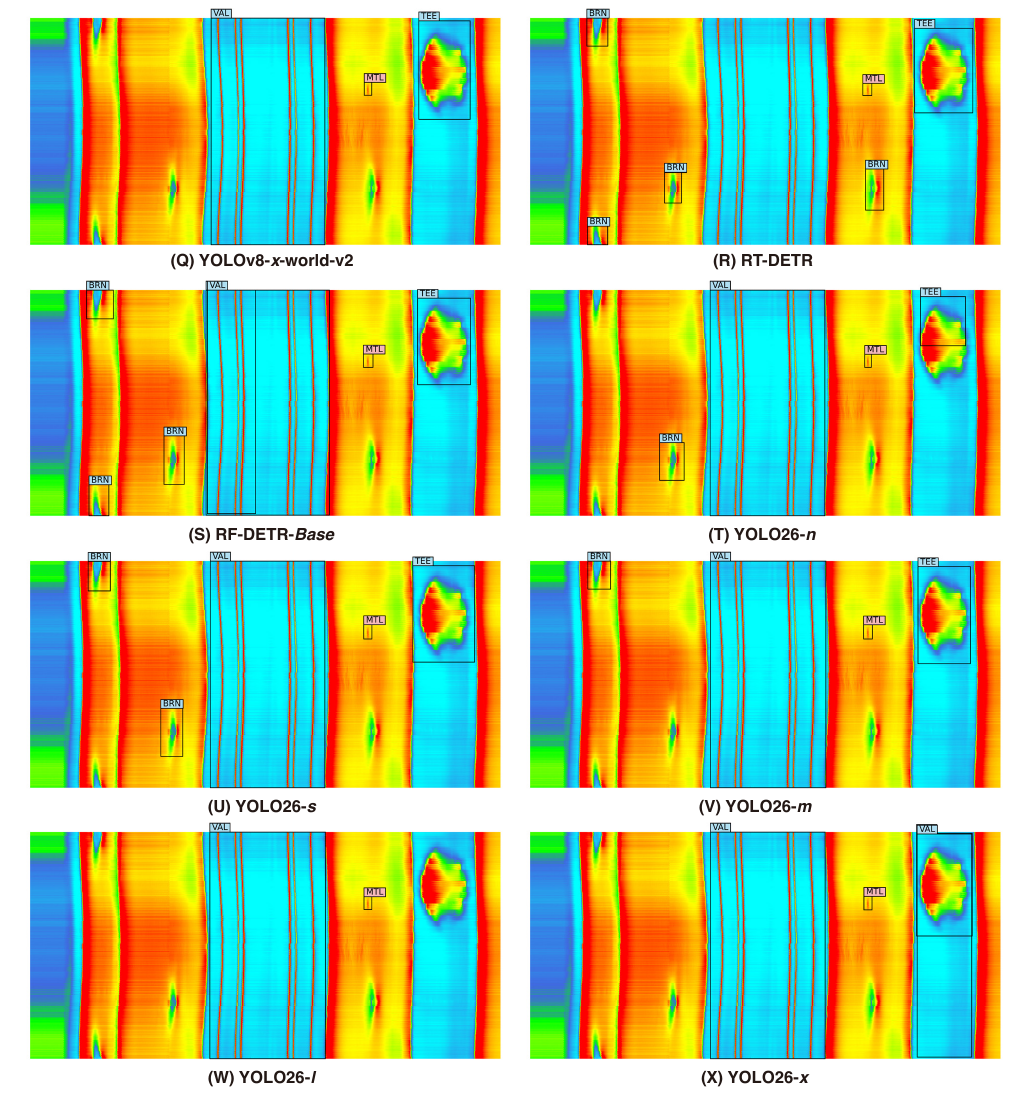}
    \caption{Qualitative benchmark results on representative component samples (Part C).}
    \vspace{0.7cm}
    \label{fig:visual_results_2_c}
\end{figure*}

\begin{figure*}[!t]
    \centering
    \vspace{0.7cm}
    \includegraphics[width=\linewidth]{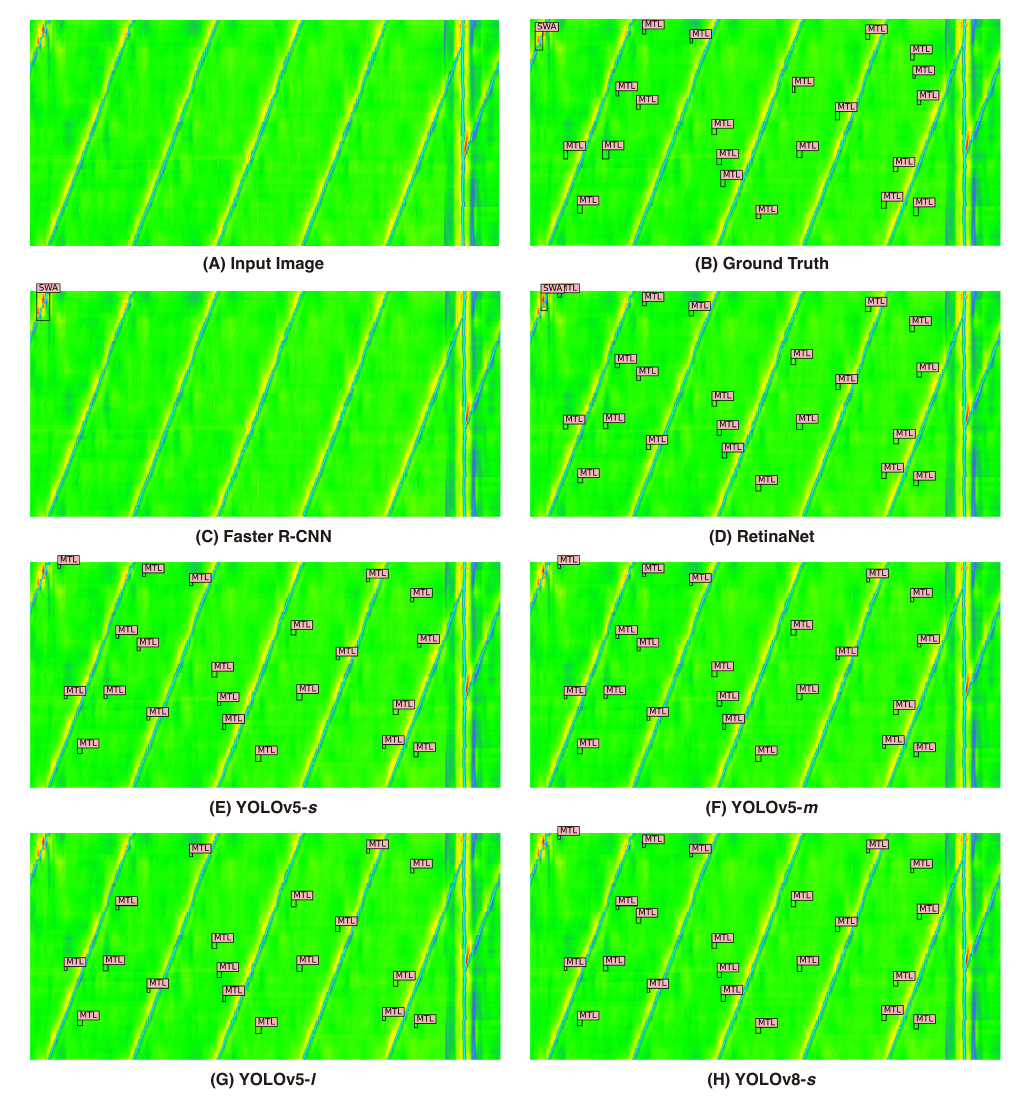}
    \caption{Qualitative benchmark results on representative tiny damage samples (Part A).}
    \vspace{0.7cm}
    \label{fig:visual_results_3_a}
\end{figure*}
\begin{figure*}[!t]
    \centering
    \vspace{0.7cm}
    \includegraphics[width=\linewidth]{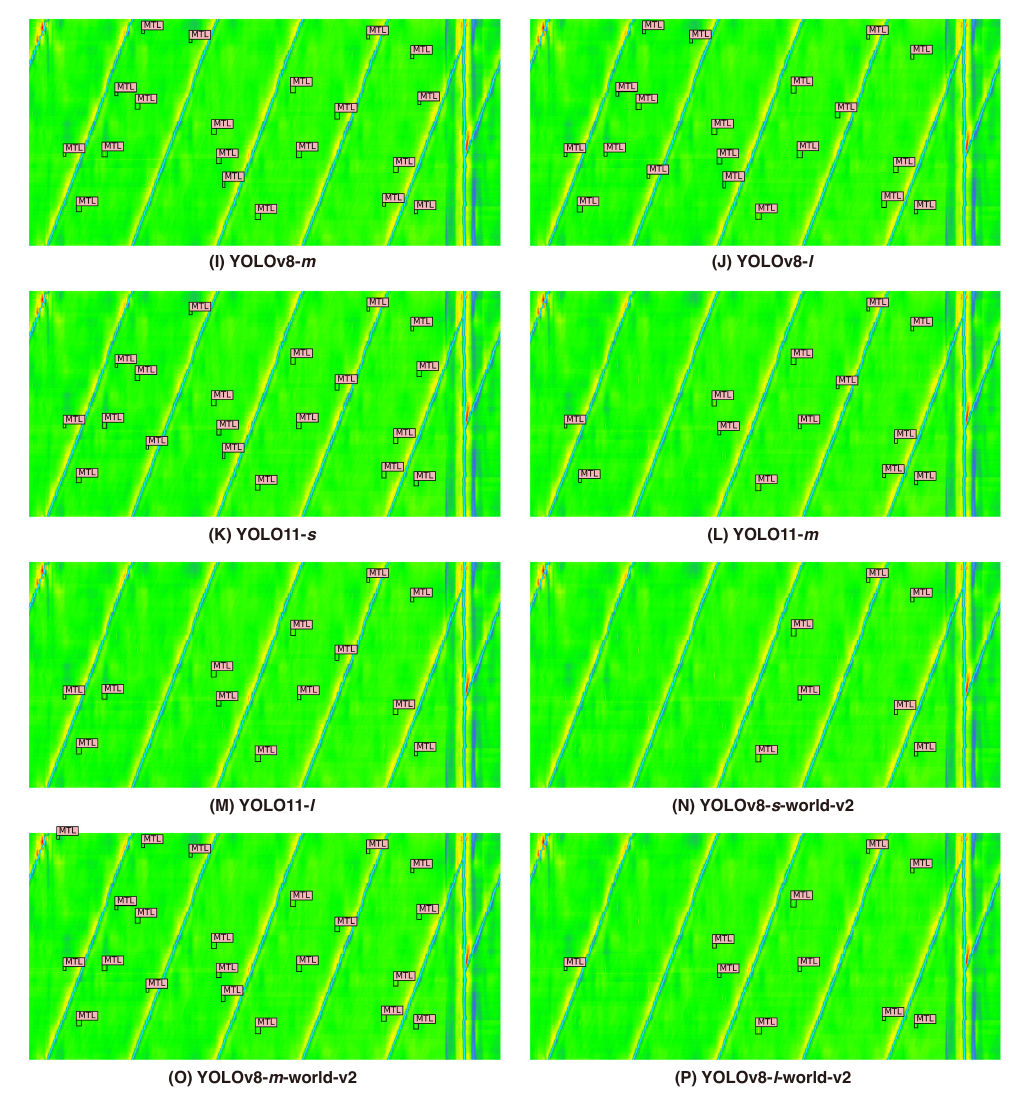}
    \caption{Qualitative benchmark results on representative tiny damage samples (Part B).}
    \vspace{0.7cm}
    \label{fig:visual_results_3_b}
\end{figure*}
\begin{figure*}[!t]
    \centering
    \vspace{0.7cm}
    \includegraphics[width=\linewidth]{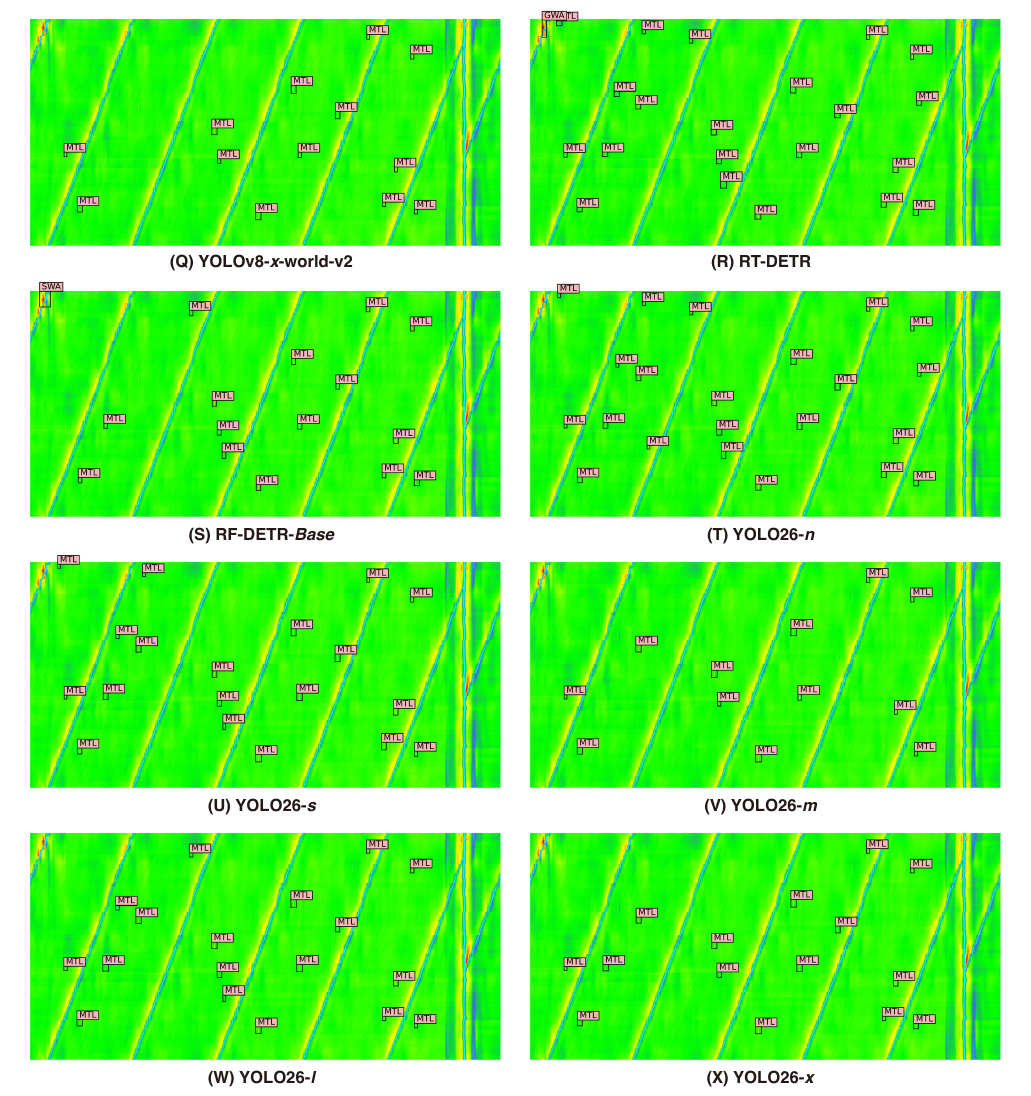}
    \caption{Qualitative benchmark results on representative tiny damage samples (Part C).}
    \vspace{0.7cm}
    \label{fig:visual_results_3_c}
\end{figure*}

\subsection{Overview of Data Distribution by Pipeline}
\label{sec:pipe}


From the perspective of individual pipelines, notable differences are observed in both the total number of objects and the category composition, as shown in Table~\ref{tab:pipeline_characteristics}. Pipelines 240K-J, 240K-I and 240K-B contain relatively large numbers of annotated targets, whereas pipelines such as 240K-C, 240K-D and 240K-F have smaller sample sizes. These discrepancies are partly attributable to variations in pipeline length and inspection coverage and also reflect differences in service conditions and degradation levels among pipeline segments. Despite the pronounced imbalance in sample size across pipelines, their category distribution patterns are generally consistent with the overall trend of the dataset.

Furthermore, as shown in Table~\ref{tab:pipeline_distribution}, substantial differences exist in the number of images and inspection mileage among pipelines, highlighting the strong heterogeneity in pipeline length and the spatial distribution of defects. This characteristic closely aligns with the real-world engineering context of pipeline inspection, which is typically characterized by long distances and high heterogeneity. When normalized by pipeline length, the overall object density is 0.131 objects per meter, but with considerable variation across pipelines. The dataset covers pipeline diameters ranging from 406 to 813 mm and service lifetimes from 1 to 20 years. The correlation analysis reveals a significant positive correlation between object density and pipeline service age (r = 0.6063, 95\% CI: 0.01008–0.8845, p = 0.0480), whereas no significant correlations are observed between object density and either pipeline diameter or pipeline length. This finding is consistent with engineering experience and established corrosion mechanisms, suggesting that corrosion defects tend to accumulate over time rather than being directly determined by geometric scale.

As illustrated in Fig.~\ref{fig:yeardiameter}, the distribution of corrosion density is further analyzed along two dimensions: pipeline service age and diameter. From the temporal perspective, corrosion density exhibits an overall increasing trend with service age. During the early operational stage (approximately 1 year), corrosion density remains generally low, mostly below 100 objects per kilometer. In the mid-term stage (approximately 8–12 years), corrosion density begins to diverge markedly and differences among pipelines with varying diameters gradually become apparent. In the later stage of service (16–20 years), some pipeline segments exceed 200 objects per kilometer, with a few samples approaching or even reaching 500 objects per kilometer, indicating that long-term operation is a major driving factor for the rapid accumulation of corrosion defects. This evolutionary pattern is highly consistent with the commonly observed “time-dependent degradation” behavior in MFL inspections.

From the perspective of diameter effects, larger-diameter pipelines (e.g., 813 mm and 711 mm) already exhibit relatively high corrosion densities during the mid-term service stage, reaching approximately 180–210 objects per kilometer at around 8–9 years. This observation suggests that pipelines with larger diameters may be exposed to more complex corrosion environments under equivalent service durations, or alternatively, that their larger effective inspection area leads to the identification of more detectable defects. Meanwhile, smaller-diameter pipelines (457 mm) show pronounced peaks in corrosion density at later service stages (17–20 years), with one pipeline segment approaching nearly 500 objects per kilometer. This finding indicates that once small-diameter pipelines enter an accelerated degradation phase during long-term operation, the growth rate of defects can become extremely steep.






\subsection{Dataset Maintenance}
As the authors and maintainers of this dataset, we affirm that while the dataset is self-contained and does not depend on any external links or content, we may provide future updates, such as adding new cases or incorporating additional tasks. These potential updates aim to enhance the value of the dataset while maintaining its long-term usability.

\section{Detailed Analysis of Experimental Results}

\subsection{Benchmark Results}
\label{sec:detail_exp}
\subsubsection{Overall Performance Summary}
\ 
\newline
Tables~\ref{tab:result_full_overall},~\ref{tab:result_full_ap50}--\ref{tab:result_full_f1} present the overall and per-category performance of each model, in terms of mAP50, mAP50:95, Precision, Recall and F1-score. The evaluation reveals a clear performance hierarchy across architectural families. The YOLOv8-World-v2 series achieves the highest overall mAP50 (\textbf{0.498}) and mAP50:95 (\textbf{0.327}). Early two-stage detectors such as Faster R-CNN and RetinaNet struggle with the extreme category imbalance and scale variance inherent to MFL images, achieving mAP50 below 0.2 and showing pronounced precision-recall imbalance. YOLO variants demonstrate balanced precision-recall profiles, while transformer-based detectors exhibit polarized behaviors. RF-DETR achieves the highest macro-recall (\textbf{0.650}) but suffers from low precision (0.378), whereas RT-DETR demonstrates complete category omission on continuous structural components (SLE, VAL, FLA achieve 0 AP50) despite excelling on discrete components such as TEE (AP50=\textbf{0.894}). Within the YOLO family, detection performance exhibits a positive correlation with model capacity, where scaling up from small to extra-large variants (e.g., model-s → model-l, YOLOv5, YOLOv8, YOLO11) generally yields consistent improvements in both mAP50 and mAP50:95. Additionally, a clear task difficulty hierarchy emerges: all detectors achieve AP50 $>$ 0.8 on structural components (VAL, SLE) but struggle to achieve AP50 $>$ 0.3 on damage categories (MTL, GWA), highlighting that metal loss detection remains the critical bottleneck.

\begin{table*}[!t]
\centering
\caption{Model comparison on PipeMFL-240K sub-dataset (Confidence score=0.25, IoU=0.5) with different scales (scale=1: 53,375 Positive + 5,930 Negative samples).}
\label{tab:small_data}
\begin{threeparttable}
\setlength{\tabcolsep}{0.8mm}{  
\begin{tabular}{p{2.4cm}@{\centering} 
               p{2cm}<{\centering}
               p{2cm}<{\centering}
               p{2cm}<{\centering}
               p{2.4cm}<{\centering}
               p{2.4cm}<{\centering}
               p{2.4cm}<{\centering}
}
\toprule
\textbf{Model}  & \textbf{Dataset Scale}  & \textbf{mAP50} & \textbf{mAP50:95} & \textbf{Precision} & \textbf{Recall} & \textbf{F1-score}  \\
 &  &  &  & \textbf{(macro, micro)} & \textbf{(macro, micro)} & \textbf{(macro, micro)} \\
\midrule 
YOLOv8-m & 1 & 0.450 & 0.288 & 0.448, 0.375& 0.596, 0.356 & 0.490, 0.365   \\
YOLOv8-m & 1/2 & 0.404 & 0.250 & 0.384, 0.315 & 0.541, 0.414 & 0.435, 0.358   \\
YOLOv8-m & 1/4 & 0.417 & 0.224 & 0.376, 0.245 & 0.563, 0.404 & 0.428, 0.305   \\
YOLOv8-m & 1/8 & 0.290 & 0.168 & 0.348, 0.232 & 0.407, 0.355& 0.351, 0.280   \\
YOLOv8-m & 1/16 & 0.169 & 0.119 & 0.229, 0.213 & 0.281, 0.361 & 0.194, 0.268   \\
RF-DETR-Base& 1 & 0.472 & 0.273 & 0.378, 0.151 & 0.650, 0.514 & 0.460, 0.233   \\
RF-DETR-Base & 1/2 & 0.447 & 0.263 & 0.327, 0.157 & 0.650, 0.493 & 0.416, 0.238   \\
RF-DETR-Base& 1/4 & 0.413 & 0.250 & 0.327, 0.136 &  0.633, 0.499& 0.408, 0.214   \\
RF-DETR-Base& 1/8 & 0.346 & 0.194 & 0.266, 0.138 & 0.552, 0.460 & 0.332, 0.213   \\
RF-DETR-Base & 1/16 & 0.256 & 0.160 & 0.220, 0.137 &  0.433, 0.405  & 0.274, 0.205 \\
YOLO26-m& 1 & 0.442 & 0.263 & 0.585, 0.436 & 0.550, 0.291 & 0.536, 0.349   \\
YOLO26-m & 1/2 & 0.418 & 0.239 & 0.512, 0.389 & 0.540, 0.323 &  0.497, 0.353  \\
YOLO26-m & 1/4 & 0.370 & 0.222 & 0.418, 0.339 & 0.494, 0.350 & 0.428, 0.345   \\
YOLO26-m& 1/8 & 0.184 &  0.126 & 0.327, 0.354 & 0.248, 0.300 &  0.239, 0.324  \\
YOLO26-m& 1/16 & 0.100 & 0.044 & 0.170, 0.283 & 0.167, 0.292 & 0.145, 0.288   \\

\bottomrule
\end{tabular}}
\end{threeparttable}
\end{table*}

\begin{table*}[!t]
\centering
\caption{Comparison of AP50 for each category in models with different dataset scales (scale=1: 53,375 Positive + 5,930 Negative samples). "-" indicates not detected.}
\label{tab:small_data_ap50}
\begin{threeparttable}
\resizebox{\textwidth}{!}{
\setlength{\tabcolsep}{0.65mm}{
\begin{tabular}{p{2cm}p{1cm}<{\centering}p{1cm}<{\centering}p{1cm}<{\centering}p{1cm}<{\centering}p{1cm}<{\centering}p{1cm}<{\centering}p{1cm}<{\centering}p{1cm}<{\centering}p{1cm}<{\centering}p{1cm}<{\centering}p{1cm}<{\centering}p{1cm}<{\centering}p{1cm}<{\centering}p{1cm}<{\centering}p{1cm}<{\centering}}
\toprule
& & \multicolumn{4}{c}{\textbf{Damage}} & \multicolumn{8}{c}{\textbf{Component}} \\
\cmidrule(lr){3-6} \cmidrule(lr){7-14}
\textbf{Model} & \textbf{Scale} & \textbf{MTL} & \textbf{CRC} & \textbf{GWA} & \textbf{SWA} & \textbf{BND} & \textbf{SLE} & \textbf{BRN} & \textbf{TEE} & \textbf{CAS} & \textbf{VAL} & \textbf{ESP} & \textbf{FLA} & \textbf{Overall}  \\
\midrule
YOLOv8-m& 1 &0.243 &0.113 &0.051 &0.090 &0.571 &0.871 &0.362 &0.771 &0.531 &0.891 &0.651 &0.250 &0.450\\
YOLOv8-m& 1/2 & 0.262 & 0.100 & 0.056 & 0.100 & 0.553 & 0.856 & 0.360 & 0.579 & 0.483 & 0.851  & 0.400  & 0.250 & 0.404\\
YOLOv8-m& 1/4 & 0.217  & 0.069  & 0.044 & 0.074 & 0.529 & 0.714 & 0.234 & 0.512 & 0.359 & 0.849 & 0.651 & 0.750 & 0.417 \\
YOLOv8-m& 1/8 & 0.188 & 0.010  & 0.036 & 0.067 & 0.286 & 0.694 & 0.257 & 0.255 & 0.224 & 0.785 & 0.174 & 0.500 & 0.290\\
YOLOv8-m& 1/16 & 0.181  & 0.030  & 0.024 & 0.053 & 0.184 & 0.733 & - & 0.053 & - & 0.723  &-  & 0.050 & 0.169\\
RF-DETR-Base& 1 & 0.246 & 0.128 & 0.070 & 0.108 & 0.501 & 0.810 & 0.479 & 0.878 & 0.365 &0.849  & 0.477 &  0.750& 0.472\\ 
RF-DETR-Base& 1/2 & 0.211 & 0.127 & 0.045 & 0.069 & 0.442 &  0.724& 0.353 & 0.861 & 0.471 &0.776  &0.785  & 0.500 & 0.447\\
RF-DETR-Base& 1/4 & 0.194  & 0.082  & 0.049 & 0.055 & 0.426 & 0.660 & 0.463 & 0.750 & 0.524 & 0.776  & 0.636  & 0.286 &0.413\\
RF-DETR-Base& 1/8 & 0.191 & 0.134  & 0.050 & 0.090 & 0.229 & 0.685 & 0.443 & 0.655 & 0.214 & 0.703  & 0.477  & 0.279 & 0.346\\
RF-DETR-Base& 1/16 & 0.151  & 0.046 & 0.031 & 0.051 & 0.119 & 0.613 & 0.338 & 0.547 & 0.158 & 0.690 & 0.333 & - & 0.256\\
YOLO26-m& 1 & 0.206 & 0.149 & 0.032 & 0.088 & 0.430 & 0.829 & 0.346 &0.775  & 0.452 & 0.864 & 0.386 & 0.750& 0.442\\  
YOLO26-m& 1/2 & 0.223  &0.109   &0.030  & 0.091 & 0.418 &  0.865& 0.333 & 0.719 & 0.401 & 0.840  &0.574  & 0.417 & 0.418\\
YOLO26-m& 1/4 & 0.212  & 0.081 & 0.032 & 0.082 & 0.455 & 0.819 & 0.278 & 0.566 & 0.271 & 0.835  & 0.478  &  0.333 & 0.370\\
YOLO26-m& 1/8  & 0.189 & 0.012 & 0.015 & 0.062 & 0.275 & 0.640 & 0.084 & - & 0.165  & 0.764  & - & - & 0.184\\
YOLO26-m& 1/16 & 0.162 & 0.009 & 0.004 & 0.026 & 0.019 & 0.497 & - & - &-  & 0.472  & - & 0.006 & 0.100\\

\bottomrule
\end{tabular}}}
\end{threeparttable}
\end{table*}

\begin{table*}[!t]
\centering
\caption{Comparison of Precision for each category in models with different dataset scales (scale=1: 53,375 Positive + 5,930 Negative samples). "-" indicates not detected.}
\label{tab:small_data_p}
\begin{threeparttable}
\resizebox{\textwidth}{!}{
\setlength{\tabcolsep}{0.65mm}{
\begin{tabular}{p{2cm}p{1cm}<{\centering}p{1cm}<{\centering}p{1cm}<{\centering}p{1cm}<{\centering}p{1cm}<{\centering}p{1cm}<{\centering}p{1cm}<{\centering}p{1cm}<{\centering}p{1cm}<{\centering}p{1cm}<{\centering}p{1cm}<{\centering}p{1cm}<{\centering}p{1cm}<{\centering}p{1cm}<{\centering}}
\toprule
& & \multicolumn{4}{c}{\textbf{Damage}} & \multicolumn{8}{c}{\textbf{Component}} \\
\cmidrule(lr){3-6} \cmidrule(lr){7-14}
\textbf{Model} & \textbf{Scale} & \textbf{MTL} & \textbf{CRC} & \textbf{GWA} & \textbf{SWA} & \textbf{BND} & \textbf{SLE} & \textbf{BRN} & \textbf{TEE} & \textbf{CAS} & \textbf{VAL} & \textbf{ESP} & \textbf{FLA} & \textbf{Overall}  \\
\midrule
YOLOv8-m& 1 &0.401 &0.111 &0.339 &0.269 &0.451 &0.639 &0.609 &0.689 &0.329 &0.789 &0.500 &0.250 &0.448\\
YOLOv8-m& 1/2 & 0.335 & 0.085 & 0.278 & 0.229 & 0.434 & 0.541 & 0.512 & 0.563 & 0.469 & 0.719  & 0.191 & 0.250 & 0.384\\
YOLOv8-m& 1/4 &0.268  & 0.064  & 0.195 & 0.148 & 0.428 & 0.565 & 0.484 & 0.647 & 0.324 & 0.767  & 0.429  & 0.188 & 0.376\\
YOLOv8-m& 1/8 & 0.255  &  0.022 & 0.143 & 0.205 & 0.240 & 0.469 & 0.778 & 0.500 & 0.333 & 0.636  & 0.200  & 0.400& 0.348\\
YOLOv8-m& 1/16 & 0.241 & 0.029  & 0.128 & 0.156 & 0.137 & 0.485 & - & 1.000 & - & 0.476  &-  &  0.100& 0.241\\
RF-DETR-Base& 1 & 0.167 & 0.108 & 0.173 & 0.064 & 0.326 & 0.398 & 0.518 & 0.694 & 0.333 & 0.697 & 0.313 & 0.750 & 0.378\\ 
RF-DETR-Base& 1/2 & 0.172  & 0.086  & 0.146 & 0.074 & 0.299 & 0.345 & 0.362 & 0.783 & 0.227 & 0.657  & 0.667  & 0.105 & 0.327\\
RF-DETR-Base& 1/4 &0.153  & 0.057  & 0.127 & 0.064 & 0.250 & 0.537 & 0.406 & 0.667 & 0.295 &0.731  & 0.539  & 0.098 & 0.327\\
RF-DETR-Base& 1/8 & 0.148 & 0.063  & 0.140 & 0.084 & 0.196 & 0.275 &0.294 &0.519 & 0.444 & 0.667 & 0.304 & 0.054  &0.266 \\
RF-DETR-Base& 1/16 & 0.152 & 0.060  & 0.129 & 0.067 & 0.201 & 0.400 &0.348 &0.600 & 0.139 & 0.210 & 0.333 &-  &0.220 \\
YOLO26-m& 1 & 0.454 & 0.156 & 0.411 & 0.396 & 0.457 & 0.781 & 0.677 & 0.861 & 0.750 & 0.697 & 0.385 & 0.750 & 0.585\\ 
YOLO26-m& 1/2 & 0.408  & 0.100  &0.355  &0.372  &0.432  &0.776  & 0.613 & 0.775 & 0.619 & 0.677 & 0.353  & 0.667 & 0.512\\
YOLO26-m& 1/4 & 0.359  & 0.076 & 0.256 & 0.300 & 0.437 & 0.520 & 0.500 & 0.800 & 0.500 & 0.667  & 0.313  & 0.286 & 0.418\\
YOLO26-m& 1/8  & 0.364 & 0.055 & 0.336 & 0.329 & 0.382 & 0.525 & 0.667 & - & 0.714  & 0.556  & - & - & 0.327\\
YOLO26-m& 1/16 & 0.299 & 0.043 & 0.165 & 0.163 & 0.261 & 0.756 & - & - &-  & 0.333  & - & 0.022 & 0.170\\

\bottomrule
\end{tabular}}}
\end{threeparttable}
\end{table*}

\begin{table*}[!t]
\centering
\caption{Comparison of Recall for each category in models with different dataset scales (scale=1: 53,375 Positive + 5,930 Negative samples). "-" indicates not detected.}
\label{tab:small_data_r}
\begin{threeparttable}
\resizebox{\textwidth}{!}{
\setlength{\tabcolsep}{0.65mm}{
\begin{tabular}{p{2cm}p{1cm}<{\centering}p{1cm}<{\centering}p{1cm}<{\centering}p{1cm}<{\centering}p{1cm}<{\centering}p{1cm}<{\centering}p{1cm}<{\centering}p{1cm}<{\centering}p{1cm}<{\centering}p{1cm}<{\centering}p{1cm}<{\centering}p{1cm}<{\centering}p{1cm}<{\centering}p{1cm}<{\centering}}
\toprule
& & \multicolumn{4}{c}{\textbf{Damage}} & \multicolumn{8}{c}{\textbf{Component}} \\
\cmidrule(lr){3-6} \cmidrule(lr){7-14}
\textbf{Model} & \textbf{Scale} & \textbf{MTL} & \textbf{CRC} & \textbf{GWA} & \textbf{SWA} & \textbf{BND} & \textbf{SLE} & \textbf{BRN} & \textbf{TEE} & \textbf{CAS} & \textbf{VAL} & \textbf{ESP} & \textbf{FLA} & \textbf{Overall}  \\
\midrule
YOLOv8-m& 1 &0.390  &0.401  &0.112  &0.221  &0.880  &0.900  &0.508  &0.820  &0.640  &1.000  &0.778  & 0.500 & 0.596\\
YOLOv8-m& 1/2 & 0.457 & 0.380 & 0.143 & 0.272 & 0.827 & 0.914 & 0.489 & 0.711 & 0.600 & 1.000 & 0.444 & 0.250 & 0.541\\
YOLOv8-m& 1/4 &0.443  & 0.392  & 0.149 & 0.281 & 0.782 & 0.897 & 0.333 & 0.579 & 0.480 & 1.000 &0.667  & 0.750 &0.563 \\
YOLOv8-m& 1/8 & 0.402  & 0.134  & 0.135 & 0.193 & 0.565 & 0.793 & 0.311 & 0.368 & 0.240 & 0.913  & 0.333  &  0.500 & 0.407 \\
YOLOv8-m& 1/16 & 0.417  & 0.209  & 0.105 & 0.191 & 0.437 & 0.845 & - & 0.053 & - & 0.870  &-  & 0.250 & 0.281\\
RF-DETR-Base& 1 & 0.553 & 0.390 & 0.236 & 0.446 & 0.852 & 0.879 & 0.644 & 0.895 & 0.600  &1.000  & 0.556 &0.750  & 0.650\\ 
RF-DETR-Base& 1/2 & 0.534  &0.427   & 0.232 & 0.387 & 0.797 &  0.845& 0.556 & 0.947 & 0.680 & 1.000  & 0.889  & 0.500 &0.650\\
RF-DETR-Base& 1/4 & 0.535  & 0.450  & 0.247 & 0.425 & 0.776 & 0.879 & 0.622 & 0.842 & 0.720 & 0.826  & 0.778  & 0.500 &0.633\\
RF-DETR-Base& 1/8 & 0.503 & 0.418  & 0.174 & 0.408 & 0.573 & 0.793 & 0.556 & 0.737 & 0.320 & 0.870  & 0.778  & 0.500 & 0.552\\
RF-DETR-Base& 1/16 & 0.451 & 0.310 & 0.139 & 0.331 & 0.406 & 0.793 & 0.533 & 0.632 & 0.360 & 0.913  & 0.333  & - & 0.433\\
YOLO26-m& 1  & 0.322 & 0.368 & 0.059 & 0.169 & 0.748 & 0.862 & 0.467 & 0.816 & 0.480 & 1.000 & 0.556 & 0.750 & 0.550\\ 
YOLO26-m& 1/2 & 0.363 & 0.340  & 0.067 & 0.186 & 0.701 & 0.897 & 0.422 & 0.816 & 0.520 & 1.000 & 0.667 & 0.500 & 0.540\\
YOLO26-m& 1/4 & 0.394  & 0.303 & 0.089 & 0.193 & 0.723 & 0.879 & 0.378 & 0.632 & 0.320 & 0.957  & 0.557  & 0.500 & 0.494\\
YOLO26-m& 1/8  & 0.354 & 0.094 & 0.036 & 0.139 & 0.466 & 0.724 & 0.089 & - & 0.200  & 0.870  & - & - & 0.248\\
YOLO26-m& 1/16 & 0.361 & 0.061 & 0.019 & 0.119 & 0.044 & 0.586 & - & - &-  & 0.565  & - & 0.250 & 0.167\\

\bottomrule
\end{tabular}}}
\end{threeparttable}
\end{table*}

\begin{table*}[!t]
\centering
\caption{Comparison of F1-score for each category in models with different dataset scales (scale=1: 53,375 Positive + 5,930 Negative samples). "-" indicates not detected.}
\label{tab:small_data_f1}
\begin{threeparttable}
\resizebox{\textwidth}{!}{
\setlength{\tabcolsep}{0.65mm}{
\begin{tabular}{p{2cm}p{1cm}<{\centering}p{1cm}<{\centering}p{1cm}<{\centering}p{1cm}<{\centering}p{1cm}<{\centering}p{1cm}<{\centering}p{1cm}<{\centering}p{1cm}<{\centering}p{1cm}<{\centering}p{1cm}<{\centering}p{1cm}<{\centering}p{1cm}<{\centering}p{1cm}<{\centering}p{1cm}<{\centering}}
\toprule
& & \multicolumn{4}{c}{\textbf{Damage}} & \multicolumn{8}{c}{\textbf{Component}} \\
\cmidrule(lr){3-6} \cmidrule(lr){7-14}
\textbf{Model} & \textbf{Scale} & \textbf{MTL} & \textbf{CRC} & \textbf{GWA} & \textbf{SWA} & \textbf{BND} & \textbf{SLE} & \textbf{BRN} & \textbf{TEE} & \textbf{CAS} & \textbf{VAL} & \textbf{ESP} & \textbf{FLA} & \textbf{Overall}  \\
\midrule
YOLOv8-m& 1& 0.395 &0.174 &0.168 &0.243 &0.596 &0.747 &0.554 &0.749 &0.434 &0.882 &0.609 &0.333 &0.490\\
YOLOv8-m& 1/2 & 0.387 & 0.139 & 0.188 & 0.249 & 0.569 & 0.679 & 0.500 & 0.628 & 0.526 & 0.836& 0.267 & 0.250 & 0.358\\
YOLOv8-m& 1/4 &0.334  & 0.111  &0.169  & 0.194 & 0.553 & 0.693 & 0.395 & 0.611 & 0.387 &0.868  &0.522  & 0.300 & 0.428\\
YOLOv8-m& 1/8 & 0.312  & 0.038  & 0.139 & 0.199& 0.337 & 0.590 & 0.444 & 0.424 & 0.279 & 0.750  & 0.250  & 0.444 & 0.351\\
YOLOv8-m& 1/16 & 0.305  &0.051   &0.115  & 0.172 & 0.208 & 0.616 & - & 0.100 & - & 0.615 & -  & 0.143 & 0.194\\
RF-DETR-Base& 1  &0.256  & 0.169 & 0.200 & 0.112 & 0.472 & 0.548 & 0.574 & 0.782 & 0.429 & 0.821 & 0.400 & 0.750 & 0.460\\ 
RF-DETR-Base& 1/2 &0.260  & 0.143  &0.180  &0.124  & 0.435 & 0.490 & 0.439 & 0.857 & 0.340 & 0.793  & 0.762  & 0.174 &0.416\\
RF-DETR-Base& 1/4 & 0.238 & 0.101  & 0.168 & 0.110 & 0.378 & 0.667 & 0.491 & 0.744 & 0.419 & 0.776  & 0.636  & 0.163 &0.408\\
RF-DETR-Base& 1/8 & 0.229 & 0.109  & 0.155 & 0.140 & 0.292 & 0.409 & 0.385 & 0.609 & 0.372 & 0.755  & 0.437  & 0.098 & 0.332\\
RF-DETR-Base& 1/16 & 0.227  & 0.100  & 0.134 & 0.111 & 0.269 & 0.532 & 0.421 & 0.615 & 0.200 & 0.341  & 0.333  & - & 0.274\\
YOLO26-m& 1 & 0.377 & 0.219 & 0.103 & 0.237 & 0.567 & 0.820 & 0.553 & 0.838  & 0.585 & 0.821 & 0.455 & 0.857 & 0.536\\
YOLO26-m& 1/2 & 0.384  & 0.155  & 0.113 & 0.248 & 0.534 & 0.832 & 0.500 & 0.795 & 0.565 & 0.807  & 0.462  & 0.571 & 0.497\\
YOLO26-m& 1/4 & 0.376 & 0.122 & 0.132 & 0.235 & 0.544 & 0.654 & 0.430 & 0.706 & 0.390 & 0.786  & 0.400  & 0.364 & 0.428\\
YOLO26-m& 1/8  & 0.359 & 0.069 & 0.065 & 0.196 & 0.420 & 0.609 & 0.157 & - & 0.313  & 0.678  & - & - & 0.239\\
YOLO26-m& 1/16 & 0.327 & 0.051 & 0.034 & 0.138 & 0.075 & 0.660 & - & - &-  & 0.419  & - & 0.041 & 0.145\\

\bottomrule
\end{tabular}}}
\end{threeparttable}
\end{table*}

\subsubsection{Per-Category Analysis (Damages and Components)}
\paragraph{\textbf{Damage-Type Features (MTL, CRC, GWA, SWA)}}

The damage categories present distinct detection challenges due to extreme scale variation and category imbalance.

Metal Loss (MTL): As the dominant category, MTL shows surprising detection difficulty. YOLOv5-l achieves the highest AP50 (\textbf{0.286}), outperforming transformer alternatives (RF-DETR: 0.246, RT-DETR: 0.202). This suggests that anchor-based detection remains superior for densely packed micro-defects. Notably, despite MTL's high frequency, no model achieves AP50 > 0.30, indicating inherent challenges in detecting sparse, low-contrast pitting signals against noisy backgrounds.

Corrosion Cluster (CRC): YOLO26-m achieves the best AP50 (\textbf{0.149}), followed by YOLO26-x (\underline{0.139}).Most models exhibit higher recall than precision, suggesting that the localization ability of CRC suffers from poorly-defined boundaries and confusion with densely packed MTL. Models appear to detect the general area of corrosion clusters effectively (high recall), but struggle to precisely delineate their spatial extent against scattered metal loss.

Weld Anomalies (GWA and SWA): F1-scores generally improve with increasing model complexity across the YOLOv5, YOLOv8 and YOLO26 series, with YOLO11 being a notable exception. Among the 12 categories, GWA presents the greatest detection challenge (maximum AP50 of merely \textbf{0.076}), as it occurs exclusively on girth welds that exhibit substantial signal variations within the heat-affected zone, thereby obscuring anomaly boundaries. This thermal interference creates unstable feature representations even when metal loss is present, rendering GWA particularly elusive for detectors relying on consistent visual patterns. SWA constitutes another challenging category, as its inclined linear patterns closely resemble those of MTL and the two frequently co-occur within the same spatial regions, further complicating differentiation and localization.

\paragraph{\textbf{Component-Type Features (BND, SLE, BRN, TEE, CAS, VAL, ESP, FLA)}}

Component detection reveals a scale-dependent performance gap:
Large Continuous Components (SLE, VAL, BND): These full-circumference features achieve high AP50 across most models (SLE:\textbf{0.880}, VAL:\textbf{0.919}, BND:\textbf{0.587}), validating that modern detectors handle large structural elements reliably. However, RT-DETR completely fails in SLE (AP50: “-”), likely due to its decoder's inability to model continuous vertical structures spanning the full circumferential boundary. 

Moderate-size Components (BRN and TEE):  These exhibit high inter-class confusion. RF-DETR achieves the highest BRN AP50 (\textbf{0.479}) and strong BRN and TEE recall (\textbf{0.644}, \underline{0.895}), but YOLO models (e.g., YOLO26-n) excel in TEE and BRN precision (\underline{0.739}, \underline{0.889}). This discrepancy suggests that transformer architectures better capture global spatial priors, such as the azimuthal constraints of BRN and TEE (concentrated at 0, 3 and 9 o'clock positions) and their contextual association with station scenes, thereby successfully retrieving more objects despite local feature ambiguity. Conversely, CNN-based detectors excel at precise geometric localization: YOLO models more accurately delineate the spherical boundaries of TEE and the spindle-shaped contours of BRN through localized edge and texture cues, resulting in tighter bounding boxes and fewer false positives on ambiguous background regions.

Long-tailed Components (FLA, ESP, CAS): These rare tail categories (combined <0.1\% of objects) expose the brittleness of current approaches. Several models achieve relatively high recall (\textbf{0.750}) in FLA, but the precision is low. This arises because FLA patterns are visually similar to partial VAL segments. During patch-based processing, VAL frequently extends beyond single windows due to their large size, producing incomplete observations that geometrically mimic FLA. Specifically, the blue rectangular base with a single vertical red line of FLA resembles truncated VAL segments that lack their complete symmetric elliptical patterns. This spatial truncation of VAL during windowing creates ambiguous local features that are easily misclassified as FLA, thus degrading detection precision for both categories. Models in the YOLO family maintain moderate performance on CAS and ESP, which indicates that YOLO architectures maintain representational capacity for tail categories even when training data follow a highly skewed distribution.

\subsubsection{Visualization Comparison}
\ 
\newline
Fig.~\ref{fig:visual_results_1_a}-Fig.~\ref{fig:visual_results_3_c} provide more qualitative visualization results. 
Fig.~\ref{fig:visual_results_1_a}-Fig.~\ref{fig:visual_results_1_c} contain dense damage-type categories. Transformer-based models RF-DETR and RT-DETR, as well as RetinaNet detect much more metal losses, corresponding to their high recall values shown in Table~\ref{tab:result_full_recall}, while the detection results are duplicated and located at the wrong region, leading to low precision, these models can detect more head-class objects, but false positives are also higher. In contrast, YOLO series has balanced ability to detect and locate, YOLOv8-l outperforms among all YOLO models on this case.

Fig.~\ref{fig:visual_results_2_a}-Fig.~\ref{fig:visual_results_2_c} contain multiple types of component categories. YOLO-series and RF-DETR detect most of the components correctly, especially VAL and SLE. They also identify one MTL object with an abnormal color pattern. Several models successfully figure out hard-to-detect BRNs, such as YOLOv5-m, YOLOV8-m, YOLO11-l, RT-DETR and RF-DETR.

Fig.~\ref{fig:visual_results_3_a}-Fig.~\ref{fig:visual_results_3_c} contain tiny and low-response MTLs. Most models can detect these tiny damages, indicating that current models can handle small objects after being trained with large-scale data.
\subsection{Data Scaling Study}
\label{sec:detail_small_data}
Tables~\ref{tab:small_data_ap50}--~\ref{tab:small_data_f1} show per-category metrics (AP50, Precision, Recall, F1-score) of models trained with different subsets. Key observations include: (i) The recall of head categories (predominantly damage types) remains stable across different data scales, attributable to the sufficient presence of these frequent objects even in small subsets. Conversely, precision for head categories improves with larger training sets, indicating enhanced localization accuracy as more data become available. (ii) For medium-to-tail categories, both precision and recall generally increase with data scale, though the rarest categories (ESP, FLA) exhibit instability due to their extreme scarcity in the testing set. (iii) Overall performance improves consistently as the training subset approaches the full dataset, with YOLO26-m showing particularly pronounced sensitivity to data volume. (iv) Notably, all models exhibit substantial gains when scaling from 1/8 to 1/4 of the data, whereas the marginal improvement rate decreases at higher scales (1/4 $\rightarrow$ 1/2 $\rightarrow$ 1) because at these stages, the model has possessed an adequate detection performance.

\subsection{Zero-shot Generalization Experiment}
\label{sec:detail_zero_test}
Tables~\ref{tab:result_full_ap50_zero}--~\ref{tab:result_full_f1_zero} show per-category metrics (AP50, Precision, Recall, F1-score) of trained models on the Zero-shot Set. Key observations include: (i) Most models achieve relatively stable performance compared to that on PipeMFL-240K Testing Set, indicating reasonable cross-pipeline generalization. (ii) Damage-type categories (MTL, CRC, GWA, SWA) exhibit performance degradation, attributable to more complex pipeline conditions and longer operation period of the zero-shot pipeline. (iii) Conversely, several models (predominantly YOLO variants) achieve higher metrics on middle-to-tail categories, demonstrating that real-time one-stage detectors are suitable for pipeline inspection. (iv) YOLO-World series again perform the best, consistent with observations on PipeMFL-240K Testing Set. (v) As model size increases (YOLO-n $\rightarrow$ x), performance improves simultaneously, indicating that larger models with more parameters are beneficial for such complex detection scenarios. (vi) DETR models are less competitive; several categories remain undetected, especially tail categories—a phenomenon also observed with lightweight YOLO-n and YOLO-s models, suggesting that limited parameters hinder feature learning for rare categories.

\subsection{Sampling Ratio Selection}
\label{sec:detail_sampling}
Tables~\ref{tab:sample_240}--\ref{tab:sample_zero_f1} present comprehensive results under varying positive-to-negative sampling ratios on both the standard and zero-shot testing sets. Several key observations include: (i) Model capacity serves as an effective buffer against negative-sample inflation, with larger models exhibiting amplified relative advantages under extreme sampling ratios. As model complexity increases, nearly all metrics improve monotonically across both datasets. Notably, under the extreme 3:7 sampling ratio, the relative performance gap between large and small models does not shrink but rather expands (e.g., on standard testing set, the macro F1 margin between YOLOv8-l and YOLOv8-s widens from 0.092 at 9:1 to 0.102 at 3:7), indicating that greater model capacity not only yields absolute gains but also enhances robustness to deteriorating positive-to-negative ratios. (ii) Increasing negative samples triggers a conservative decision behavior within models: micro Precision rises at the cost of catastrophic Recall declines, leading to comprehensive metric degradation. As the negative-sample proportion grows, models increasingly suppress false positives, evidenced by monotonically rising micro Precision. However, this risk-averse behavior incurs severe missed-detection penalties, with macro Recall dropping precipitously  and consequently dragging mAP and F1 down uniformly. This confirms that a "rather miss than misclassify" strategy is counterproductive in detection tasks, as the cost of false negatives substantially outweighs that of false positives. (iii) Category performance exhibits a "Matthew effect" and collapse cascade: tail categories are progressively sacrificed while dominant Component classes remain resilient. Performance degradation under negative-sample pressure is highly non-uniform. Component categories with salient, large-scale features, particularly VAL and SLE, maintain AP50 predominantly above 0.8 across all sampling ratios. Conversely, tail categories suffer disproportionately, with their AP50 collapsing toward zero and eventually vanishing from predictions  as negative samples accumulate. The model strategically abandons hard-to-detect tail classes to preserve overall Precision, creating a widening performance chasm between dominant and tail categories.

\begin{table*}[!t]
\centering
\caption{Model comparison on PipeMFL-240K Zero-shot Set (Confidence score=0.25, IoU=0.5). The best results in each column are highlighted in bold and the second-best values are underlined.}
\label{tab:extra_ap_zero}
\begin{threeparttable}
\setlength{\tabcolsep}{0.7mm}{  
\begin{tabular}{p{3.2cm}@{\centering} 
               p{1.2cm}<{\centering}
               p{1.5cm}<{\centering}
               p{1.5cm}<{\centering}
               p{2.4cm}<{\centering}
               p{2.4cm}<{\centering}
               p{2.4cm}<{\centering}
               p{1cm}<{\centering}
               p{1cm}<{\centering}
}
\toprule
\textbf{Model} & \textbf{Year} & \textbf{mAP50} & \textbf{mAP50:95} & \textbf{Precision} & \textbf{Recall} & \textbf{F1-score} & \textbf{Param} & \textbf{FLOPs}  \\
 &  &  &  & \textbf{(macro, micro)} & \textbf{(macro, micro)} & \textbf{(macro, micro)} & \textbf{[M]} & \textbf{[G]}  \\
\midrule 
Faster R-CNN~\cite{2015FASTRCNN} & 2017 & 0.078 & 0.041 & 0.035, 0.004 &0.214, 0.029  & 0.053, 0.007 & 41.4 & 535.9   \\
RetinaNet~\cite{retina} & 2017 & 0.071 & 0.055 & 0.076, 0.100 & 0.198, \textbf{0.455} & 0.069, 0.164 & 32.4 &  519.8  \\
YOLOv5-s~\cite{yolov5}     & 2020 &0.369  &0.216  &0.354, 0.195  &0.532, 0.361  &0.389, 0.253  & 9.1  &23.9    \\
YOLOv5-m~\cite{yolov5}   & 2020 & 0.438 & 0.240 & 0.399, 0.185 & \underline{0.661}, 0.347 & 0.460, 0.241 & 25.1 & 64.0   \\
YOLOv5-l~\cite{yolov5}   & 2020 & 0.413 & 0.249 & 0.359,  0.196 & 0.619, 0.402 & 0.414, 0.264 & 53.1 & 134.7  \\
YOLOv8-s~\cite{yolov8_ultralytics}   & 2023 & 0.347 & 0.204 &0.354, 0.165 & 0.432, 0.403& 0.367, 0.234 & 11.1&28.5    \\
YOLOv8-m~\cite{yolov8_ultralytics}   & 2023 &0.442  &0.255  &0.352, 0.155  &0.630, 0.350  &0.407, 0.216  &25.9  & 78.7    \\
YOLOv8-l~\cite{yolov8_ultralytics}   & 2023 & 0.411&0.251 &0.424, 0.200 & 0.482, 0.409 &0.445, 0.268 & 43.6& 164.9    \\
YOLO11-s~\cite{yolo11}   & 2024 &0.340 &0.190 & 0.361, 0.187 & 0.494, 0.373 & 0.339, 0.249 & 9.4 & 21.3  \\
YOLO11-m~\cite{yolo11}   & 2024 & 0.366 & 0.205 & 0.456, 0.237 & 0.439, 0.286 & 0.432, 0.259 & 20.0  &  67.7  \\
YOLO11-l~\cite{yolo11}   & 2024 & 0.450 & 0.239 & 0.398, 0.165 & 0.610, 0.366 & 0.441, 0.227 & 25.3  &  86.6 \\
YOLOv8-s-world-v2~\cite{2024yoloworld}   & 2024 & 0.471 & 0.278 & 0.355, 0.163 &0.568, 0.377 & 0.415, 0.228 & 12.7 & 34.7   \\
YOLOv8-m-world-v2~\cite{2024yoloworld}   & 2024 & \textbf{0.516}  &\textbf{0.300}  &0.407, 0.220  &\textbf{0.666}, 0.339  &0.464, 0.266  & 28.4 & 88.8   \\
YOLOv8-l-world-v2~\cite{2024yoloworld}   & 2024 & \underline{0.477} & 0.266 & 0.431, 0.169 & 0.634, \underline{0.410} & \underline{0.478}, 0.239  & 46.8 & 179.0  \\
YOLOv8-x-world-v2~\cite{2024yoloworld}   & 2024 & 0.475 & \underline{0.285} & \underline{0.460}, \textbf{0.236} & 0.608, 0.358& \textbf{0.481}, \textbf{0.284} & 72.9 &  277.4 \\
RT-DETR~\cite{2024RTDETR}  & 2024 & 0.127 & 0.067 & 0.146, 0.063 & 0.206, 0.407 & 0.140, 0.109 & 32.0 & 103.5   \\
RF-DETR-Base~\cite{rfdetr} & 2025 & 0.277 & 0.190 & 0.198, 0.054 & 0.482, 0.329 & 0.237, 0.093 & 28.6 & 89.3 \\
YOLO26-n~\cite{yolo26}   & 2026 & 0.304 & 0.186 & 0.332, 0.213 & 0.375, 0.327 & 0.331, 0.258 & 2.4& 9.5  \\
YOLO26-s~\cite{yolo26}   & 2026 & 0.307 & 0.211 & 0.445, 0.188 & 0.384, 0.350 & 0.399, 0.244 & 9.5 & 22.5  \\
YOLO26-m~\cite{yolo26}   & 2026 & 0.392 & 0.227 & 0.395, 0.194 & 0.525, 0.315 & 0.411, 0.240 & 9.5 & 22.5  \\
YOLO26-l~\cite{yolo26}   & 2026 & 0.208 & 0.147 & 0.430, 0.201 & 0.275, 0.361 & 0.305, 0.260 & 24.8 & 86.1 \\
YOLO26-x~\cite{yolo26}   & 2026 & 0.410 & 0.249 & \textbf{0.475}, \underline{0.229}& 0.609, 0.346 & 0.476, \underline{0.275} & 55.6 & 193.4  \\
\bottomrule
\end{tabular}}
\end{threeparttable}
\end{table*}

\begin{table*}[!t]
\centering
\caption{Comparison of AP50 for each category in different models on PipeMFL-240K Zero-shot Set. The best results in each column are highlighted in bold and the second-best values are underlined. "-" indicates not detected.}
\label{tab:result_full_ap50_zero}
\begin{threeparttable}
\resizebox{\textwidth}{!}{
\setlength{\tabcolsep}{0.65mm}{
\begin{tabular}{p{3.3cm}p{1cm}<{\centering}p{1cm}<{\centering}p{1cm}<{\centering}p{1cm}<{\centering}p{1cm}<{\centering}p{1cm}<{\centering}p{1cm}<{\centering}p{1cm}<{\centering}p{1cm}<{\centering}p{1cm}<{\centering}p{1cm}<{\centering}p{1cm}<{\centering}p{1cm}<{\centering}p{1cm}<{\centering}}
\toprule
& \multicolumn{4}{c}{\textbf{Damage}} & \multicolumn{8}{c}{\textbf{Component}} \\
\cmidrule(lr){2-5} \cmidrule(lr){6-13}
\textbf{Model} & \textbf{MTL} & \textbf{CRC} & \textbf{GWA} & \textbf{SWA} & \textbf{BND} & \textbf{SLE} & \textbf{BRN} & \textbf{TEE} & \textbf{CAS} & \textbf{VAL} & \textbf{ESP} & \textbf{FLA} & \textbf{Overall}  \\
\midrule 
Faster R-CNN~\cite{2015FASTRCNN} & - & 0.021 & 0.001 & 0.004 & 0.254 & 0.074 & 0.182 & 0.143 & - & - & - & - & 0.078 \\
RetinaNet~\cite{retina} & 0.198 & 0.028 & 0.025 & 0.037 & 0.061 & 0.500 & -  & - & - & - & - & - & 0.071\\
YOLOv5-s~\cite{yolov5}& 0.141 &\underline{0.076} &0.029 &0.029 &0.119 &\textbf{1.000} &0.333 &0.563 &0.143 &\textbf{1.000} & - &0.889 &0.369 \\
YOLOv5-m~\cite{yolov5}  & 0.131 & 0.064 & 0.027 & 0.028 & 0.266 & \textbf{1.000} & 0.410 & \textbf{1.000} & 0.050 & \textbf{1.000} & 0.333 & 0.950 & 0.438 \\
YOLOv5-l~\cite{yolov5} & \textbf{0.215} & 0.022 & 0.019 & 0.071 & 0.346 & \textbf{1.000} & 0.525 & 0.844 & 0.100 & 0.750 & 0.500 & 0.563 & 0.413\\
YOLOv8-s~\cite{yolov8_ultralytics} &0.158 &0.012 &0.019 &0.054 &0.257 &\textbf{1.000} &0.377 &0.283 &- &0.889 &- &\textbf{1.000} &0.347\\
YOLOv8-m~\cite{yolov8_ultralytics} &0.132 &\textbf{0.078} &0.016 &0.042 &0.287 &\textbf{1.000} &0.485 &0.542 &0.100 &\textbf{1.000} &\textbf{1.000} &0.625 &0.442\\
YOLOv8-l~\cite{yolov8_ultralytics} & \underline{0.204} &0.046 &0.023 &\textbf{0.088} &\textbf{0.447} & \textbf{1.000} &0.505 &\underline{0.875} & - &\textbf{1.000} &- & \textbf{1.000}&0.411\\
YOLO11-s~\cite{yolo11} &0.148 & 0.035& 0.024& 0.048& 0.210& \textbf{1.000}& 0.250&0.250 &0.111 &\textbf{1.000} &- & \textbf{1.000}&0.340\\
YOLO11-m~\cite{yolo11} & 0.108 & 0.030 & - & 0.057 & 0.316 & \textbf{1.000} & 0.454 & 0.344 & - & \textbf{1.000} & \textbf{1.000} & 0.083 & 0.366\\
YOLO11-l~\cite{yolo11} & 0.140 &0.037 &0.004 &0.058 & 0.238& \textbf{1.000}& 0.496&0.325 & 0.100& \textbf{1.000}& \textbf{1.000}& \textbf{1.000}& 0.450\\
YOLOv8-s-world-v2~\cite{2024yoloworld} & 0.155 & 0.044 & 0.015 & 0.038 & 0.263 & \textbf{1.000} & \textbf{0.573} & 0.859 &  - & 0.950 & \textbf{1.000} & 0.750 & 0.471 \\
YOLOv8-m-world-v2~\cite{2024yoloworld} & 0.172 & 0.043 & 0.017 & 0.057 & 0.367 & \textbf{1.000} & \underline{0.551} & 0.844 &  0.143 & \textbf{1.000} & \textbf{1.000} & \textbf{1.000} & \textbf{0.516} \\
YOLOv8-l-world-v2~\cite{2024yoloworld} & 0.196 & 0.043 & 0.010 & 0.072 & 0.338 & \textbf{1.000} & 0.517 & 0.594 &  \textbf{0.200} & 0.750 & \textbf{1.000} & \textbf{1.000} & \underline{0.477} \\
YOLOv8-x-world-v2~\cite{2024yoloworld} & 0.198 & 0.009 & \textbf{0.039} & \underline{0.083} & 0.336 & \textbf{1.000} & 0.488 & 0.594 &  \textbf{0.200} & \textbf{1.000} & \textbf{1.000} & 0.750 & 0.475 \\
RT-DETR~\cite{2024RTDETR} & 0.152 & - & 0.031 & 0.071 & - & - & 0.396 & \underline{0.875} & - & - & - & - & 0.127\\
RF-DETR-Base~\cite{rfdetr} & 0.073 & 0.068 & 0.032 & 0.010 & 0.071 & \textbf{1.000} & 0.421 & 0.542 & 0.111 &\textbf{1.000}  & - &  - & 0.277\\ 
YOLO26-n~\cite{yolo26}  & 0.128 & 0.046 & 0.024 & 0.045 & 0.175 & \textbf{1.000} & 0.235 & - & - & \textbf{1.000} & - & \textbf{1.000} & 0.304\\ 
YOLO26-s~\cite{yolo26}   & 0.138 &0.044  & 0.007 & 0.063 & 0.286 & \textbf{1.000} & 0.488 & 0.344 & - & 0.750 & - & 0.563 & 0.307\\ 
YOLO26-m~\cite{yolo26}   & 0.106 & 0.064 & \underline{0.036} & 0.038 & 0.166 & \textbf{1.000} & 0.465 &0.625  & - & \textbf{1.000} & 0.200 & \textbf{1.000} & 0.392\\ 
YOLO26-l~\cite{yolo26} & 0.144 & 0.031 & 0.001 & 0.061 & 0.175 & \textbf{1.000} & 0.083 & 0.125 & - & 0.750 & - & 0.125 & 0.208\\ 
YOLO26-x~\cite{yolo26}   & 0.179 &  0.007 &  0.003& 0.076 & \underline{0.424} & \textbf{1.000} & 0.424 & 0.583 & 0.125 & \textbf{1.000} & 0.100 & \textbf{1.000} & 0.410\\  
\bottomrule
\end{tabular}}}
\end{threeparttable}
\end{table*}

\begin{table*}[!t]
\centering
\caption{Comparison of Precision for each category in different models on PipeMFL-240K Zero-shot Set. The best results in each column are highlighted in bold and the second-best values are underlined. "-" indicates not detected.}
\label{tab:result_full_p_zero}
\begin{threeparttable}
\resizebox{\textwidth}{!}{
\setlength{\tabcolsep}{0.65mm}{
\begin{tabular}{p{3.3cm}p{1cm}<{\centering}p{1cm}<{\centering}p{1cm}<{\centering}p{1cm}<{\centering}p{1cm}<{\centering}p{1cm}<{\centering}p{1cm}<{\centering}p{1cm}<{\centering}p{1cm}<{\centering}p{1cm}<{\centering}p{1cm}<{\centering}p{1cm}<{\centering}p{1cm}<{\centering}p{1cm}<{\centering}}
\toprule
& \multicolumn{4}{c}{\textbf{Damage}} & \multicolumn{8}{c}{\textbf{Component}} \\
\cmidrule(lr){2-5} \cmidrule(lr){6-13}
\textbf{Model} & \textbf{MTL} & \textbf{CRC} & \textbf{GWA} & \textbf{SWA} & \textbf{BND} & \textbf{SLE} & \textbf{BRN} & \textbf{TEE} & \textbf{CAS} & \textbf{VAL} & \textbf{ESP} & \textbf{FLA} & \textbf{Overall}  \\
\midrule
Faster R-CNN~\cite{2015FASTRCNN} & - & 0.019 & 0.001 & 0.002 & 0.128 &  0.039 & 0.125 & 0.100 & - & - & - & - & 0.035\\
RetinaNet~\cite{retina} & 0.130 &0.124 & 0.022 & 0.021 & 0.500 & 0.111 & - & - & - & - & - & - & 0.076   \\
YOLOv5-s~\cite{yolov5} &0.209 &0.145 &0.064 &0.093 &0.323 &0.333 &0.500 &0.714 &0.071 &\textbf{1.000} & - &\textbf{1.000} &0.354\\
YOLOv5-m~\cite{yolov5}  & 0.196 & 0.156 & 0.044 & 0.094 & 0.354 & 0.500 & 0.546 & \textbf{1.000} & 0.050 & 0.800 & 0.250 & 0.800 & 0.399\\
YOLOv5-l~\cite{yolov5} & 0.205 & 0.174 & 0.049 & 0.114 & 0.515& 0.333 & 0.583 & 0.778 & 0.100 & 0.600 & 0.250 &0.600 & 0.359\\
YOLOv8-s~\cite{yolov8_ultralytics} & 0.171& 0.138& 0.023 &0.098 &0.486 &0.333 & 0.600&0.600 &- &0.800 &- & \textbf{1.000}&0.354\\
YOLOv8-m~\cite{yolov8_ultralytics} &0.172 &0.133 &0.029 &0.063 &0.425 &0.250 &0.533 &0.833 &0.053 &0.800 &0.333 &0.600 &0.352\\
YOLOv8-l~\cite{yolov8_ultralytics} & 0.207&0.128 &0.064 &0.126 &0.556 &\textbf{1.000} & 0.636&0.875 & - & \textbf{1.000} & - & 0.500&0.424\\
YOLO11-s~\cite{yolo11} &0.189 &\textbf{0.234} &0.111 &0.133 &0.432 & 0.200& 0.455& \textbf{1.000}& 0.071& 0.500& -& \textbf{1.000}& 0.361\\
YOLO11-m~\cite{yolo11} & 0.236 & 0.185 & - & \textbf{0.234} & \underline{0.593}  & \textbf{1.000} & 0.636 & 0.750 & - & \textbf{1.000} & \underline{0.500} & 0.333 & 0.456\\
YOLO11-l~\cite{yolo11} & 0.174 & 0.128 & 0.021 & 0.093& 0.390& 0.333& 0.700& 0.600& 0.100& 0.571& \textbf{1.000}& 0.667& 0.398\\
YOLOv8-s-world-v2~\cite{2024yoloworld} & 0.179 & 0.073 & 0.030 & 0.089 & 0.313 & 0.333 & 0.727 & 0.778 & - & 0.800 & 0.333 & 0.600 & 0.355\\
YOLOv8-m-world-v2~\cite{2024yoloworld} & \underline{0.237} & 0.119 & 0.051 & 0.121 & 0.475 & 0.250 & 0.667 & 0.778 & 0.059 & 0.800 & 0.333 & \textbf{1.000} & 0.407\\
YOLOv8-l-world-v2~\cite{2024yoloworld} & 0.175 & 0.180 & 0.030 & 0.105 & 0.462 & 0.333 & 0.636 & 0.750 & \textbf{0.200} & \textbf{1.000} & \underline{0.500} & 0.800 & 0.431\\
YOLOv8-x-world-v2~\cite{2024yoloworld} & \textbf{0.243} & 0.192 & 0.068 & 0.161 & 0.586 & 0.500 & 0.667 & 0.750 & 0.100 & \textbf{1.000} & \underline{0.500} & 0.750 & \underline{0.460}\\
RT-DETR~\cite{2024RTDETR} & 0.087 & - & 0.017 & 0.016 & - & - & \underline{0.750} & 0.875 & - & - & - & - & 0.146 \\
RF-DETR-Base~\cite{rfdetr} & 0.071 & 0.090 & 0.029 & 0.010 & 0.205 & 0.100 & 0.444 & 0.714 & 0.050 &0.667  & - &  - & 0.198\\ 
YOLO26-n~\cite{yolo26}  & 0.218 & 0.198 & \textbf{0.286} & 0.145 & 0.400 & 0.500 & 0.571 & - & - & 0.667 & - & \textbf{1.000} & 0.332\\ 
YOLO26-s~\cite{yolo26}   & 0.188 & 0.196 & \underline{0.154} & 0.154 & 0.536 & \textbf{1.000} & \textbf{0.857} & 0.750 & - & 0.750 & - & 0.750 & 0.445\\ 
YOLO26-m~\cite{yolo26}   & 0.196 & 0.166 & \underline{0.154} & 0.154 & 0.406 & 0.333 & 0.667 & \textbf{1.000} & - & 0.800 & 0.200 & 0.667 & 0.395\\ 
YOLO26-l~\cite{yolo26}    & 0.203 & \underline{0.217} & 0.022 & 0.164 & 0.556 & \textbf{1.000} & 0.500 & \textbf{1.000} & - & \textbf{1.000} & - & 0.500 & 0.430\\ 
YOLO26-x~\cite{yolo26}   & 0.229 & 0.140 &  0.040& \underline{0.207} & \textbf{0.621} & \textbf{1.000} & 0.600 & 0.833 & \underline{0.125} & \textbf{1.000} & 0.100 &0.800  & \textbf{0.475}\\  

\bottomrule
\end{tabular}}}
\end{threeparttable}
\end{table*}

\begin{table*}[!t]
\centering
\caption{Comparison of Recall for each category in different models on PipeMFL-240K Zero-shot Set. The best results in each column are highlighted in bold and the second-best values are underlined. "-" indicates not detected.}
\label{tab:result_full_recall_zero}
\begin{threeparttable}
\resizebox{\textwidth}{!}{
\setlength{\tabcolsep}{0.65mm}{
\begin{tabular}{p{3.3cm}p{1cm}<{\centering}p{1cm}<{\centering}p{1cm}<{\centering}p{1cm}<{\centering}p{1cm}<{\centering}p{1cm}<{\centering}p{1cm}<{\centering}p{1cm}<{\centering}p{1cm}<{\centering}p{1cm}<{\centering}p{1cm}<{\centering}p{1cm}<{\centering}p{1cm}<{\centering}p{1cm}<{\centering}}
\toprule
& \multicolumn{4}{c}{\textbf{Damage}} & \multicolumn{8}{c}{\textbf{Component}} \\
\cmidrule(lr){2-5} \cmidrule(lr){6-13}
\textbf{Model} & \textbf{MTL} & \textbf{CRC} & \textbf{GWA} & \textbf{SWA} & \textbf{BND} & \textbf{SLE} & \textbf{BRN} & \textbf{TEE} & \textbf{CAS} & \textbf{VAL} & \textbf{ESP} & \textbf{FLA} & \textbf{Overall}  \\
\midrule
Faster R-CNN~\cite{2015FASTRCNN} & - & \underline{0.358} & 0.054 & 0.154 & 0.424 & \textbf{1.000} & 0.333 & 0.250 &-  & - & - & - &0.214 \\
RetinaNet~\cite{retina} & \textbf{0.488} & 0.192 & 0.156 & \underline{0.286} & 0.285 & 0.121 & - & \textbf{1.000} & - & - & - & -  &0.198 \\
YOLOv5-s~\cite{yolov5} &0.383  &0.236  &0.143 &0.195  &0.303  &\textbf{1.000}  &0.500  &0.625  &\textbf{1.000} &\textbf{1.000}  &-  &\textbf{1.000}  &0.532 \\
YOLOv5-m~\cite{yolov5} &0.366  &0.243  &0.107 &0.200 &0.515  &\textbf{1.000}  &0.500 &\textbf{1.000}  &\textbf{1.000}  &\textbf{1.000}  &\textbf{1.000}  &\textbf{1.000}  &\underline{0.661} \\
YOLOv5-l~\cite{yolov5} &0.431  &0.086  &0.196 &0.239 &0.515 &\textbf{1.000} &0.583  &0.875  &\textbf{1.000}  & 0.750 &\textbf{1.000}  & 0.750  & 0.619 \\
YOLOv8-s~\cite{yolov8_ultralytics} & 0.434 &0.064 &0.054 &0.245 & 0.515 & \textbf{1.000} &0.500& 0.375& - &\textbf{1.000} & - &\textbf{1.000} &0.432\\
YOLOv8-m~\cite{yolov8_ultralytics} & 0.368 & 0.278 & 0.125 & 0.228 &  0.515 & \textbf{1.000} & \textbf{0.667} & 0.625 & \textbf{1.000} & \textbf{1.000} & \textbf{1.000} & 0.750 & 0.630\\
YOLOv8-l~\cite{yolov8_ultralytics} & 0.434& 0.144& 0.125 &0.271  & \underline{0.606} &\textbf{1.000} & 0.583& 0.875& -&\textbf{1.000} &- &\textbf{1.000} &0.482\\
YOLO11-s~\cite{yolo11} & 0.402 &0.093  &0.071  & 0.211 & 0.485& \textbf{1.000} & 0.417 & 0.250& \textbf{1.000} & \textbf{1.000} & - &\textbf{1.000} & 0.494\\
YOLO11-m~\cite{yolo11} & 0.306 & 0.096 & - & 0.171 &  0.485 & \textbf{1.000} & 0.583 & 0.375 & - & \textbf{1.000} & \textbf{1.000} & 0.250 & 0.439\\
YOLO11-l~\cite{yolo11} & 0.386 &0.185  &0.054  & 0.253 & 0.485 & \textbf{1.000} & 0.583 & 0.375 & \textbf{1.000} & \textbf{1.000} & \textbf{1.000} & \textbf{1.000} & 0.610 \\
YOLOv8-s-world-v2~\cite{2024yoloworld} & 0.398  & 0.214 & 0.232 & 0.231 & 0.455 & \textbf{1.000} & \textbf{0.667}  & 0.875 & - & \textbf{1.000} & \textbf{1.000} & 0.750 & 0.568\\
YOLOv8-m-world-v2~\cite{2024yoloworld} & 0.353  & 0.192 & 0.089 & 0.244 & 0.576 & \textbf{1.000} & \textbf{0.667} & 0.875 & \textbf{1.000} & \textbf{1.000} & \textbf{1.000} & \textbf{1.000} & \textbf{0.666}\\
YOLOv8-l-world-v2~\cite{2024yoloworld} & \underline{0.437}  & 0.173 & 0.125 & 0.246 & 0.546 & \textbf{1.000} & 0.583  & 0.750 & \textbf{1.000} & 0.750 & \textbf{1.000} & \textbf{1.000} & 0.634\\
YOLOv8-x-world-v2~\cite{2024yoloworld} & 0.383 & 0.032 & 0.125 & 0.245 & 0.515 & \textbf{1.000} & 0.500 & 0.750 & \textbf{1.000} & \textbf{1.000} & \textbf{1.000} & 0.750 & 0.608\\
RT-DETR~\cite{2024RTDETR} & 0.429 & - & \underline{0.268} & \textbf{0.395} & - & - & 0.500 & 0.875 & - & - & - & - & 0.206 \\
RF-DETR-Base~\cite{rfdetr} & 0.339 & 0.332 & \textbf{0.339} & 0.206 & 0.273 & \textbf{1.000} & \textbf{0.667} & 0.625 & \textbf{1.000}  &\textbf{1.000}  & - & -  & 0.482\\ 
YOLO26-n~\cite{yolo26}  & 0.348 & 0.153 & 0.036 & 0.204 & 0.424 & \textbf{1.000} & 0.333 & - & - & \textbf{1.000} & - & \textbf{1.000} & 0.375\\ 
YOLO26-s~\cite{yolo26}  & 0.332 & 0.150 & 0.036 & 0.222 & 0.455 & \textbf{1.000} & 0.500 & 0.375 & - & 0.750 & - & 0.750 & 0.384\\ 
YOLO26-m~\cite{yolo26}   & 0.332 & 0.220 & 0.036 & 0.191 & 0.394 & \textbf{1.000} & 0.500 & 0.625 & - & \textbf{1.000} & \textbf{1.000} & \textbf{1.000} & 0.525\\ 
YOLO26-l~\cite{yolo26}    & 0.391 & 0.096 & 0.018  & 0.204 & 0.303 &\textbf{1.000}  & 0.167 & 0.125 & - & 0.750 & - & 0.250  &  0.275\\ 
YOLO26-x~\cite{yolo26}   &0.373  & 0.038 & 0.018 & 0.205 & 0.546 & \textbf{1.000} & 0.500 & 0.625 & \textbf{1.000} & \textbf{1.000} & \textbf{1.000} & \textbf{1.000} & 0.609\\  
\bottomrule
\end{tabular}}}
\end{threeparttable}
\end{table*}

\begin{table*}[!t]
\centering
\caption{Comparison of F1-score for each category in different models on PipeMFL-240K Zero-shot Set. The best results in each column are highlighted in bold and the second-best values are underlined. "-" indicates not detected.}
\label{tab:result_full_f1_zero}
\begin{threeparttable}
\resizebox{\textwidth}{!}{
\setlength{\tabcolsep}{0.65mm}{
\begin{tabular}{p{3.3cm}p{1cm}<{\centering}p{1cm}<{\centering}p{1cm}<{\centering}p{1cm}<{\centering}p{1cm}<{\centering}p{1cm}<{\centering}p{1cm}<{\centering}p{1cm}<{\centering}p{1cm}<{\centering}p{1cm}<{\centering}p{1cm}<{\centering}p{1cm}<{\centering}p{1cm}<{\centering}p{1cm}<{\centering}}
\toprule
& \multicolumn{4}{c}{\textbf{Damage}} & \multicolumn{8}{c}{\textbf{Component}} \\
\cmidrule(lr){2-5} \cmidrule(lr){6-13}
\textbf{Model} & \textbf{MTL} & \textbf{CRC} & \textbf{GWA} & \textbf{SWA} & \textbf{BND} & \textbf{SLE} & \textbf{BRN} & \textbf{TEE} & \textbf{CAS} & \textbf{VAL} & \textbf{ESP} & \textbf{FLA} & \textbf{Overall}  \\
\midrule
Faster R-CNN~\cite{2015FASTRCNN} &-  & 0.037 & 0.002 & 0.004 & 0.197 & 0.074 & 0.182 & 0.143 & - &-  &-  & - & 0.053\\
RetinaNet~\cite{retina} & 0.205 & 0.151 & 0.040 & 0.040 & 0.195 & 0.200 & - & - & - & - & - & - & 0.069\\
YOLOv5-s~\cite{yolov5} &0.270 &0.180 &\textbf{0.088} &0.126 &0.312 &0.500 &0.500 &0.667 &0.133 &\textbf{1.000} & -&0.889 &0.389\\
YOLOv5-m~\cite{yolov5} &0.255 & \textbf{0.190} &0.062 &0.128 &0.420 &0.667 &0.522 &\textbf{1.000} &0.095 &0.889 &0.400 &0.889 &0.460\\
YOLOv5-l~\cite{yolov5} & 0.278 &0.115 & 0.079 & 0.155 &0.515 &0.500 & 0.583 &0.824 &0.182 & 0.667 & 0.400 &0.667 &0.414\\
YOLOv8-s~\cite{yolov8_ultralytics} & 0.246 &0.087 &0.032  & 0.140 & 0.500& 0.500 & 0.545 & 0.462& - & 0.889 & - &\textbf{1.000} & 0.367\\
YOLOv8-m~\cite{yolov8_ultralytics}& 0.234 &0.180 &0.047 &0.099 &0.466 &0.400 &0.593 &0.714 &0.100 &0.889 &0.500 &0.667 &0.407\\
YOLOv8-l~\cite{yolov8_ultralytics} & 0.281  & 0.136 & 0.085 & 0.172 & \underline{0.580} & \textbf{1.000} &0.609  & 0.875 & - & \textbf{1.000} & -  & \textbf{1.000} & 0.445\\
YOLO11-s~\cite{yolo11} & 0.257 & 0.133 & 0.087 & 0.163 &0.457 &0.333 & 0.435 & 0.400 & 0.133& 0.667 & - & \textbf{1.000} & 0.339 \\
YOLO11-m~\cite{yolo11} & 0.266 & 0.126 & - & \underline{0.197} & 0.533 & \textbf{1.000} & 0.609 & 0.500 & - & \textbf{1.000} & \underline{0.667} & 0.286 & 0.432 \\
YOLO11-l~\cite{yolo11} & 0.240 & 0.152 & 0.030 & 0.136 &  0.432& 0.500 & 0.636 & 0.462 & 0.182 & 0.727 & \textbf{1.000} &  0.800& 0.441\\
YOLOv8-s-world-v2~\cite{2024yoloworld} & 0.247 & 0.109 & 0.053 & 0.129 & 0.370 & 0.500 & \textbf{0.696} & 0.824 & - & 0.889 & 0.500 & 0.667 & 0.415\\
YOLOv8-m-world-v2~\cite{2024yoloworld} & 0.283 & 0.147 & 0.065 & 0.162 & 0.521 & 0.400 & \underline{0.667} & 0.824 & 0.111 & 0.889 & 0.500 & \textbf{1.000} & 0.464\\
YOLOv8-l-world-v2~\cite{2024yoloworld} & 0.249 & 0.176 & 0.049 & 0.147 & 0.500 & 0.500 & 0.609 & 0.750 & \textbf{0.333} & 0.857 & \underline{0.667} & 0.889 & \underline{0.478}\\
YOLOv8-x-world-v2~\cite{2024yoloworld} & \textbf{0.298} & 0.055 & \textbf{0.088} & 0.195 & 0.548 & 0.667 & 0.571 & 0.750 & 0.182 & \textbf{1.000} & \underline{0.667} & 0.750 & \textbf{0.481}\\
RT-DETR~\cite{2024RTDETR} & 0.145 & - & 0.033 &0.031  & - & - & 0.600 & \underline{0.876} & - & - & - & - & 0.140\\
RF-DETR-Base~\cite{rfdetr} &0.118  & 0.142 & 0.053 & 0.020 & 0.234 & 0.182 & 0.533 & 0.667 & 0.095 & 0.800 & - & - & 0.237\\ 
YOLO26-n~\cite{yolo26}  & 0.268 & 0.173 & 0.063 & 0.169 & 0.412 & 0.667 & 0.421 & - & - & 0.800 & - & \textbf{1.000} & 0.331\\ 
YOLO26-s~\cite{yolo26}   & 0.250 & 0.170 & 0.058 & 0.182 & 0.492 & \textbf{1.000} & 0.632 & 0.500 & - & 0.750 & - & 0.750 & 0.399\\ 
YOLO26-m~\cite{yolo26}   & 0.246 & \textbf{0.190} & 0.058 & 0.170 & 0.400 & 0.500 & 0.571 & 0.769  & - & 0.889 & 0.333 & 0.800 & 0.411 \\ 
YOLO26-l~\cite{yolo26}    & 0.268 & 0.133 & 0.020 & 0.181  & 0.392 & \textbf{1.000} & 0.250 & 0.222 & - & 0.857 & - & 0.333 & 0.305\\ 
YOLO26-x~\cite{yolo26}   & \underline{0.284} & 0.060 & 0.025 & \textbf{0.206} & \textbf{0.581} & \textbf{1.000} & 0.545 &0.714  &  \underline{0.222}& \textbf{1.000} & 0.182 &0.889  & 0.476\\  
\bottomrule
\end{tabular}}}
\end{threeparttable}
\end{table*}


\begin{table*}[!t]
\centering
\caption{Model comparison on PipeMFL-240K (Confidence score=0.25, IoU=0.5) with different sampling ratios (ratio=9:1: 90\% Positive + 10\% Negative samples).}
\label{tab:sample_240}
\begin{threeparttable}
\setlength{\tabcolsep}{0.8mm}{  
\begin{tabular}{p{2.4cm}@{\centering} 
               p{2cm}<{\centering}
               p{2cm}<{\centering}
               p{2cm}<{\centering}
               p{2.4cm}<{\centering}
               p{2.4cm}<{\centering}
               p{2.4cm}<{\centering}
}
\toprule
\textbf{Model}  & \textbf{Ratio}  & \textbf{mAP50} & \textbf{mAP50:95} & \textbf{Precision} & \textbf{Recall} & \textbf{F1-score}  \\
 &  &  &  & \textbf{(macro, micro)} & \textbf{(macro, micro)} & \textbf{(macro, micro)} \\
\midrule 
YOLOv8-s & 9:1 & 0.402 & 0.248 & 0.413, 0.361& 0.544, 0.369 & 0.449, 0.365   \\
YOLOv8-s & 5:5 & 0.341 & 0.216 & 0.433, 0.438 & 0.420, 0.286 & 0.395, 0.346   \\
YOLOv8-s & 3:7 & 0.185 & 0.140 & 0.306, 0.560 & 0.232, 0.181 & 0.221, 0.274   \\
YOLOv8-m & 9:1 & 0.450 & 0.288 & 0.448, 0.375& 0.596, 0.356 & 0.490, 0.365   \\
YOLOv8-m & 5:5 & 0.387 & 0.249 & 0.635, 0.489 & 0.455, 0.278 & 0.480, 0.354   \\
YOLOv8-m & 3:7 & 0.211 & 0.156 & 0.457, 0.565 & 0.263, 0.192 & 0.269, 0.282   \\
YOLOv8-l & 9:1 & 0.475 & 0.298 & 0.529, 0.410& 0.587, 0.368 & 0.541, 0.388   \\
YOLOv8-l & 5:5 & 0.414 & 0.262 & 0.542, 0.457 & 0.502, 0.307 & 0.487, 0.367   \\
YOLOv8-l & 3:7 & 0.248 & 0.180 & 0.638, 0.584 & 0.295, 0.158 & 0.323, 0.270   \\

\bottomrule
\end{tabular}}
\end{threeparttable}
\end{table*}

\begin{table*}[!t]
\centering
\caption{Comparison of AP50 for each category in models on PipeMFL-240K with different sampling ratios (ratio=9:1: 90\% Positive + 10\% Negative samples). "-" indicates not detected.}
\label{tab:sample_240_ap50}
\begin{threeparttable}
\resizebox{\textwidth}{!}{
\setlength{\tabcolsep}{0.65mm}{
\begin{tabular}{p{2cm}p{1cm}<{\centering}p{1cm}<{\centering}p{1cm}<{\centering}p{1cm}<{\centering}p{1cm}<{\centering}p{1cm}<{\centering}p{1cm}<{\centering}p{1cm}<{\centering}p{1cm}<{\centering}p{1cm}<{\centering}p{1cm}<{\centering}p{1cm}<{\centering}p{1cm}<{\centering}p{1cm}<{\centering}p{1cm}<{\centering}}
\toprule
& & \multicolumn{4}{c}{\textbf{Damage}} & \multicolumn{8}{c}{\textbf{Component}} \\
\cmidrule(lr){3-6} \cmidrule(lr){7-14}
\textbf{Model} & \textbf{Ratio} & \textbf{MTL} & \textbf{CRC} & \textbf{GWA} & \textbf{SWA} & \textbf{BND} & \textbf{SLE} & \textbf{BRN} & \textbf{TEE} & \textbf{CAS} & \textbf{VAL} & \textbf{ESP} & \textbf{FLA} & \textbf{Overall}  \\
\midrule
YOLOv8-s& 9:1 &0.252  &0.091 &0.043  &0.085  &0.514  &0.863  &0.352  &0.485  &0.420  &0.816  &0.569  & 0.333 & 0.402\\
YOLOv8-s& 5:5 & 0.212 & 0.069 & 0.035 & 0.057 & 0.501 & 0.857 & 0.270 & 0.606 & - & 0.817 & 0.667 & - & 0.341\\
YOLOv8-s& 3:7 & 0.148  & 0.045  & 0.023 & 0.029 & 0.383 & 0.857 & - & - & - & 0.714 & -  &  - &0.185 \\
YOLOv8-m& 9:1 & 0.243  & 0.113  & 0.051 & 0.090 & 0.571 & 0.871 & 0.362 & 0.771 & 0.531 & 0.891  & 0.651  &  0.250 & 0.450 \\
YOLOv8-m& 5:5 & 0.218  & 0.065  & 0.044 & 0.071 & 0.484 & 0.854 & 0.339 & 0.594 & 0.312 & 0.857  & 0.557  & 0.250 & 0.387\\
YOLOv8-m & 3:7 & 0.155 & 0.039 & 0.034 & 0.037 & 0.422 & 0.858 & 0.137 & 0.079 & -  &0.774 & - &- & 0.211\\ 
YOLOv8-l& 9:1 & 0.260  &0.105   & 0.065 & 0.108 & 0.587 &  0.854& 0.279 & 0.828 & 0.460 & 0.875  & 0.630  & 0.650 &0.475\\
YOLOv8-l& 5:5 & 0.231  & 0.081  & 0.050 & 0.084 & 0.508 & 0.878 & 0.290 & 0.645 & 0.353 & 0.840  & 0.514  & 0.500 &0.414\\
YOLOv8-l& 3:7 & 0.131 & 0.058  & 0.022 & 0.020 & 0.446 & 0.843 & 0.156 & 0.360 & 0.040 & 0.787  & 0.111  & - & 0.248\\

\bottomrule
\end{tabular}}}
\end{threeparttable}
\end{table*}

\begin{table*}[!t]
\centering
\caption{Comparison of Precision for each category in models on PipeMFL-240K with different sampling ratios (ratio=9:1: 90\% Positive + 10\% Negative samples). "-" indicates not detected.}
\label{tab:sample_240_p}
\begin{threeparttable}
\resizebox{\textwidth}{!}{
\setlength{\tabcolsep}{0.65mm}{
\begin{tabular}{p{2cm}p{1cm}<{\centering}p{1cm}<{\centering}p{1cm}<{\centering}p{1cm}<{\centering}p{1cm}<{\centering}p{1cm}<{\centering}p{1cm}<{\centering}p{1cm}<{\centering}p{1cm}<{\centering}p{1cm}<{\centering}p{1cm}<{\centering}p{1cm}<{\centering}p{1cm}<{\centering}p{1cm}<{\centering}p{1cm}<{\centering}}
\toprule
& & \multicolumn{4}{c}{\textbf{Damage}} & \multicolumn{8}{c}{\textbf{Component}} \\
\cmidrule(lr){3-6} \cmidrule(lr){7-14}
\textbf{Model} & \textbf{Ratio} & \textbf{MTL} & \textbf{CRC} & \textbf{GWA} & \textbf{SWA} & \textbf{BND} & \textbf{SLE} & \textbf{BRN} & \textbf{TEE} & \textbf{CAS} & \textbf{VAL} & \textbf{ESP} & \textbf{FLA} & \textbf{Overall}  \\
\midrule
YOLOv8-s& 9:1 &0.375  &0.122  &0.321  &0.295  &0.396  &0.461  &0.618  &0.500  &0.400  &0.719  &0.462  & 0.286 & 0.413\\
YOLOv8-s& 5:5 & 0.454 & 0.128 & 0.360 & 0.428 & 0.432 & 0.536 & 0.593 & 0.667 & - & 0.742 & 0.857 & - & 0.433\\
YOLOv8-s& 3:7 &0.581  & 0.190  & 0.510 & 0.582 & 0.454 & 0.646 & - & - & - & 0.714 & -  & - &0.306 \\
YOLOv8-m& 9:1 & 0.401  & 0.111  & 0.339 & 0.269 & 0.451 & 0.639 & 0.609 & 0.689 & 0.329 & 0.789  & 0.500  &  0.250 & 0.448 \\
YOLOv8-m& 5:5 & 0.501  & 0.181  & 0.432 & 0.476 & 0.458 & 0.559 & 0.760 & 0.793 & 0.727 & 0.733  & 1.000  & 1.000 & 0.635\\
YOLOv8-m & 3:7 & 0.580 & 0.208 & 0.543 & 0.567 & 0.487 & 0.680 & 0.667 & 1.000 & -  &0.750  & - & -  & 0.457\\ 
YOLOv8-l& 9:1 & 0.423  &0.175   & 0.332 & 0.338 & 0.491 &  0.642 & 0.613 & 0.846 & 0.546 & 0.793  & 0.546  & 0.600 &0.529\\
YOLOv8-l& 5:5 & 0.470  & 0.161  & 0.409 & 0.447 & 0.449 & 0.509 & 0.581 & 0.730 & 0.429 & 0.759  & 0.556  & 1.000 &0.542\\
YOLOv8-l& 3:7 & 0.606 & 0.249  & 0.562 & 0.603 & 0.484 & 0.769 & 0.727 & 0.875 & 1.000 & 0.778  & 1.000  & - & 0.638\\

\bottomrule
\end{tabular}}}
\end{threeparttable}
\end{table*}

\begin{table*}[!t]
\centering
\caption{Comparison of Recall for each category in models on PipeMFL-240K with different sampling ratios (ratio=9:1: 90\% Positive + 10\% Negative samples). "-" indicates not detected.}
\label{tab:sample_240_r}
\begin{threeparttable}
\resizebox{\textwidth}{!}{
\setlength{\tabcolsep}{0.65mm}{
\begin{tabular}{p{2cm}p{1cm}<{\centering}p{1cm}<{\centering}p{1cm}<{\centering}p{1cm}<{\centering}p{1cm}<{\centering}p{1cm}<{\centering}p{1cm}<{\centering}p{1cm}<{\centering}p{1cm}<{\centering}p{1cm}<{\centering}p{1cm}<{\centering}p{1cm}<{\centering}p{1cm}<{\centering}p{1cm}<{\centering}}
\toprule
& & \multicolumn{4}{c}{\textbf{Damage}} & \multicolumn{8}{c}{\textbf{Component}} \\
\cmidrule(lr){3-6} \cmidrule(lr){7-14}
\textbf{Model} & \textbf{Ratio} & \textbf{MTL} & \textbf{CRC} & \textbf{GWA} & \textbf{SWA} & \textbf{BND} & \textbf{SLE} & \textbf{BRN} & \textbf{TEE} & \textbf{CAS} & \textbf{VAL} & \textbf{ESP} & \textbf{FLA} & \textbf{Overall}  \\
\midrule
YOLOv8-s& 9:1 &0.413  &0.305  &0.097  &0.208  &0.816  &0.914  &0.467  &0.579  &0.560  &1.000  &0.667  & 0.500 & 0.544\\
YOLOv8-s& 5:5 & 0.325 & 0.209 & 0.067 & 0.108 & 0.778 & 0.897 & 0.356 & 0.632 & - & 1.000 & 0.667 & - & 0.420\\
YOLOv8-s& 3:7 &0.206 & 0.096  & 0.037 & 0.046 & 0.654 & 0.879 & - & - & - & 0.870 & -  & - &0.232 \\
YOLOv8-m& 9:1 & 0.390  & 0.401  & 0.112 & 0.221 & 0.880 & 0.900 & 0.508 & 0.820 & 0.640 & 1.000  & 0.778  &  0.500 & 0.596 \\
YOLOv8-m& 5:5 & 0.312  & 0.167  & 0.076 & 0.121 & 0.772 & 0.897 & 0.422 & 0.605 & 0.320 & 0.957  & 0.557  & 0.250 & 0.455\\
YOLOv8-m & 3:7 & 0.215 & 0.096 & 0.054 & 0.057 & 0.686 & 0.879 & 0.178 & 0.079 & -  &0.913  & - & - & 0.263\\ 
YOLOv8-l& 9:1 & 0.403  &0.310   & 0.136 & 0.218 & 0.889 &  0.897& 0.422 & 0.868 & 0.480 & 1.000  & 0.667  & 0.750 &0.587\\
YOLOv8-l& 5:5 & 0.343  & 0.230  & 0.091 & 0.143 & 0.804 & 0.931 & 0.400 & 0.711 & 0.360 & 0.957  & 0.556  & 0.500 &0.502\\
YOLOv8-l& 3:7 & 0.175 & 0.141  & 0.034 & 0.031 & 0.689 & 0.862 & 0.178 & 0.368 & 0.040 & 0.913  & 0.111  & - & 0.295\\

\bottomrule
\end{tabular}}}
\end{threeparttable}
\end{table*}

\begin{table*}[!t]
\centering
\caption{Comparison of F1-score for each category in models on PipeMFL-240K with different sampling ratios (ratio=9:1: 90\% Positive + 10\% Negative samples). "-" indicates not detected.}
\label{tab:sample_240_f1}
\begin{threeparttable}
\resizebox{\textwidth}{!}{
\setlength{\tabcolsep}{0.65mm}{
\begin{tabular}{p{2cm}p{1cm}<{\centering}p{1cm}<{\centering}p{1cm}<{\centering}p{1cm}<{\centering}p{1cm}<{\centering}p{1cm}<{\centering}p{1cm}<{\centering}p{1cm}<{\centering}p{1cm}<{\centering}p{1cm}<{\centering}p{1cm}<{\centering}p{1cm}<{\centering}p{1cm}<{\centering}p{1cm}<{\centering}}
\toprule
& & \multicolumn{4}{c}{\textbf{Damage}} & \multicolumn{8}{c}{\textbf{Component}} \\
\cmidrule(lr){3-6} \cmidrule(lr){7-14}
\textbf{Model} & \textbf{Ratio} & \textbf{MTL} & \textbf{CRC} & \textbf{GWA} & \textbf{SWA} & \textbf{BND} & \textbf{SLE} & \textbf{BRN} & \textbf{TEE} & \textbf{CAS} & \textbf{VAL} & \textbf{ESP} & \textbf{FLA} & \textbf{Overall}  \\
\midrule
YOLOv8-s& 9:1 &0.393  &0.175  &0.148  &0.244  &0.533  &0.613  &0.532  &0.537  &0.467  &0.836  &0.545  & 0.364 & 0.449\\
YOLOv8-s& 5:5 & 0.378 & 0.159 & 0.112 & 0.172 & 0.555 & 0.671 & 0.444 & 0.649 & - & 0.852 & 0.750 & - & 0.395\\
YOLOv8-s& 3:7 &0.305  & 0.127  & 0.068 & 0.085 & 0.536 & 0.745 & - & - & - & 0.784 & -  & - &0.221 \\
YOLOv8-m& 9:1 & 0.395  & 0.174  & 0.168 & 0.243 & 0.596 & 0.747 & 0.554 & 0.749 & 0.434 & 0.882  & 0.609  &  0.333 & 0.490 \\
YOLOv8-m& 5:5 & 0.384  & 0.174  & 0.129 & 0.193 & 0.575 & 0.689 & 0.543 & 0.687 & 0.444 & 0.830  & 0.714  & 0.400 & 0.480\\
YOLOv8-m & 3:7 & 0.313 & 0.131 & 0.098 & 0.104 & 0.570 & 0.767 & 0.281 & 0.146 & -  &0.824  & - & -  & 0.269\\ 
YOLOv8-l& 9:1 & 0.413  &0.224   & 0.193 & 0.265 & 0.633 &  0.748& 0.500 & 0.857 & 0.511 & 0.885  & 0.600  & 0.667 &0.541\\
YOLOv8-l& 5:5 & 0.397  & 0.190  & 0.149 & 0.217 & 0.577 & 0.659 & 0.474 & 0.720 & 0.391 & 0.846  & 0.556  & 0.667 &0.487\\
YOLOv8-l& 3:7 & 0.272 & 0.180  & 0.064 & 0.058 & 0.568 & 0.813 & 0.286 & 0.518 & 0.077 & 0.840  & 0.200  & - & 0.323\\

\bottomrule
\end{tabular}}}
\end{threeparttable}
\end{table*}

\begin{table*}[!t]
\centering
\caption{Model comparison on zero-shot testing set (Confidence score=0.25, IoU=0.5) with different sampling ratios (ratio=9:1: 90\% Positive + 10\% Negative samples).}
\label{tab:sample_zero}
\begin{threeparttable}
\setlength{\tabcolsep}{0.8mm}{  
\begin{tabular}{p{2.4cm}@{\centering} 
               p{2cm}<{\centering}
               p{2cm}<{\centering}
               p{2cm}<{\centering}
               p{2.4cm}<{\centering}
               p{2.4cm}<{\centering}
               p{2.4cm}<{\centering}
}
\toprule
\textbf{Model}  & \textbf{Ratio}  & \textbf{mAP50} & \textbf{mAP50:95} & \textbf{Precision} & \textbf{Recall} & \textbf{F1-score}  \\
 &  &  &  & \textbf{(macro, micro)} & \textbf{(macro, micro)} & \textbf{(macro, micro)} \\
\midrule 
YOLOv8-s & 9:1 & 0.347 & 0.204 & 0.354, 0.165& 0.432, 0.403 & 0.367, 0.234   \\
YOLOv8-s & 5:5 & 0.234 & 0.161 & 0.287, 0.222 & 0.302, 0.273 & 0.270, 0.273   \\
YOLOv8-s & 3:7 & 0.217 & 0.151 & 0.401, 0.355 & 0.243, 0.301 & 0.265, 0.325   \\
YOLOv8-m & 9:1 & 0.442 & 0.255 & 0.352, 0.155& 0.630, 0.350 & 0.407, 0.216   \\
YOLOv8-m & 5:5 & 0.368 & 0.218 & 0.401, 0.211 & 0.432, 0.382 & 0.384, 0.271   \\
YOLOv8-m & 3:7 & 0.222 & 0.157 & 0.357, 0.332 & 0.251, 0.319 & 0.263, 0.325   \\
YOLOv8-l & 9:1 & 0.411 & 0.251 & 0.424, 0.200& 0.482, 0.409 & 0.445, 0.268  \\
YOLOv8-l & 5:5 & 0.345 & 0.199 & 0.483, 0.228 & 0.398, 0.366 & 0.387, 0.281   \\
YOLOv8-l & 3:7 & 0.227 & 0.170 & 0.474, 0.323 & 0.259, 0.336 & 0.330, 0.341   \\

\bottomrule
\end{tabular}}
\end{threeparttable}
\end{table*}

\begin{table*}[!t]
\centering
\caption{Comparison of AP50 for each category in models on zero-shot testing set with different sampling ratios (ratio=9:1: 90\% Positive + 10\% Negative samples). "-" indicates not detected.}
\label{tab:sample_zero_ap50}
\begin{threeparttable}
\resizebox{\textwidth}{!}{
\setlength{\tabcolsep}{0.65mm}{
\begin{tabular}{p{2cm}p{1cm}<{\centering}p{1cm}<{\centering}p{1cm}<{\centering}p{1cm}<{\centering}p{1cm}<{\centering}p{1cm}<{\centering}p{1cm}<{\centering}p{1cm}<{\centering}p{1cm}<{\centering}p{1cm}<{\centering}p{1cm}<{\centering}p{1cm}<{\centering}p{1cm}<{\centering}p{1cm}<{\centering}p{1cm}<{\centering}}
\toprule
& & \multicolumn{4}{c}{\textbf{Damage}} & \multicolumn{8}{c}{\textbf{Component}} \\
\cmidrule(lr){3-6} \cmidrule(lr){7-14}
\textbf{Model} & \textbf{Ratio} & \textbf{MTL} & \textbf{CRC} & \textbf{GWA} & \textbf{SWA} & \textbf{BND} & \textbf{SLE} & \textbf{BRN} & \textbf{TEE} & \textbf{CAS} & \textbf{VAL} & \textbf{ESP} & \textbf{FLA} & \textbf{Overall}  \\
\midrule
YOLOv8-s& 9:1 &0.158  &0.012  &0.019  &0.054  &0.257  &1.000  &0.377  &0.283  & - &0.889  & -  & 1.000 & 0.347\\
YOLOv8-s& 5:5 & 0.193 & 0.029 & 0.005 & 0.079 & 0.175 & 1.000 & 0.417 & 0.344 & - & 0.563 & - & - & 0.234\\
YOLOv8-s& 3:7 &0.213  & 0.012  & 0.018 & 0.061 & 0.170 & 1.000 & - & 0.125 & - & 1.000 & -  & - &0.217 \\
YOLOv8-m& 9:1 & 0.132  & 0.078  & 0.016 & 0.042 & 0.287 & 1.000 & 0.485 & 0.542 & 0.100 & 1.000  & 1.000  &  0.625 & 0.442 \\
YOLOv8-m& 5:5 & 0.219  & 0.032  & 0.005 & 0.095 & 0.292 & 1.000 & 0.500 & 0.518 & - & 1.000  & - & 0.750 & 0.368\\
YOLOv8-m & 3:7 & 0.216 & 0.004 & 0.009 & 0.078 & 0.361 & 1.000 & - & - & -  &1.000  & - &- & 0.222\\ 
YOLOv8-l& 9:1 & 0.204  &0.046   & 0.023 & 0.088 & 0.447 &  1.000 & 0.505 & 0.875 & - & 1.000  & -  & 1.000 &0.411\\
YOLOv8-l& 5:5 & 0.213  & 0.024  & 0.004 & 0.098 & 0.302 & 1.000 & 0.167 & 0.625 & - & 0.950  & -  & 0.750 &0.345\\
YOLOv8-l& 3:7 & 0.226 & 0.003  & 0.018 & 0.104 & 0.293 & 1.000 & 0.083 & - & - & 1.000  & - & - & 0.227\\

\bottomrule
\end{tabular}}}
\end{threeparttable}
\end{table*}

\begin{table*}[!t]
\centering
\caption{Comparison of Precision for each category in models on zero-shot testing set with different sampling ratios (ratio=9:1: 90\% Positive + 10\% Negative samples). "-" indicates not detected.}
\label{tab:sample_zero_p}
\begin{threeparttable}
\resizebox{\textwidth}{!}{
\setlength{\tabcolsep}{0.65mm}{
\begin{tabular}{p{2cm}p{1cm}<{\centering}p{1cm}<{\centering}p{1cm}<{\centering}p{1cm}<{\centering}p{1cm}<{\centering}p{1cm}<{\centering}p{1cm}<{\centering}p{1cm}<{\centering}p{1cm}<{\centering}p{1cm}<{\centering}p{1cm}<{\centering}p{1cm}<{\centering}p{1cm}<{\centering}p{1cm}<{\centering}p{1cm}<{\centering}}
\toprule
& & \multicolumn{4}{c}{\textbf{Damage}} & \multicolumn{8}{c}{\textbf{Component}} \\
\cmidrule(lr){3-6} \cmidrule(lr){7-14}
\textbf{Model} & \textbf{Ratio} & \textbf{MTL} & \textbf{CRC} & \textbf{GWA} & \textbf{SWA} & \textbf{BND} & \textbf{SLE} & \textbf{BRN} & \textbf{TEE} & \textbf{CAS} & \textbf{VAL} & \textbf{ESP} & \textbf{FLA} & \textbf{Overall}  \\
\midrule
YOLOv8-s& 9:1 &0.171  &0.138  &0.023  &0.098  &0.486  &0.333  &0.600  &0.600  & -  &0.800  & -  & 1.000 & 0.354\\
YOLOv8-s& 5:5 & 0.234 & 0.227 & 0.044 & 0.260 & 0.375 & 0.500 & 0.625 & 0.750 & - & 0.429 & - & - & 0.287\\
YOLOv8-s& 3:7 &0.353  & 0.304  & 0.111 & 0.522 & 0.526 & 1.000 & - & 1.000 & - & 1.000 & - &  - &0.401 \\
YOLOv8-m& 9:1 & 0.172  & 0.133  & 0.029 & 0.063 & 0.425 & 0.250 & 0.533 & 0.833 & 0.053 & 0.800  & 0.333  &  0.600 & 0.352 \\
YOLOv8-m& 5:5 & 0.212  & 0.278  & 0.040 & 0.191 & 0.514 & 0.333 & 0.857 & 0.714 & - & 0.667  &-  & 1.000 & 0.401\\
YOLOv8-m & 3:7 & 0.326 & 0.667 & 0.111 & 0.483 & 0.696 & 1.000 & - & - & - & 1.000 & - & - & 0.357\\ 
YOLOv8-l& 9:1 & 0.207  &0.128   & 0.064 & 0.126 & 0.556 &  1.000 & 0.636 & 0.875 & - & 1.000  & -  & 0.500 &0.424\\
YOLOv8-l& 5:5 & 0.233  & 0.414  & 0.046 & 0.236 & 0.567 & 0.500 & 1.000 & 1.000 & - & 0.800  & -  & 1.000 &0.483\\
YOLOv8-l& 3:7 & 0.314 & 1.000  & 0.500 & 0.408 & 0.667 & 1.000 & 1.000 & - & - & 0.800  & -  & - & 0.474\\

\bottomrule
\end{tabular}}}
\end{threeparttable}
\end{table*}

\begin{table*}[!t]
\centering
\caption{Comparison of Recall for each category in models on zero-shot testing set with different sampling ratios (ratio=9:1: 90\% Positive + 10\% Negative samples). "-" indicates not detected.}
\label{tab:sample_zero_r}
\begin{threeparttable}
\resizebox{\textwidth}{!}{
\setlength{\tabcolsep}{0.65mm}{
\begin{tabular}{p{2cm}p{1cm}<{\centering}p{1cm}<{\centering}p{1cm}<{\centering}p{1cm}<{\centering}p{1cm}<{\centering}p{1cm}<{\centering}p{1cm}<{\centering}p{1cm}<{\centering}p{1cm}<{\centering}p{1cm}<{\centering}p{1cm}<{\centering}p{1cm}<{\centering}p{1cm}<{\centering}p{1cm}<{\centering}}
\toprule
& & \multicolumn{4}{c}{\textbf{Damage}} & \multicolumn{8}{c}{\textbf{Component}} \\
\cmidrule(lr){3-6} \cmidrule(lr){7-14}
\textbf{Model} & \textbf{Ratio} & \textbf{MTL} & \textbf{CRC} & \textbf{GWA} & \textbf{SWA} & \textbf{BND} & \textbf{SLE} & \textbf{BRN} & \textbf{TEE} & \textbf{CAS} & \textbf{VAL} & \textbf{ESP} & \textbf{FLA} & \textbf{Overall}  \\
\midrule
YOLOv8-s& 9:1 &0.434  &0.064  &0.054  &0.245  &0.515  &1.000  &0.500  &0.375  & -  &1.000  & -  & 1.000 & 0.432\\
YOLOv8-s& 5:5 & 0.424 & 0.070 & 0.054 & 0.171 & 0.364 & 1.000 & 0.417 & 0.375 & - & 0.750 & - & - & 0.302\\
YOLOv8-s& 3:7 &0.370  & 0.022  & 0.018 & 0.083 & 0.303 & 1.000 & - & 0.125 & - & 1.000 & - & - & 0.243 \\
YOLOv8-m& 9:1 & 0.368  & 0.278  & 0.125 & 0.228 & 0.515 & 1.000 & 0.667 & 0.625 & 1.000 & 1.000  & 1.000  &  0.750 & 0.630 \\
YOLOv8-m& 5:5 & 0.451  & 0.064  & 0.018 & 0.231 & 0.546 & 1.000 & 0.500 & 0.625 & - & 1.000  & -  & 0.750 & 0.432\\
YOLOv8-m & 3:7 & 0.386 & 0.006 & 0.018 & 0.114 & 0.485 & 1.000 & - & - & - &1.000  & - & -  & 0.251\\ 
YOLOv8-l& 9:1 & 0.434  & 0.144  & 0.125 & 0.271 & 0.606 & 1.000 & 0.583 & 0.875 & - & 1.000  & -  & 1.000 &0.482\\
YOLOv8-l& 5:5 & 0.430  & 0.038  & 0.036 & 0.220 & 0.515 & 1.000 & 0.167 & 0.625 & - & 1.000  & -  & 0.750 &0.398\\
YOLOv8-l& 3:7 & 0.403 & 0.003  & 0.018 & 0.173 & 0.424 & 1.000 & 0.083 & - & - & 1.000  & - & - & 0.259\\

\bottomrule
\end{tabular}}}
\end{threeparttable}
\end{table*}

\begin{table*}[!t]
\centering
\caption{Comparison of F1-score for each category in models on zero-shot testing set with different sampling ratios (ratio=9:1: 90\% Positive + 10\% Negative samples). "-" indicates not detected.}
\label{tab:sample_zero_f1}
\begin{threeparttable}
\resizebox{\textwidth}{!}{
\setlength{\tabcolsep}{0.65mm}{
\begin{tabular}{p{2cm}p{1cm}<{\centering}p{1cm}<{\centering}p{1cm}<{\centering}p{1cm}<{\centering}p{1cm}<{\centering}p{1cm}<{\centering}p{1cm}<{\centering}p{1cm}<{\centering}p{1cm}<{\centering}p{1cm}<{\centering}p{1cm}<{\centering}p{1cm}<{\centering}p{1cm}<{\centering}p{1cm}<{\centering}}
\toprule
& & \multicolumn{4}{c}{\textbf{Damage}} & \multicolumn{8}{c}{\textbf{Component}} \\
\cmidrule(lr){3-6} \cmidrule(lr){7-14}
\textbf{Model} & \textbf{Ratio} & \textbf{MTL} & \textbf{CRC} & \textbf{GWA} & \textbf{SWA} & \textbf{BND} & \textbf{SLE} & \textbf{BRN} & \textbf{TEE} & \textbf{CAS} & \textbf{VAL} & \textbf{ESP} & \textbf{FLA} & \textbf{Overall}  \\
\midrule
YOLOv8-s& 9:1 &0.246  &0.087  &0.032  &0.140  &0.500  &0.500  &0.545  &0.462  & -  &0.889  & -  & 1.000 & 0.367\\
YOLOv8-s& 5:5 & 0.301 & 0.107 & 0.048 & 0.206 & 0.369 & 0.667 & 0.500 & 0.500 & - & 0.545 & - & - & 0.270\\
YOLOv8-s& 3:7 &0.361  & 0.042  & 0.031 & 0.142 & 0.385 & 1.000 & - & 0.222 & - & 1.000 & - & - &0.265 \\
YOLOv8-m& 9:1 & 0.234  & 0.180  & 0.047 & 0.099 & 0.466 & 0.400 & 0.593 & 0.714 & 0.100 & 0.889  & 0.500  &  0.667 & 0.407 \\
YOLOv8-m& 5:5 & 0.288  & 0.104  & 0.025 & 0.209 & 0.529 & 0.500 & 0.632 & 0.667 & - & 0.800  &-  & 0.857 & 0.384\\
YOLOv8-m & 3:7 & 0.353 & 0.013 & 0.031 & 0.185 & 0.571 & 1.000 & - & - & -  &1.000  & - & -  & 0.263\\ 
YOLOv8-l& 9:1 & 0.281  &0.136   & 0.085 & 0.172 & 0.580 &  1.000& 0.609 & 0.875 & - & 1.000  & -  & 1.000 &0.445\\
YOLOv8-l& 5:5 & 0.302  & 0.070  & 0.040 & 0.228 & 0.540 & 0.667 & 0.286 & 0.769 & - & 0.889  & - & 0.857 &0.387\\
YOLOv8-l& 3:7 & 0.353 & 0.006  & 0.035 & 0.243 & 0.518 & 1.000 & 0.154 & - & - & 0.889  & -  & - & 0.330\\

\bottomrule
\end{tabular}}}
\end{threeparttable}
\end{table*}

\section{Broader Impact}
This work introduces PipeMFL-240K, a large-scale, publicly available, carefully annotated dataset and benchmark for complex object detection in pipeline MFL imagery. By covering various damage types, structural components, inspection scenes and pipe configurations under real in-service conditions, this dataset provides a standardized foundation for advancing research on automated pipeline inspection and structural integrity assessment.

The authors do not anticipate any negative social impacts resulting from the release of this dataset. On the contrary, the expected impact is largely positive. PipeMFL-240K enables the development and rigorous evaluation of robust inspection algorithms that can better handle extreme category imbalance, scale variation and strong structural context, which are common challenges in real-world industrial inspection scenarios. By reducing the reliance on manual inspection and expert-driven annotation, the proposed benchmark can facilitate more efficient and consistent condition monitoring of critical pipeline infrastructure.

In a broader industrial and societal context, this work has the potential to improve the safety and reliability of energy transportation systems by supporting earlier and more accurate detection of safety-critical defects. Such advances may help prevent pipeline failures, reduce environmental risks and lower maintenance costs. Furthermore, by providing an open and reproducible benchmark, this dataset promotes transparent comparison across methods and accelerates the translation of machine learning research into practical, trustworthy inspection systems, particularly in large-scale or resource-constrained operational environments.

\end{document}